\documentclass[3p,times,twocolumn]{elsarticle}

\usepackage{geometry}

\usepackage{amsmath,amsfonts,amssymb,amsthm}
\usepackage{mathtools}
\usepackage{bm}
\usepackage[T1]{fontenc}

\usepackage[linesnumbered,ruled,vlined]{algorithm2e}
\SetAlFnt{\small}
\SetAlCapFnt{\small}
\SetAlCapNameFnt{\small}

\let\oldnl\nl
\newcommand{\nonl}{\renewcommand{\nl}{\stepcounter{AlgoLine}\let\nl\oldnl}}

\usepackage{xcolor}
\usepackage{graphicx}
\usepackage{color}

\usepackage{array}
\usepackage{tabularx}
\usepackage{multirow}
\usepackage{float}
\usepackage{stfloats}

\usepackage{subfigure}
\usepackage[caption=false,font=normalsize,labelfont=sf,textfont=sf]{subfig}

\usepackage{url}
\usepackage{verbatim}
\usepackage{CJKutf8}

\SetKwInput{KwIn}{Input}
\SetKwInput{KwOut}{Output}
\SetKwInput{KwRet}{Return}
\DontPrintSemicolon  

\SetKwComment{tcp}{// }{} 
\SetCommentSty{mycommentstyle}

\usepackage{xr}
\makeatletter
\newcommand*{\addFileDependency}[1]{
  \typeout{(#1)}
  \@addtofilelist{#1}
  \IfFileExists{#1}{}{\typeout{No file #1.}}
}
\makeatother

\newcommand*{\myexternaldocument}[1]{
    \externaldocument{#1}
    \addFileDependency{#1.tex}
    \addFileDependency{#1.aux}
}
\myexternaldocument{Supplementary_materials}

\journal{Journal}

\begin{document}

\begin{frontmatter}
\title{An adaptive experience-based discrete genetic algorithm for multi-trip picking robot task scheduling in smart orchards}

\author[1]{Peng Chen}
\author[2,1]{Jing Liang\corref{cor1}}
\ead{liangjing@zzu.edu.cn}
\author[1]{Kang-Jia Qiao}
\author[3]{Hui Song}
\author[1]{Cai-Tong Yue}
\author[1,4]{Kun-Jie Yu}
\author[5]{Ponnuthurai Nagaratnam Suganthan} 
\author[6,7]{Witold Pedrycz} 

\cortext[cor1]{Corresponding author}

\address[1]{School of Electrical and Information Engineering, Zhengzhou University, Zhengzhou 450007, China}
\address[2]{School of Electrical Engineering and Automation, Henan Institute of Technology, Xinxiang 453003, China}
\address[3]{School of Engineering, RMIT University, Melbourne, VIC, 3000, Australia}
\address[4]{Longmen Laboratory, Luoyang 471000, China}
\address[5]{KINDI Center for Computing Research, College of Engineering, Qatar University, Doha, Qatar}
\address[6]{Department of Electrical and Computer Engineering, University of Alberta, Edmonton, Canada}
\address[7]{Systems Research Institute of the Polish Academy of Sciences, Warsaw, Poland}

\begin{abstract}
The continuous innovation of smart robotic technologies is driving the development of smart orchards, significantly enhancing the potential for automated harvesting systems. While multi-robot systems offer promising solutions to address labor shortages and rising costs, the efficient scheduling of these systems presents complex optimization challenges. This research investigates the multi-trip picking robot task scheduling (MTPRTS) problem. The problem is characterized by its provision for robot redeployment while maintaining strict adherence to makespan constraints, and encompasses the interdependencies among robot weight, robot load, and energy consumption, thus introducing substantial computational challenges that demand sophisticated optimization algorithms.To effectively tackle this complexity, metaheuristic approaches, which often utilize local search mechanisms, are widely employed. Despite the critical role of local search in vehicle routing problems, most existing algorithms are hampered by redundant local operations, leading to slower search processes and higher risks of local optima, particularly in large-scale scenarios. To overcome these limitations, we propose an adaptive experience-based discrete genetic algorithm (AEDGA) that introduces three key innovations: (1) integrated load-distance balancing initialization method, (2) a clustering-based local search mechanism, and (3) an experience-based adaptive selection strategy. To ensure solution feasibility under makespan constraints, we develop a solution repair strategy implemented through three distinct frameworks. Comprehensive experiments on 18 proposed test instances and 24 existing test problems demonstrate that AEDGA significantly outperforms eight state-of-the-art algorithms.
\end{abstract}


\end{frontmatter}

\section{Introduction}
Despite the transformative impact of smart agricultural technologies on automated harvesting systems~\cite{sharma2022technological,muangprathub2019iot}, high-value fruit harvesting in orchard environments remains heavily reliant on manual labor~\cite{davidson2016proof}, creating persistent operational challenges. These challenges are particularly acute given the escalating labor costs and widespread workforce shortages~\cite{calvin2010us}. While existing works have explored various autonomous harvesting solutions~\cite{baeten2008autonomous}, single-robot implementations have proven insufficient for large-scale orchard operations~\cite{dai2023multi}. This limitation has driven increased attention toward multi-robot systems~\cite{li2020multi,qian2024cooperative,zhao2024multi}, where the fundamental challenge lies in optimizing task distribution and scheduling to maximize operational efficiency while minimizing resource utilization~\cite{wang2024multi,guo2024effective}.

The multi-trip picking robot task scheduling (MTPRTS) in smart orchards, though currently understudied, represents a critical step toward next-generation agriculture. Its significance extends beyond immediate operational optimization to potentially advance the broader field of agricultural robotics, including applications in pruning, spraying, and fruit harvesting, making it a cornerstone technology for future smart farming systems. From an algorithmic perspective, MTPRTS can be viewed as a specialized variant of the capacitated vehicle routing problem (CVRP)~\textcolor{blue}{\cite{zhao2025q,zhang2025deep}} adapted for agricultural scenarios. Its complexity surpasses that of the traditional CVRP, introducing additional characteristics to this NP-hard problem~\cite{cattaruzza2016vehicle}. The distinguishing characteristic of MTPRTS lies in its allowance for multiple trips per robot within a specified makespan~\cite{raeesi2019multi}, considering the interrelationship among robot weight, robot load, and energy consumption~\textcolor{blue}{\cite{chen2025reinforced}}, thus presenting significant algorithmic complexities within the optimization paradigm. This extension introduces interconnected challenges: simultaneous task allocation, multiple-trip coordination, capacity constraint management, and the critical relationship between real-time load and energy consumption. Moreover, the cumulative duration of trips assigned to each robot must satisfy predetermined operational time constraints.

Currently, due to the limited research devoted to MTPRTS, our investigation centered on examining the recent advances in multi-trip CVRP (MTVRP) research: 

The MTVRP predominantly arises in logistics and transportation operations within small to moderate-scale urban environments. This class of problems similarly incorporates multiple deployments of individual robots within the same problem instance. Given its computational complexity, exact algorithms are impractical for obtaining optimal solutions within operational time constraints~\cite{cattaruzza2018vehicle}, making heuristic algorithms better suited for practical applications~\cite{wassan2017multiple}. The evolutionary development of MTVRP solutions began with Fleischmann et al.~\cite{fleischmann1990vehicle}, who combined an improved savings heuristic with the bin packing (BP) heuristic. This foundation led to Taillard et al.'s~\cite{taillard1996vehicle} hierarchical approach: 1) using a tabu search (TS) algorithm to generate multiple trips from VRP solutions; 2) generating multiple VRP solutions based on these trips; and 3) employing the BP process to generate feasible MTVRP solutions. Subsequently, Salhi et al.~\cite{salhi2007ga} applied a genetic algorithm (GA), where chromosomes represent ordered circular sectors and VRP is solved within these sectors using the savings heuristic, followed by BP to generate MTVRP solutions. Olivera et al.~\cite{olivera2007adaptive} introduced an adaptive memory programming approach, combining TS and BP heuristics to solve MTVRP by generating an initial VRP solution through a "sweeping method" and continuously improving it within a memory pool.

Although these algorithms provide viable solutions to MTVRP, their effectiveness is constrained by the absence of systematic local search mechanisms. Local search plays a crucial role in solution refinement by iteratively exploring neighboring solutions. To enhance the solution quality, Cattaruzza et al.~\cite{cattaruzza2014memetic} proposed a memory-based algorithm, which partitions the customer sequence and improves solutions of MTVRP through local search. Following a similar line of thought, Francois et al.~\cite{franccois2016large} introduced specific operators for large neighborhood search algorithms, yielding two distinct algorithms for solving route generation and trip allocation phases. Florian et al.~\cite{arnold2019knowledge} argued that local search alone could generate high-quality solutions quickly, integrating three local search techniques into a single algorithm. Building upon this foundation, Mohammad et al.~\cite{sajid2022routing} used a GA for optimal customer ordering and employed simulated annealing (SA) for local search to avoid local optima. Li et al.~\cite{li2022ant} applied a greedy strategy to locally adjust the preferences of ants, but this slowed progress when dealing with larger customer sets. Furthermore, Zhao et al.~\cite{zhao2024hybrid} incorporated an extensive set of eight local search operations, applying them systematically to offspring solutions throughout each generation. These algorithms improve trip structures by applying local search to all trips in the solution space. While effective for smaller problem sizes, the redundancy of local search operations slows down the search process and can lead to local optima as the problem size increases.

The methodologies developed for MTVRP provide valuable insights and significant reference value for designing MTPRTS algorithms. To overcome the aforementioned limitations, this paper introduces an adaptive experience-based discrete genetic algorithm (AEDGA), consisting of route generation (RG) and route scheduling (RS). During RG, the integrated load-distance balancing initialization method accelerates population convergence by generating high-quality initial solutions. Instead of exhaustively applying local search to all solutions, AEDGA dynamically selects local search targets based on historical experience, significantly reducing computational overhead. Furthermore, a clustering-based local search mechanism is employed to prioritize high-potential trips and their adjacent trips, enabling focused exploration of promising solution spaces while avoiding redundant searches in less promising areas. The synergy of these innovative mechanisms enables AEDGA to effectively handle large-scale problems. Three frameworks with infeasible solution repair procedure are then proposed for RS to allocate the generated trips to a limited team of robots using the BP approach. These two phases are combined to address the MTPRTS problem in smart orchards. The main contributions of this paper are as follows:
\begin{itemize}
 \item A mathematical model for solving the MTPRTS problem is proposed, tailored to the practical conditions of orchards. Furthermore, a mixed integer linear programming (MILP) model is formulated based on this mathematical framework.
 \item An AEDGA for route generation and optimization is proposed, including the integrated load-distance balancing initialization method, a clustering-based local search mechanism, and an experience-based adaptive selection strategy. Three frameworks for repairing infeasible solutions during trip allocation are also designed.
 \item A test set consisting of $18$ proposed test instances and $24$ existing test problems is proposed. Experimental results demonstrate that AEDGA is competitive with \textcolor{blue}{eight} representative algorithms.
\end{itemize}

The structure of this paper is as follows: Section~\ref{section2} provides a description and modeling of the problem. Section~\ref{Problem solving methodology} details the processing procedure of the AEDGA. Section~\ref{section 4} presents the experimental design and analysis of the results. Finally, Section~\ref{section 5} concludes the paper and discusses the prospects for future research.

\section{Problem description and modeling}
\label{section2}
\subsection{Problem description}

\begin{figure}[htp]
    \centering
    \includegraphics[width=8.5cm]{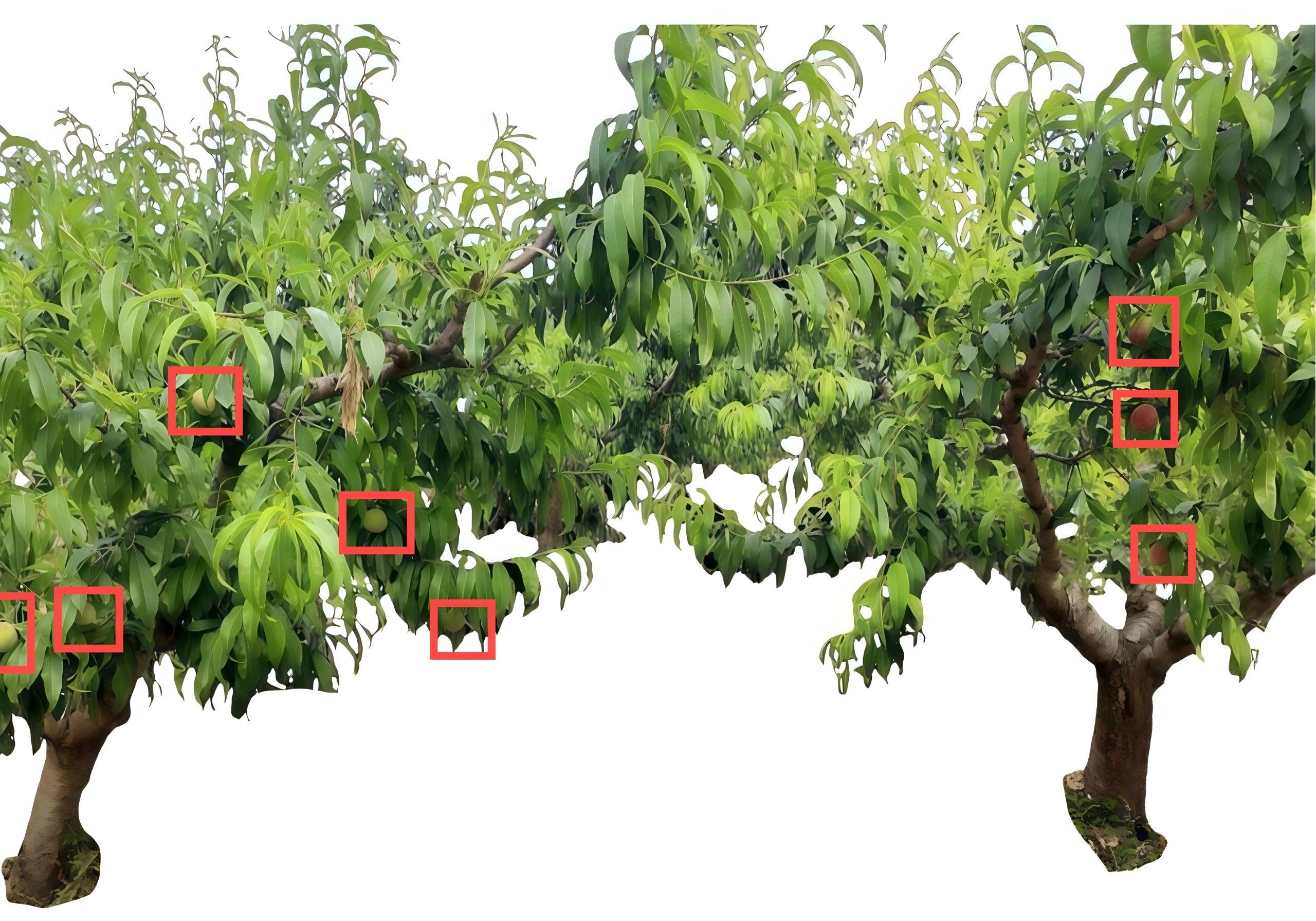}
    \caption{Orchard real shot}
    \label{Orchard real shot}
\end{figure}

\begin{figure}[htp]
    \centering
    \includegraphics[width=9cm]{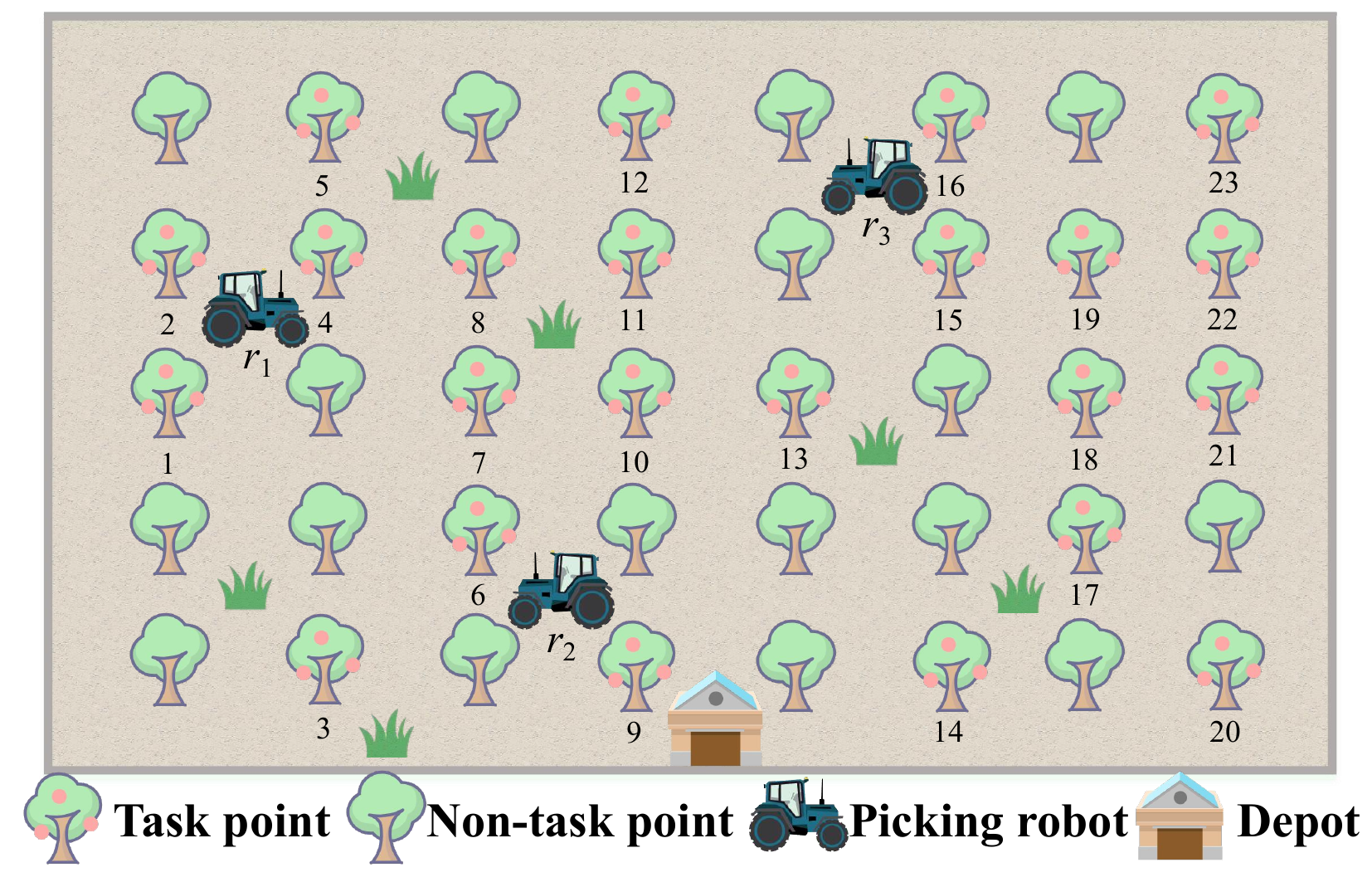}
    \caption{\textcolor{blue}{Schematic diagram of orchard scene}}
    \label{fruit trees}
\end{figure}

The MTPRTS problem can be described as follows: based on our field study, variations in environmental factors cause slight differences in the ripening times and yields of trees within the same orchard, as illustrated in Fig.~\ref{Orchard real shot}. We consider an orchard as shown in Fig.~\ref{fruit trees}, where each tree is uniformly planted. A tree whose proportion of ripe fruit reaches a specified threshold is designated as a picking task. In this context, trees bearing red fruit indicate task points, while others are considered obstacles.

To ensure the freshness of the fruit, all fruit at task points must be picked. The orchard contains $n$ task points, with $m$ picking robots stationed in the depot. Each task point is assigned to a single picking robot. All robots depart simultaneously from the depot in an unloaded state, complete the tasks along their assigned trips, and then return to the depot to unload. If tasks remain uncompleted, the robots redeploy from the depot to initiate another trip for the next set of tasks. \textcolor{blue}{For instance, when robot $r_1$ needs to complete the task set $\{1, 2, 3, 4, 5, 6, 7\}$, capacity limitations necessitate its execution across two optimized trips: $\{1, 2, 5, 4\}$ and $\{3, 6, 7\}$.} The objective is to allocate task points to robots in a way that ensures all tasks are completed within a limited time while minimizing the robots' total energy consumption.

For generality, several assumptions are made:
\begin{itemize}
    \item All robots are homogeneous, fully functional, and do not experience breakdowns while in motion.
    \item Each task point is assigned to one robot.
    \item Once a robot reaches full capacity, it must return to the depot to unload.
    \item The energy consumption of each robot is proportional to its weight and load~\cite{dorling2016vehicle}.
    \item All the trees are of the same type, but their yields vary.
    \item Variations in task demand over short periods are ignored. 
    \item When the total task volume remains constant, the energy required to pick all the fruit is assumed to be a fixed value. This assumption impacts only the simulation results and not the algorithm design. Therefore, in calculating total energy consumption, only the energy expended by the robots during their travel is considered.
\end{itemize}

\subsection{Problem modeling}
Based on the above description, a mathematical model for MTPRTS is proposed, which is formulated as a MILP model (the solution representation for heuristic approach is presented in subsection~\ref{Solution representation}). Given a set of tasks $N = \{1,\ldots, n\}$ and the depot $\{0\}$, the model is constructed as follows:

\vspace{2ex}
\textbf{Decision Variables}
\begin{flalign*}
&x_{ij}  = \begin{cases}
1, & \text{if robot travels directly from node } i \text{ to node } j \\
0, & \text{otherwise}
\end{cases} &\\
& \qquad \forall i,j \in N \cup \{0\}, i \neq j &
\end{flalign*}
\begin{flalign*}
&y_i  = \begin{cases}
1, & \text{if task point } i \text{ is the first visit after depot} \\
0, & \text{otherwise}
\end{cases} &\\
& \qquad \forall i \in N &\\[2ex]
&b_i  = \begin{cases}
1, & \text{if robot returns to depot from task point } i \\
0, & \text{otherwise}
\end{cases} &\\
& \qquad \forall i \in N &\\[2ex]
&L_i  \geq 0: \text{ load of robot when leaving node } i &\\
& \qquad \forall i \in N \cup \{0\} &\\[1ex]
&u_i  \geq 0: \text{ auxiliary variable for subtour elimination} &\\
& \qquad \forall i \in N &\\
&z_{kr}  = \begin{cases}
1, & \text{if robot } k \text{ executes trip } r \\
0, & \text{otherwise}
\end{cases} &\\
&E_r: \text{ time-consuming on trip } r &\\
&E_{\text{max}}: \text{ makespan constraint of a problem}&
\end{flalign*}

\textbf{Parameters}
\begin{flalign*}
& d_{ij}: \text{ distance between nodes } i \text{ and } j && \\
& q_i:  \text{ yield of task } i && \\
& Q: \text{ robot capacity} && \\
& W: \text{ robot weight} && \\
& R: \text{ set of all possible trips} && \\
& K: \text{ set of all robots} &&
\end{flalign*}
\vspace{-1em}

\textbf{Objective Function}

\begin{flalign}
\label{eq1}
& \min Z =\begin{aligned}[t]
& \sum_{i=1}^n \sum_{j \neq i, j \neq 0} d_{ij}(W + L_i)x_{ij} + \\
& \sum_{j=1}^n d_{0j}Wx_{0j} + \sum_{i=1}^n d_{i0}(W + L_i)b_i 
\end{aligned}&&
\end{flalign}

\textbf{Constraints}
\begin{flalign}
\label{eq2}
& \sum_{j \in N \setminus \{i\}} x_{ij} + b_i = 1 \quad \forall i \in N && \\[1.8ex] \label{eq3}
& \sum_{i \in N \setminus \{j\}} x_{ij} = 1 \quad \forall j \in N && \\[1.8ex] \label{eq4}
& y_j = x_{0j} \quad \forall j \in N && \\[1.8ex] \label{eq5}
& L_0 = 0 && \\[1ex]\label{eq6}
& L_j = q_j y_j + \sum_{i=1, i \neq j}^n (L_i + q_j)x_{ij} \quad \forall j \in N && \\[1ex]
\label{eq7}
& L_i \leq Q \quad \forall i \in N && \\[1.8ex]\label{eq8}
& u_i - u_j + nx_{ij} \leq n-1 \quad \forall i,j \in N, i \neq j && \\[1.8ex]\label{eq9}
& \sum_{i=1}^n b_i = \sum_{j=1}^n x_{0j} && \\ \label{eq10}
& x_{ij} \in \{0,1\} \quad \forall i,j \in N \cup \{0\}, i \neq j && 
\end{flalign}
\begin{flalign}
\label{eq11}
& y_i \in \{0,1\} \quad \forall i \in N && \\[1.8ex]
\label{eq12}
& b_i \in \{0,1\} \quad \forall i \in N && \\[1.8ex]
\label{eq13}
& L_i \geq 0 \quad \forall i \in N \cup \{0\} && \\[1.8ex]\label{eq14}
& u_i \geq 0 \quad \forall i \in N &&\\[1.8ex]\label{eq15}
& \sum_{k} z_{kr} = 1, \quad \forall r \in R && \\[1.8ex]\label{eq16}
& \sum_{r} E_r z_{kr} \leq E_{\text{max}}, \quad \forall k \in K && \\[1.8ex]\label{eq17}
& E_r = \begin{aligned}[t]
& \!\sum\limits_{(i,j) \in r} d_{ij}(W + L_i)x_{ij} \,+\, d_{0j}W \\
& + d_{i0}(W + L_i), \quad \forall r \in R
\end{aligned} &&
\end{flalign}
\vspace{-0.1em}
\textbf{Where:}
\begin{itemize}
\item (1): Minimization of transportation costs along the trip depends on the robot's weight and its load
\item (2): Each robot must either leave for another task point or return to depot
\item (3): Each task point must be visited exactly once
\item (4): Identifies first task points after depot visits
\item (5): Each robot starts from the depot with no load
\item (6): Load propagation along the trip
\item (7): Robot capacity constraint
\item (8): MTZ subtour elimination constraints
\item (9): Number of trips starting from depot equals number of returns
\item (10-14): Domain constraints
\item (15): Each trip must be visited exactly once
\item (16): Makespan constraint
\item (17): Energy calculation of a trip based on trip traversal and load
\end{itemize}

\section{Methodology}
\label{Problem solving methodology}

This section presents the solution representation methodology at the beginning. Then three key components for the route generation (RG) phase are introduced: the integrated load-distance balancing initialization method, the clustering-based local search mechanism, and the experience-based adaptive selection strategy. To facilitate understanding, a description of the solution update and evaluation methods is interspersed within these subsections. For RS phase, an infeasibility repair procedure along with three repair frameworks to better satisfy makespan constraints is detailed. After that, we outline the overall process of the algorithm. Finally, the computational complexity of AEDGA is presented.

\subsection{Solution representation}
\label{Solution representation}
\begin{figure}[htp]
    \centering
    \includegraphics[width=8.5cm]{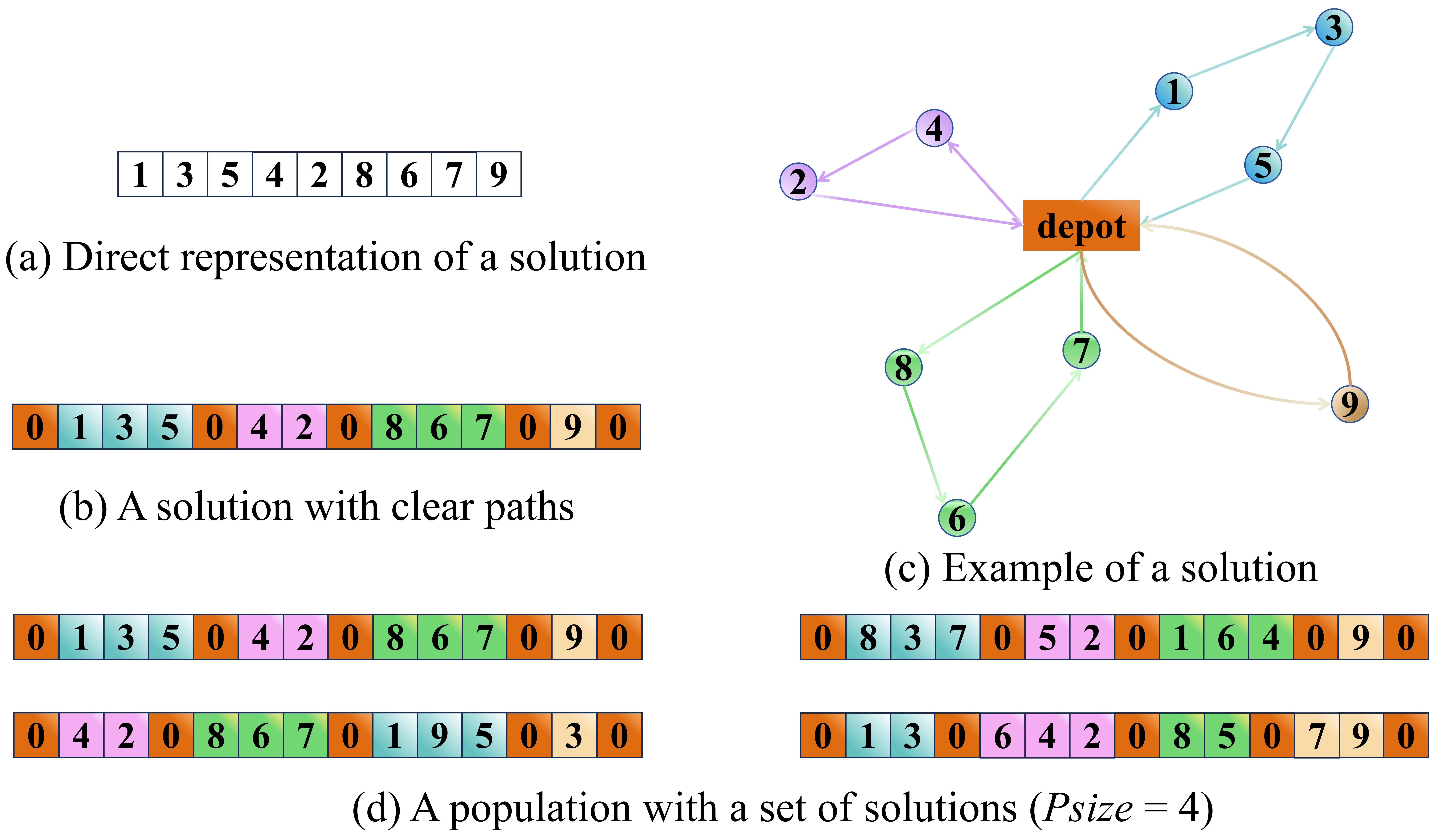}
    \caption{Solution representation}
    \label{solution representation}
\end{figure}

In this problem, each fruit tree is assigned a unique identifier. To maintain simplicity, the sequence of these identifiers represents a solution. In a solution, each fruit tree appears exactly once, as shown in Fig.~\ref{solution representation}(a). This sequence indicates the picking order of each fruit tree. Due to load constraints, all fruit trees may need to be harvested either through the deployment of multiple robots or through multiple trips executed by the same robot. To differentiate trips, a '0' symbol is inserted within the sequence to represent the depot, as illustrated in Fig.~\ref{solution representation}(b). This structure ensures that each robot departs from the depot, completes a series of picking tasks, and eventually returns to the depot, as depicted in Fig.~\ref{solution representation}(c). During solution evaluation, the total demand of trees between any two adjacent $'0'$ elements in the sequence must not exceed the robot’s load limit. Otherwise, the solution is considered infeasible and lacks practical utility.

The proposed AEDGA is a population-based optimization approach, where the algorithm operates on a population with a set of solutions and utilizes information exchange between individuals to guide the search process. Each solution corresponds to an individual in the population, with the number of solutions representing the population size ($\textcolor{blue}{\mathcal{P}}$), as shown in Fig.~\ref{solution representation}(d). Different solutions represent different servicing sequences for the fruit trees, with the inclusion of $'0'$ elements adding further diversity to the solutions. This representation is not only simple and intuitive but also effectively reflects the problem-solving approach, providing a solid foundation for subsequent algorithmic operations.

\subsection{Integrated load-distance balancing initialization method}

A high-quality and diverse initial population accelerates algorithm convergence and leads to better results. To generate a complete feasible solution, a sequence of $n$ tasks must be created with the appropriate placement of $'0'$ elements.

\begin{figure*}[htp]
    \centering
    \includegraphics[width=18cm]{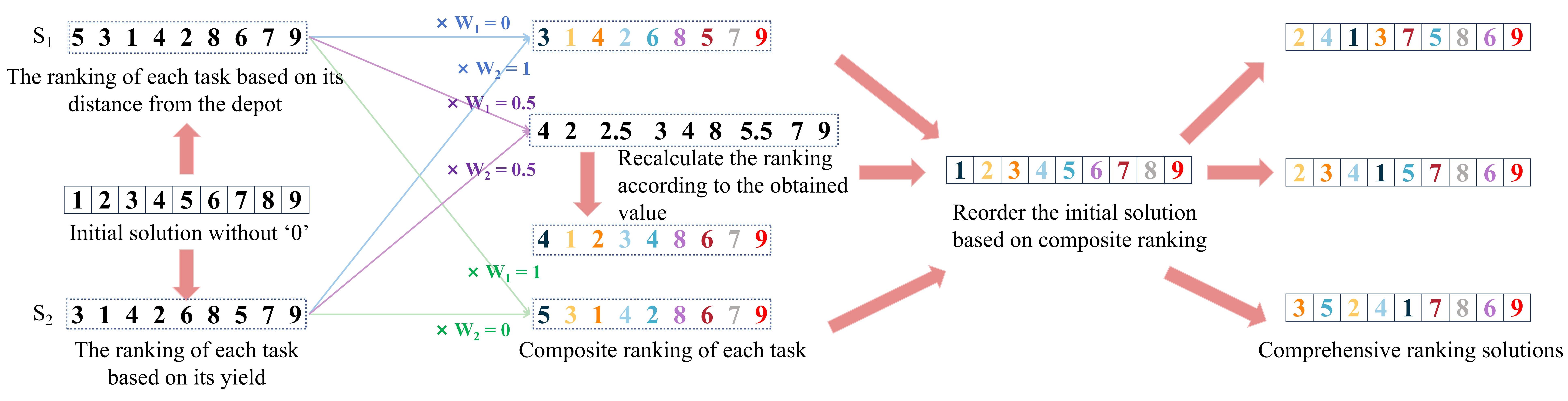}
    \caption{An example for obtaining comprehensive ranking solution}
    \label{fig.ILBIM}
\end{figure*}

To establish a promising initial population, an integrated load-distance balancing initialization method (ILBIM) is designed. First, an initial solution is created with $n$ tasks using direct solution representation. Then tasks in the solution are sorted by distance, ensuring that points farther from the depot are ranked higher, forming the ranking sequence $S_1$. Similarly, the tasks are sorted by yield per tree, with points requiring smaller loads ranked higher, producing ranking sequence $S_2$. Consequently, the optimal strategy prioritizes harvesting fruit trees that are situated at greater distances from the depot but carry smaller loads, while deferring the collection of proximate trees with larger yields to later stages. The ranking of each task point in $S_1$ and $S_2$ is then weighted and combined to yield a composite ranking. Based on this, the initial solution is reordered to obtain a set of solutions with comprehensive ranking ($\textcolor{blue}{\Pi}$), which takes both task distance and load into account: 

The weight is configured as an arithmetic sequence with a difference of 1/($\textcolor{blue}{\mathcal{P}}$-1), applied across each solution to diversify the initial population. For example, weight $W_1$ for $S_1$ is set as \{$0, 1/(\textcolor{blue}{\mathcal{P}}-1), ..., 1$\}, and weight $W_2$ for $S_2$ is set as \{$1, 1-1/(\textcolor{blue}{\mathcal{P}}-1), ..., 0$\}, based on their complementary relationship. Each solution’s weights are applied throughout the initialization operations. Finally, the composite ranking for each task point is determined by ranking in $S_1$ $\times W_1$ plus ranking in $S_2$ $\times W_2$. The tasks in the initial solution are sorted in ascending order of their composite ranking to produce the $\textcolor{blue}{\Pi}$. Fig.~\ref{fig.ILBIM} provides an example with $n=9$ and $\textcolor{blue}{\mathcal{P}}=3$.

The solution construction process is employed to rearrange the solution sequences and insert depot visits $'0'$ at feasible positions, which is conducted based on the obtained $\textcolor{blue}{\Pi}$. For each solution, starting from the depot with an empty robot, the highest-ranked task point $i$ in the sequence is chosen first. After completing the picking task at this point, it contributes to the load $L_i$. The remaining task loads are then evaluated based on their share of the robot's available capacity: 
\begin{equation}
    Pr_j=L_j/(1-L_i) \quad (\, j=1,...,n, \enspace and \enspace j \neq i \,)
    \label{eqpj}
\end{equation}
Any task with $Pr>1$ is ignored for the next step to prevent overload. For tasks with $Pr\leq1$, tasks are ranked in two ways: ranking tasks by proximity to $i$ to obtain $S_1'$ and by load ratio in descending order to obtain $S_2'$. Similarly, each solution generates a comprehensive sequence $S_3'$ using respective weights, and gets the top-ranked task in $S_3'$ (point $k$). If $d_{0i} \geq d_{ik}$, the robot proceeds to $k$. Otherwise, it returns to the depot. This process repeats until the load constraint or proximity condition necessitates returning to the depot. When the robot arrives at the depot again, a trip is completed, and all executed tasks along the current trip are removed from the $\textcolor{blue}{\Pi}$. Then the robot restarts from the depot with an empty load, repeating the process until all tasks in the $\textcolor{blue}{\Pi}$ are assigned to trips. 

\begin{figure}[htp]
    \centering
    \includegraphics[width=9cm]{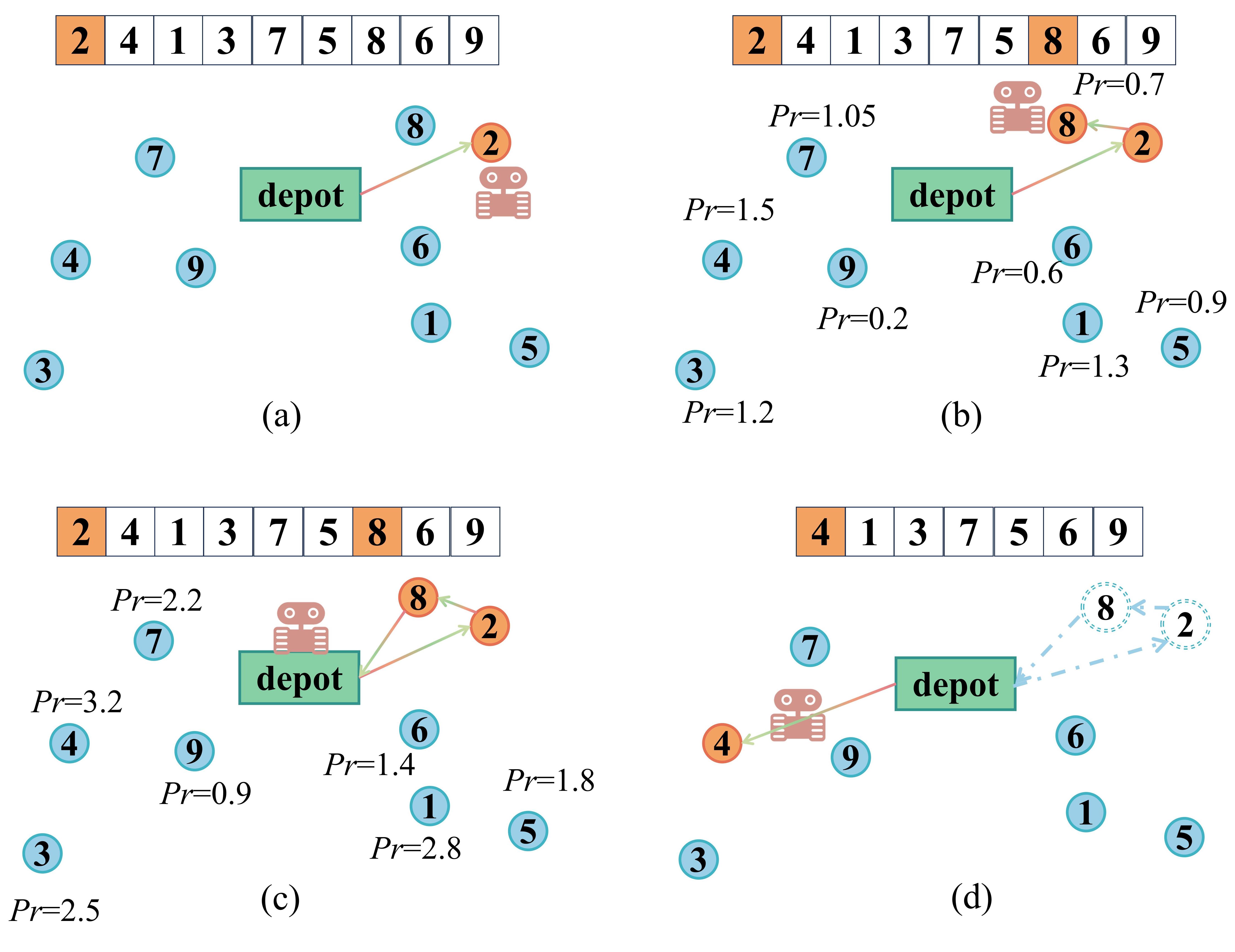}
    \caption{An example for trip construction}
    \label{fig.solution construction}
\end{figure}

The completed trip construction is demonstrated in Fig.~\ref{fig.solution construction} with example weights ($W_1=W_2=0.5$), where the sequence represents a $\textcolor{blue}{\Pi}$. The first target point for the robot is task point $2$, as seen in Fig.~\ref{fig.solution construction}(a). After completing the task, the remaining task points are \{$8, 6, 5, 9$\} sorted by proximity to task $2$ with $Pr\leq1$. Obviously, task point $8$ is suitable for the next task, as illustrated in Fig.~\ref{fig.solution construction}(b). After that, only task point $9$ remains feasible, as shown in Fig.~\ref{fig.solution construction}(c). Due to the greater distance between points $8$ and $9$ than between task point $8$ and the depot, the robot returns to the depot. A trip is thereby formed. Subsequently, update the $\textcolor{blue}{\Pi}$ and repeat the trip construction, as demonstrated in Fig.~\ref{fig.solution construction}(d), until all tasks are executed.

The algorithmic flow of ILBIM is presented in Algorithm~\ref{alg:main_initialization} and Algorithm~\ref{alg:construct_solution}.

\begin{algorithm}[htb]
\caption{Integrated load-distance balancing initialization method}
\label{alg:main_initialization}
\SetAlgoLined
\KwIn{\parbox[t]{0.8\linewidth}{%
 \hspace*{0.37em}    Set of nodes: $N = \{1,2,\ldots,n\} \cup \{0\}$ \\
  \hspace*{0.37em}   Population size: $\textcolor{blue}{\mathcal{P}}$\\
  \hspace*{0.37em}   Distance matrix: $d_{ij}$\\
  \hspace*{0.37em}   Yield of each task: $q$\\
  \hspace*{0.37em}   Robot capacity: $Q$\\
   \hspace*{0.37em}  Robot weight: $W$
}}
\KwOut{Initial population}
\BlankLine
$S_1 \leftarrow$ Sort tasks by $d_{0i}$ (descending)\\
\nonl $S_2 \leftarrow$ Sort tasks by $q$ (ascending)\\
\nonl $rank_1 \leftarrow S_1$ \tcp*{The ranking of each task is obtained according to $S_1$}
\nonl $rank_2 \leftarrow S_2$\\
\nonl $Population \leftarrow \emptyset$\\
\For{$j \leftarrow 1$ \KwTo $\textcolor{blue}{\mathcal{P}}$}{
    \nonl $\beta_1 \leftarrow (p-1)/(\textcolor{blue}{\mathcal{P}}-1)$\\
    \nonl $\beta_2 \leftarrow 1 - \beta_1$\\
    \For{each task $i$ in $N$}{
        \nonl $Rank_i \leftarrow \beta_1 \times Rank_1(i) + \beta_2 \times Rank_2(i)$\\
    }
    \nonl $\textcolor{blue}{\Pi} \leftarrow$ Sort tasks by $Rank_i$ (ascending)\\
    \nonl Obtain an $individual$ with $\textcolor{blue}{\Pi}$ \tcp*{Algorithm \ref{alg:construct_solution}}
    \nonl Add $individual$ to $Population$\\
}
\BlankLine
\KwRet{$Population$}
\end{algorithm}

\begin{algorithm}[htb]
\caption{Construction of a solution}
\label{alg:construct_solution}
\SetAlgoLined
\KwIn{\parbox[t]{0.8\linewidth}{%
\hspace*{0.37em} A comprehensive ranking solution: $\textcolor{blue}{\Pi}$:\\ 
\hspace*{0.37em} Other parameters same as Algorithm \ref{alg:main_initialization}}}
\KwOut{A feasible solution}
\BlankLine
 $Trips \leftarrow \emptyset$\\
\nonl $CR \leftarrow \emptyset$ \tcp*{Initialization current trip}
\nonl $L_0 \leftarrow 0$\\
\nonl $CL \leftarrow 0$ \tcp*{Initialization current load}
\While{$\textcolor{blue}{\Pi}$ is not empty}{
    \nonl $i \leftarrow$ First task in $\textcolor{blue}{\Pi}$\\
    \If{$CL + q_i \leq Q$}{
        \nonl $CL \leftarrow CL + q_i$\\
        \nonl $L_i \leftarrow CL$\\
        \nonl $CR \leftarrow CR \cup \{i\}$\\
        \nonl Remove $i$ from $\textcolor{blue}{\Pi}$\\
        \While{$\textcolor{blue}{\Pi}$ is not empty}{
            \nonl $FT \leftarrow \emptyset$ \tcp*{Feasible tasks}
            \For{each task $j$ in $\textcolor{blue}{\Pi}$}{
                \If{$CL + q_j \leq Q$}{
                    \nonl Add $j$ to $FT$\\
                }
            }
            \If{$FT = \emptyset$}{
                \nonl \textbf{break}\\
            }
            \nonl Obtain $Pr$ via Eq.~(\ref{eqpj}) \tcp*{Calculate occupancy rate}
            \nonl Sort $FT$ by $d_{ij}$ to get $S_1'$\\
            \nonl Sort $FT$ by $P$ to get $S_2'$\\
            \nonl Calculate weighted ranks for tasks in $FT$\\
            \nonl $k \leftarrow$ Task with minimum weighted rank\\
            \If{$d_{i0} < d_{ik}$}{
                \nonl \textbf{break}\\
            }
            \nonl $CR \leftarrow CR \cup \{k\}$\\
            \nonl $CL \leftarrow CL + q_k$\\
            \nonl $L_k \leftarrow CL$\\
            \nonl Remove $k$ from $\textcolor{blue}{\Pi}$\\
            \nonl $i \leftarrow k$\\
        }
        \nonl $CR \leftarrow \{0\} \cup CR \cup \{0\}$\\
        \nonl $Trips \leftarrow Trips \cup \{CR\}$\\
        \nonl $CR \leftarrow \emptyset$\\
        \nonl $CL \leftarrow 0$\\
    }
}
\BlankLine
\KwRet{$\text{The solution constructed by}\ Trips$}
\end{algorithm}

\subsection{Solution update and evaluation}
During each iteration of the AEDGA algorithm, an offspring population $P'$ is generated based on the parent population $P$. The two populations are then combined and sorted in ascending order of objective values. Using an elitist selection strategy, the top $\textcolor{blue}{\mathcal{P}}$ individuals are selected as the parent population for the next iteration.

The solution evaluation is conducted based on sequences formed by two consecutive $'0'$ elements and all elements between them. However, during the population update process, infeasible solutions may arise, where a robot's load exceeds its maximum limit in a trip due to random algorithmic operations. In such cases, it is assumed that upon approaching the load limit, the robot will return to the depot for unloading and immediately proceed with its journey to accomplish the remaining tasks along the current trip. Once remaining tasks in the current trip are completed, the robot must return to the depot, regardless of any unused load capacity, as a penalty for the infeasible solution.

\subsection{Clustering-based local search mechanism}
Traditional GAs rely on global search, which offers strong exploration capabilities~\cite{bae2007integrated,liu2023improved}. However, they still face: 1) insufficient exploration of promising solutions in the early search phase, resulting in overlooking numerous potential high-quality solutions; and 2) difficulty in deeply refining quality solutions in the later phase, which may lead to the algorithm becoming trapped in local optima, missing potentially superior solution spaces~\cite{katoch2021review}. Particularly in large and uneven solution spaces, the randomness of global search may steer population evolution in the wrong direction, reducing optimization efficiency. Therefore, we introduce a clustering-based local search mechanism (CLSM) to target high-potential regions, accelerating convergence by focusing on deep searches for specific solutions and enhancing both solution quality and diversity.

In the context of the specific problem addressed in this study, CLSM is employed to adjust the trip structure of the most cost-effective trips among promising solutions. This local search mechanism adopts a single-solution search strategy, optimizing only one selected solution at a time. By integrating this mechanism with GA, population is optimized more effectively, reducing overall energy consumption and improving task execution efficiency.

For each selected solution, we assess the structural rationality of each trip. Specifically, for each trip, we use the K-means clustering method to divide the task nodes along the trip into two clusters (excluding the depot position and any trips with only one task point). Dividing the nodes into two clusters allows us to efficiently and intuitively analyze the spatial distribution of task points on the current trip. We then calculate the centroids of the two clusters, denoted as $C_1$ and $C_2$, and evaluate the distance between them:
\begin{equation}
D(r) = \| C_1 - C_2 \|
\label{eq19}
\end{equation}
When a significant distance exists between the class centroids, there is a greater likelihood that the corresponding trip structure contains opportunities for improvement. Consequently, these trips are given precedence in the local optimization process. This strategic approach enables the local search procedure to concentrate specifically on trips exhibiting potential structural inefficiencies, thus avoiding the computational burden of indiscriminate trip adjustments while simultaneously improving both algorithmic efficiency and solution precision.

Subsequent to the target trip selection, an assessment is conducted to determine the optimal direction of adjustment between the two clusters. Let clusters $A$ and $B$ denote the respective sets of task points contained within the target trip. The distances between the cluster centroids and the depot location $D_0$ are computed as follows:
\begin{equation}
D_i = \| C_i - D_0 \| \quad (\,i=1,2\,)
\label{eq20}
\end{equation}
and use the equation
\begin{equation}
B = \arg \max (D_i)
\label{eq21}
\end{equation}
to determine the cluster farther from the depot (e.g., cluster $B$). The algorithm proceeds to compute the centroids of all remaining trips and identifies the trip whose centroid $C_c$ demonstrates the shortest distance to the centroid $C_2$ of cluster $B$. From the trip set $\{ R \}$, the most suitable candidate trip $R_c$ for recombination with $B$ is determined by minimizing the distance to $C_2$:
\begin{equation}
R_c = \arg \min ( \| C_2 - C_c \| )
\end{equation}
Then the redistribution of task points between the two trips is facilitated through a discrete crossover operation. This operation reconstructs task sequences within both trips, aiming to achieve load balance while optimizing the overall trip structure.

Following the discrete crossover operation, the revised customer assignments may generate two new trips. To enhance these recombined trips, we employ an established traveling salesman problem (TSP) methodology~\cite{stutzle1999aco} to optimize each newly formed trip. The TSP optimization procedure restructures the customer visitation sequence within each trip, seeking to achieve near-optimal trips while simultaneously reducing total energy consumption and return times.

\begin{figure}[htp]
    \centering
    \includegraphics[width=9cm]{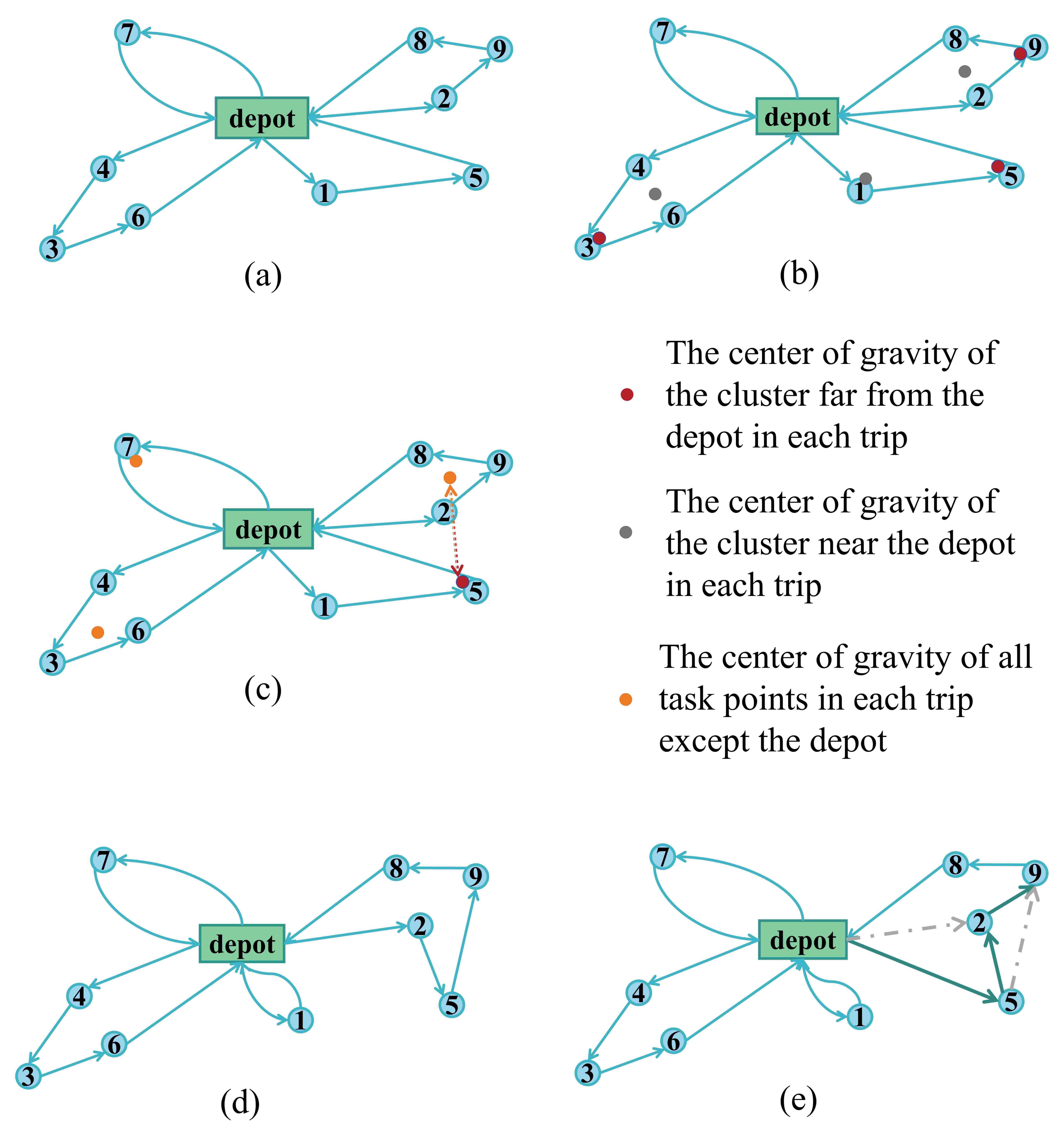}
    \caption{An example of CLSM}
    \label{An example of CLSM}
\end{figure}

Fig.~\ref{An example of CLSM} is an example that illustrates the working process of CLSM. Fig.~\ref{An example of CLSM}(a) presents the selected solution. Subsequently, for each trip containing more than one task point, the constituent points are partitioned into two distinct clusters. Within each trip, the centroid more distant from the depot is denoted by a red point, while the proximate centroid is represented by a gray point, as illustrated in Fig.~\ref{An example of CLSM}(b). The trip \{$0,1,5,0$\} exhibits the maximum distance between two cluster centroids. Subsequently, we compute the centroids of task nodes (excluding the depot) for all other trips, and identify the one nearest to the red centroid of the current trip, as demonstrated in Fig.~\ref{An example of CLSM}(c). A crossover operation is then performed between the task points of these two trips, generating a new trip configuration, as depicted in Fig.~\ref{An example of CLSM}(d). Finally, the internal structure of the newly formed trip undergoes further optimization, as shown in Fig.~\ref{An example of CLSM}(e).

The CLSM establishes a more systematic approach to solution optimization, simultaneously addressing the load balance and structural efficiency of individual trips while leveraging information from neighboring trips. The solution quality is further refined through the integration of discrete crossover operations and TSP-based optimization procedures. The comprehensive methodology of this mechanism is formally delineated in Algorithm~\ref{CLS}.

\begin{algorithm}[htb]
\caption{Clustering-based local search mechanism}
\label{CLS}
\SetAlgoLined
\KwIn{\parbox[t]{0.8\linewidth}{%
\hspace*{0.37em} Current solution to be optimized, $Solution$\\
\hspace*{0.37em} Depot location, $D_0$\\
\hspace*{0.37em} Set of all trips in solution, $R$}}
\KwOut{Improved solution}
\BlankLine
$TR \gets \emptyset$ \tcp*{Target trip}
\nonl $MaxDistance \gets 0$\\
\For{each trip $r \in R$}{
    \If{number of tasks in $r > 1$ \tcp*{excluding depot}}{
        \nonl Obtain \{$C_1$,$C_2$\} with $r$ \tcp*{Get two clusters $A$ and $B$ with centroids $C_1$ and $C_2$ by K-means clustering}
        \nonl Obtain $D(r)$ via Eq.~(\ref{eq19})\\
        \If{$D(r) > MaxDistance$}{
            \nonl $MaxDistance \gets D(r)$\\
            \nonl $TR \gets r$\\
        }
    }
}
\If{$TR \neq \emptyset$}{
    \nonl Obtain \{$D_1$,$D_2$\} via Eq.~(\ref{eq20})\\
    \eIf{$D_1 > D_2$}{
        \nonl $FC \gets C_1$\\
    }{
        \nonl $FC \gets C_2$\\
    }
    \nonl $MinDist \gets \infty$\\
    \nonl $CaRo \gets \emptyset$ \tcp*{Candidate trip}
    \For{each trip $r \in R \setminus \{TR\}$}{
        \nonl Calculate centroid $C_r$ of trip $r$\\
        \If{$||FC - C_r|| < MinDist$}{
            \nonl $MinDist \gets ||FC - C_r||$\\
            \nonl Obtain $CaRo$ with $r$\\
        }
    }
    \nonl \{$NR_1, NR_2$\} $\gets$ \textbf{Crossover}($TR, CaRo$) \tcp*{New route generation}
    \nonl \{$NR_1, NR_2$\} $\gets$ \textbf{Optimize}($NR_1, NR_2$) \tcp*{Optimize new trips with ACO}
    \nonl Update solution by replacing old trips with optimized new trips\\
}
\BlankLine
\KwRet{Improved solution}
\end{algorithm}

\subsection{Experience-based adaptive selection strategy}

In the initial generation of AEDGA, an elitist strategy is employed, wherein the individual exhibiting the minimum fitness value from the current population is selected for the clustering-based local search mechanism. However, persistent application of this approach across subsequent generations may lead to premature convergence to local optima, thereby constraining further improvements in solution quality. To enhance search diversity and mitigate premature convergence, an experience-based adaptive selection strategy (EASS) is introduced. This methodology dynamically modulates both the selection range and probability of local search targets throughout the evolutionary process, facilitating exploration across multiple promising individuals. The strategy employs a historical count sequence to monitor and update individual performance, which subsequently determines selection weights. When integrated with Lehmer mean~\cite{gao2019chaotic} based adaptive selection, this approach effectively reduces the likelihood of convergence to local optima.

Starting from the second generation of algorithm search, AEDGA dynamically selects solutions for further optimization. Specifically, the process involves generating a fixed uniform sequence ranging from 0.1 to \( p \), where \( p = 0.6 \) as an example: \(\{0.1, 0.2, 0.3, 0.4, 0.5, 0.6\}\). This sequence, designated as the range sequence, corresponds to the top \(\{10\%, 20\%, 30\%, 40\%, 50\%, 60\%\}\) of individuals in the population based on fitness ranking. During each generation, the algorithm stochastically samples a value from the range sequence. For example, if $0.2$ is chosen, a random individual within the top $20\%$ of the current population will be selected as the local search target.

To track and optimize the effectiveness of each proportion value, the algorithm maintains an external count sequence, \texttt{archive = [0, 0, 0, 0, 0, 0]}, which corresponds to the values in the range sequence. Each time a proportion value is selected for local search, if the generated individual outperforms the best individual in the current population, the corresponding count in the sequence is incremented. For example, if $0.2$ was chosen and resulted in an improved individual, the count sequence updates to \texttt{archive = [0, 1, 0, 0, 0, 0]}. Otherwise, the count remains unchanged. As evolution progresses, this count sequence accumulates each proportion’s historical performance, reflecting the contribution of each value in optimization. In this way, the algorithm adaptively increases the selection weight of well-performing values, improving the efficiency of choosing individuals that are more likely to enhance solution quality.

For adjusting selection probabilities, the algorithm applies the Lehmer mean to compute the selection probability of each proportion value. Initially, the algorithm derives the selection weights, \texttt{archiveF}, based on the count sequence \texttt{archive}, defined as:
\begin{equation}
archiveF = \frac{archive}{\sum(archive)}
\label{eq23}
\end{equation}
The selection probability for each proportion value is subsequently derived through the application of the Lehmer mean. The mathematical formulation for the selection probability is expressed as:
\begin{equation}
    F=(1-c) \times archiveF + c \times \frac{\sum(archiveF^2)}{\sum(archiveF)}
    \label{eq24}
\end{equation}
where \( c \) is a constant used to balance selection probability uniformity with the weight of historical performance, which is set to $\frac{1}{\textcolor{blue}{\mathcal{P}}}$ in this research. This Lehmer mean-based adaptive adjustment strategy allows the algorithm to gradually favor effective proportion values in the long term, optimizing the individual selection strategy progressively.

The EASS significantly enhances algorithm diversity while improving local search quality. Through the integration of a count sequence, Lehmer mean calculations, and adaptive probability adjustment, the algorithm can dynamically select search targets for local search operations based on historical performance. The comprehensive methodology is illustrated in Algorithm~\ref{alg:EASS}.

\begin{algorithm}[htb]
\caption{Experience-based adaptive selection strategy}
\label{alg:EASS}
\SetAlgoLined
\KwIn{\parbox[t]{0.8\linewidth}{%
\hspace*{0.37em} Current population, $Pop$\\
\hspace*{0.37em} Maximum selection range, $p$\\
\hspace*{0.37em} Best solution, $bestInd$\\
\hspace*{0.37em} Population size, $\textcolor{blue}{\mathcal{P}}$\\
\hspace*{0.37em} Other parameters same as Algorithm \ref{CLS}}}
\KwOut{Best solution $bestInd$ obtained by local search and count sequence $archive$}
\BlankLine
\eIf{$gen = 1$}{
    \nonl $Ind \gets bestInd$\\
}{
    \nonl $ranges \gets [0, 0.1, ..., p]$\\
    \nonl $n \gets$ length of $ranges$\\
    \nonl $archive \gets [0,0,..., 0]$\\
    \nonl $archiveF \gets [1/n,1/n,..., 1/n]$\\
    \nonl $F \gets$ Eq.~(\ref{eq24})\\
    \nonl $\textcolor{blue}{\omega} \gets$ Randsanple($ranges$,$F$)\\
    \nonl $i \gets$ index of $\textcolor{blue}{\omega}$ in $ranges$\\
    \nonl $k \gets \textcolor{blue}{\omega} \times \textcolor{blue}{\mathcal{P}}$\\
    \nonl $ES \gets$ Top $k$ individuals from $Pop$\\
    \nonl $Ind \gets$ Random individual from $ES$\\
}
\nonl $newInd \gets$ CLSM($Ind$) \tcp*{Algorithm \ref{CLS}}
\nonl Obtain $fitness(newInd)$ via Eq.~(\ref{eq1}) \\
\If{$fitness(newInd) < fitness(bestInd)$}{
    \nonl $archive(i) \gets archive(i) + 1$\\
    \nonl Obtain $archiveF$ via Eq.~(\ref{eq23})\\
    \nonl $bestInd \gets newInd$\\
}
\BlankLine
\KwRet{$bestInd, archive$}
\end{algorithm}

\subsection{Route scheduling framework}
\label{rs framework}
In the process of solving the MTPRTS, although the RG phase generates energy-optimized trips, these must be systematically distributed across multiple robots to accommodate the inherent makespan~\cite{altekin2022multi} constraints of the practical implementation. The fundamental objective in this stage is to effectively allocate the RG-generated trips to each robot subject to the constraint that each robot's cumulative execution time remains within the maximum allowable time. Given the relatively small scale of the model, determined by the number of robots and routes per solution, route scheduling is implemented directly using MILP based on Eqs.~(\ref{eq15}) to~(\ref{eq17}) to perform makespan evaluation. While solely applying traditional bin-packing methods may lead to infeasible solutions, an infeasibility repair procedure is proposed to adjust trips and task allocations. This strategy ensures that the assignment plan meets constraint requirements while maintaining efficiency.

In the infeasibility repair procedure, trips are initially sorted in descending order of energy consumption, prioritizing trips with higher consumption. Given a trip \( A \) requiring optimization and an initially empty trip \( B \). The repair process involves incrementally moving task points from trip \( A \) to trip \( B \) to reduce total energy consumption. In each operation, the last task from trip \( A \) (excluding the depot) is removed and added to the beginning of trip \( B \) (also excluding the depot). The combined total energy consumption \( Z \) of the two trips starts at infinity, with updates computed after each transfer operation. If the updated sum falls below \( Z \), the value of \( Z \) is revised and the transfer process continues with the subsequent task. The adjustment process terminates when no further reduction in total energy consumption is observed. After this operation, the makespan constraint is re-evaluated; if satisfied, the repair process concludes. Otherwise, the optimization proceeds to the next trip.

The experimental validation presented in subsection~\ref{results of RS} establishes the superior performance characteristics of the first method relative to the other frameworks under consideration. We therefore incorporate this methodology into the AEDGA, developing an integrated solution approach for MTPRTS optimization.

\begin{algorithm}[htbp]
\caption{Infeasibility repair procedure}
\label{alg:repair}
\SetAlgoLined
\KwIn{\parbox[t]{0.8\linewidth}{%
\hspace*{0.37em} Current solution to be optimized, $Solution$\\
\hspace*{0.37em} Set of routes generated from RG phase, $R$\\
\hspace*{0.37em} Makespan, $E_{max}$\\
\hspace*{0.37em} Number of robots, $m$\\
\hspace*{0.37em} Number of routes, $\Upsilon$}}
\KwOut{A solution with updated route set $R$ and its energy consumption $Z$}
\BlankLine
Sort $R$ in descending order by energy consumption (\{$Z_1$,$...$,$Z_\Upsilon$\})\\
\For{each route $A \in R$}{
    \nonl Create empty route $B$\\
    \nonl $Z_{com} \gets \infty$\\
    \nonl $improved \gets true$\\
    \While{$improved$}{
        \nonl $improved \gets false$\\
        \nonl $task \gets$ last task in $A$ (excluding depot)\\
        \nonl Remove $task$ from $A$\\
        \nonl Insert $task$ at beginning of $B$ (after depot)\\
        \nonl $E_{new} \gets Z_A+Z_B$\\
        \If{$E_{new} \leq Z_{com}$}{
            \nonl $Z_{com} \gets E_{new}$\\
            \nonl $improved \gets true$\\
        }
        \nonl Makespan evaluation via Eqs.~(\ref{eq15}) to~(\ref{eq17})\\
        \If{Makespan $\leq E_{max}$}{
            \nonl \textbf{break} \tcp*{Feasible solution found}
        }
    }
    \eIf{Makespan $\geq E_{max}$}{
        \eIf{Route not optimized exists}{
            \nonl Continue with next route optimization\\
        }{
            \nonl $Z = Inf$ \tcp*{Feasible solution not found}
        }
    }{
        \nonl $Z = \displaystyle\sum_{i=1}^{\Upsilon} Z_i$  \tcp*{Eq.~(\ref{eq1})}
    }
}
\BlankLine
\KwRet{$R$ and  $Z$}
\end{algorithm}

To guarantee adherence to the makespan constraint in the final allocation scheme, three distinct methodological frameworks are proposed for optimizing the assignment process:
\begin{itemize}

\item Generational solution processing and infeasibility remediation: Subsequent to each RG solution generation, the derived trip scheme is subjected to immediate allocation, followed by makespan evaluation. When solution infeasibility is detected, the previously delineated repair strategy is employed for transformation into a feasible solution. This systematic integration of allocation and repair procedures at the generational level facilitates continuous feasibility monitoring, thereby maintaining solution validity throughout the evolutionary trajectory.

\item Generational infeasibility elimination: In this methodological framework, subsequent to the completion of each generational RG solution, the generated trips are subjected to allocation procedures and makespan evaluation, followed by comprehensive feasibility assessment. Diverging from the previous methodology, this approach implements direct elimination of infeasible solutions rather than remediation, thereby maintaining exclusively feasible solutions and promoting natural algorithmic convergence toward the feasible solution domain.

\item Global solution integration with post-processing infeasibility remediation: This methodological framework postpones allocation procedures until the comprehensive generation of RG solutions is complete. Subsequently, a global makespan evaluation and feasibility assessment is executed across the solution set. The repair methodology is then uniformly implemented to transform infeasible solutions into feasible alternatives. The selection of the optimal solution is based on the identification of the feasible solution that minimizes total energy consumption. This framework demonstrates enhanced computational efficiency through the implementation of concentrated remediation operations to obtain a globally optimal feasible solution, thus circumventing the computational burden associated with generational repair processes.

\end{itemize}

\subsection{Complete flow of AEDGA}
The AEDGA initiates with population initialization through the ILBIM. After that, each solution undergoes objective value computation, makespan evaluation, and repair mechanisms for infeasible solutions. The algorithm then implements EASS  to identify promising solutions for optimization, followed by neighborhood exploration using CLSM. Traditional GA operations, including crossover and mutation, are subsequently applied to generate offspring solutions. The environmental selection process evaluates and ranks all individuals based on their fitness metrics, retaining the top $\textcolor{blue}{\mathcal{P}}$ solutions for the next generation. The systematic workflow of this optimization process is briefly outlined in Algorithm~\ref{alg:AEDGA}.

\begin{algorithm}[htbp]
\caption{Complete flow of AEDGA}
\label{alg:AEDGA}
\SetAlgoLined
\KwIn{\hspace*{0.37em} Problem parameters}
\KwOut{Best solution found}
\BlankLine
tic \tcp*{Start timing}
\nonl $It \gets 1$\\
\While{toc $\leq$ $Ft_{max}$}{
    \eIf{$It=1$}{
        \nonl Population initialization \tcp*{Algorithms~\ref{alg:main_initialization} and~\ref{alg:construct_solution}}
        \nonl Objective value calculation via Eq.~(\ref{eq1})\\
        \nonl Makespan evaluation via Eqs.~(\ref{eq15}) to~(\ref{eq17})\\
        \nonl Infeasible solution repair \tcp*{Algorithm~\ref{alg:repair}}
        \nonl $It \gets It + 1$\\
    }{
        \nonl Select a candidate solution for local search \tcp*{Algorithms~\ref{CLS} and~\ref{alg:EASS}}
        \nonl $P_{new} \gets GA(P)$\\
        \nonl $P_{new} \gets P_{new} \cup P$\\
        \nonl Objective value calculation via Eq.~(\ref{eq1})\\
        \nonl Makespan evaluation via Eqs.~(\ref{eq15}) to~(\ref{eq17})\\
        \nonl Infeasible solution repair \tcp*{Algorithm~\ref{alg:repair}}
        \nonl $P \gets P_{new}$ \tcp*{Environmental selection}
    }
}
\BlankLine
\KwRet{Best solution in $P$}
\end{algorithm}

\subsection{Algorithm complexity}

This section presents a concise analysis of AEDGA's computational complexity. The primary computational components encompass solution initialization, offspring generation, local search operations, and repair procedure.

For solution initialization, obtaining a comprehensive ranking solution incurs a sorting complexity of \textcolor{blue}{$O(n \cdot \log n)$} per solution. Given $\textcolor{blue}{\mathcal{P}}$ solutions, this component exhibits a computational complexity of \textcolor{blue}{$O(\mathcal{P} \cdot n \cdot \log n)$}. The trip construction process for each solution necessitates sorting of remaining task points for every task point, yielding an approximate complexity of \textcolor{blue}{$O(n^2 \cdot \log n)$}. Considering $\textcolor{blue}{\mathcal{P}}$ solutions, this phase's aggregate complexity becomes \textcolor{blue}{$O(\mathcal{P} \cdot n^2 \cdot \log n)$}. Consequently, the total computational complexity for solution initialization is \textcolor{blue}{$O(\mathcal{P} \cdot n^2 \cdot \log n)$}.

The offspring generation phase, primarily implementing the GA process, demonstrates an approximate complexity of \textcolor{blue}{$O(\mathcal{P} \cdot n)$}.

For the local search operation, the complexity per iteration is derived as follows: assuming an average of $f$ points per trip in a solution (where $f \leq n$) and $g$ iterations for $K$-means clustering, the $K$-means complexity approximates to \textcolor{blue}{$O(f \cdot g)$}. The implementation of ACO~\cite{stutzle1999aco} for trip optimization, based on validated parameters, utilizes $f$ iterations with an ant colony size of $\textcolor{blue}{\mathcal{P}}$, resulting in an approximate complexity of \textcolor{blue}{$O(\mathcal{P} \cdot f^3)$}. Given $I$ ($I \approx 0.2 \times \frac{n}{f}$) iterations of this process, the local search operation exhibits a complexity of \textcolor{blue}{$O(\mathcal{P} \cdot n \cdot f^2)$} $\leq$ \textcolor{blue}{$O(\mathcal{P} \cdot n^3)$}.

The repair procedure exhibits low computational complexity with an upper bound of \textcolor{blue}{$O(n)$}.

In summation, the algorithm demonstrates a maximal complexity of \textcolor{blue}{$O(\mathcal{P} \cdot n^3)$}. This relatively high computational complexity is primarily attributed to the utilization of ACO as the trip optimization methodology. Therefore, if necessary, using a simplified path optimization approach can significantly reduce the algorithm's complexity.

\section{Experimental results and analysis}
\label{section 4}
In this section, the experimental setup and comparative algorithms will be first introduced. Then we conduct sensitivity analysis and ablation experiments to evaluate key algorithmic components. Existing MTVRP methods typically place greater emphasis on performance comparisons in the first phase~\cite{sajid2022routing,he2023memetic}. Based on this common practice, we perform detailed comparative experiments between AEDGA and state-of-the-art algorithms in the route generation phase, followed by comprehensive result analysis. Finally, we validate the most suitable repair framework for AEDGA in the route scheduling phase.

\begin{table}[!hbpt]
\centering
\caption{Specifications for the experimental problems}
\label{Specifications for the experimental problems}
\setlength{\tabcolsep}{5.5pt}
\begin{tabular}{lllllll}
\hline
$Pro$    & $n$   & $\mu_d$ & $\lambda_d$ & $\mu_y$ & $\lambda_y$ & $Q$    \\
\hline
1       & 16  & 25.328    & 32.5576  & 15.375      & 31         & 35   \\
2       & 19  & 26.6635   & 43.9318  & 16.3158     & 31         & 160  \\
3       & 20  & 26.8869   & 43.9318  & 15.5        & 31         & 160  \\
4       & 21  & 26.0426   & 43.9318  & 14.1905     & 30         & 160  \\
5       & 22  & 26.0488   & 43.9318  & 14          & 30         & 160  \\
6       & 22  & 27.7502   & 49.366   & 1022.7273   & 2500       & 3000 \\
7       & 23  & 24.9664   & 43.9318  & 13.6087     & 30         & 40   \\
8       & 40  & 23.8694   & 43.9318  & 15.45       & 41         & 140  \\
9       & 45  & 24.8765   & 43.9318  & 15.3778     & 41         & 150  \\
10      & 50  & 22.7812   & 37.108   & 19.02       & 37         & 100  \\
11      & 50  & 22.7812   & 37.108   & 19.02       & 37         & 150  \\
12      & 50  & 22.7812   & 37.108   & 19.02       & 37         & 120  \\
13      & 51  & 24.0235   & 43.9318  & 15.2353     & 41         & 80   \\
14      & 55  & 22.6081   & 37.108   & 18.9455     & 37         & 115  \\
15      & 55  & 22.6081   & 37.108   & 18.9455     & 37         & 70   \\
16      & 55  & 22.6081   & 37.108   & 18.9455     & 37         & 170  \\
17      & 55  & 22.6081   & 37.108   & 18.9455     & 37         & 160  \\
18      & 60  & 23.4909   & 42.4264  & 18.9        & 37         & 120  \\
19      & 60  & 23.5088   & 42.4264  & 18.9        & 37         & 80   \\
20      & 65  & 24.254    & 43.2666  & 18.7538     & 37         & 130  \\
21      & 70  & 24.205    & 43.2666  & 18.7571     & 37         & 135  \\
22      & 76  & 24.2057   & 43.2666  & 17.9474     & 37         & 350  \\
23      & 76  & 24.2057   & 43.2666  & 17.9474     & 37         & 280  \\
24      & 101 & 24.9471   & 49.93    & 14.4356     & 41         & 400  \\
25      & 41  & 11.2492   & 19.8255  & 54.2439     & 70         & 300  \\
26      & 61  & 12.7403   & 21.4895  & 54.1475     & 70         & 300  \\
27      & 81  & 15.1656   & 26.7335  & 55.3951     & 70         & 300  \\
28      & 91  & 18.0054   & 32.7314  & 54.2308     & 70         & 300  \\
29      & 136 & 19.8554   & 37.4373  & 53.9632     & 70         & 300  \\
30      & 181 & 17.9296   & 33.4011  & 54.4088     & 70         & 300  \\
31      & 161 & 31.4785   & 53.4042  & 54.1056     & 70         & 300  \\
32      & 251 & 31.2362   & 54.85    & 54.1162     & 70         & 300  \\
33      & 321 & 24.9057   & 47.3523  & 54.7695     & 70         & 300  \\
34      & 251 & 29.4289   & 55.0552  & 54.9243     & 70         & 300  \\
35      & 376 & 34.84     & 63.5316  & 54.0904     & 70         & 300  \\
36      & 501 & 29.8008   & 57.0954  & 54.5349     & 70         & 300  \\
37      & 361 & 45.4521   & 79.2253  & 54.9086     & 70         & 300  \\
38      & 541 & 45.8196   & 83.4058  & 54.5915     & 70         & 300  \\
39      & 721 & 35.6037   & 68.5423  & 55.1956     & 70         & 300  \\
40      & 491 & 45.0968   & 86.1795  & 54.4094     & 70         & 300  \\
41      & 736 & 50.3167   & 93.4155  & 54.7079     & 70         & 300  \\
42      & 981 & 42.1328   & 79.7979  & 54.5352     & 70         & 300  \\
\hline
\end{tabular}
\end{table}

 \subsection{Experimental setup}
\label{Experimental setup}

To comprehensively evaluate the performance of AEDGA, we integrate $24$ established benchmark problems from dataset P~\cite{augerat2013vrp} with $18$ newly constructed test instances. The proposed instances vary in difficulty and include orchard sizes ranging from $20\times20$ to $70\times70$ square meters, with $\{100, 225, 400, 625, 900, 1225\}$ trees respectively. Each scenario also covers three maturity rates for the trees: $0.4, 0.6,$ and $0.8$. Thus, each combination of tree count and maturity rate represents a distinct test problem to evaluate the algorithm's task allocation performance under different work scenarios and task quantities. 

\begin{figure*}[!hbpt]
    \centering
    \includegraphics[width=18cm]{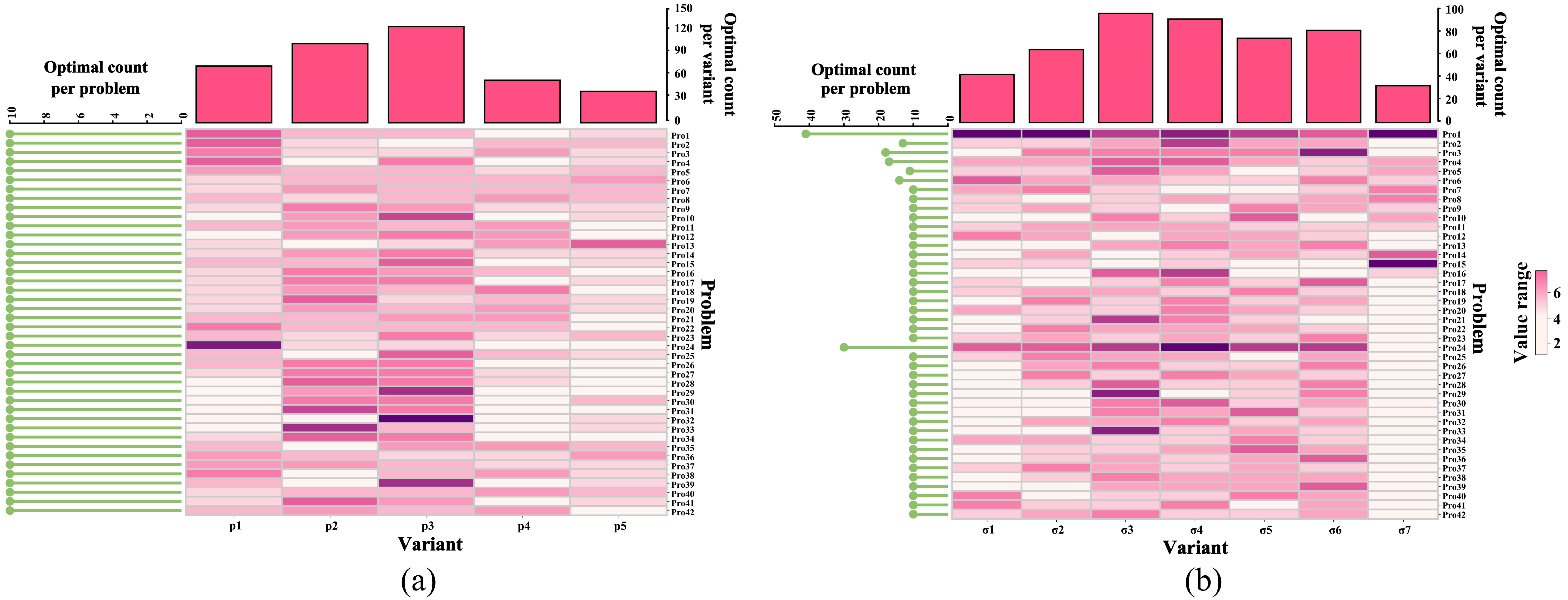}
    \caption{Test results of parameters}
    \label{Test results of parameters}
\end{figure*}

As previously noted, individual mature trees exhibit varying yields, with production levels randomly distributed between $40$ and $70$ units, while their spatial distribution is also randomly determined. Furthermore, for each test instance, the depot position is stochastically determined along the boundary edges of the orchard configuration. To ensure experimental consistency, the yield parameters and spatial coordinates of trees and depot are systematically documented for all $18$ generated test instances, facilitating algorithmic comparisons under identical experimental conditions. 

The specific form of the problems is shown in Table~\ref{Specifications for the experimental problems}. Due to space limitations, $Pro$ is used to denote $Problem$ in the first column. Problems $1-24$ are derived from the P test set, whereas the remaining instances are randomly generated for this research. The parameter $n$ represents the number of task points in each instance. $\mu_d$ and $\lambda_d$ denote the average and maximum distances from task points to the depot, respectively. Similarly, $\mu_y$ represents the average mature fruit tree yield, and $\lambda_y$ indicates the maximum yield across these trees. The parameter $Q$ defines the upper load capacity limit of the robots. In addition, the self-weight of the robot in each problem is set to $\frac{Q}{3}$. These diverse parameter configurations generate instances of varying complexity, facilitating a comprehensive evaluation of algorithmic robustness.

Following this, the proposed AEDGA algorithm is evaluated using design of experiments (DOE) and the Wilcoxon test, widely used methods for analyzing combinatorial optimization problems~\cite{xu2021bi} and comparing algorithm performance~\cite{qiao2022dynamic,song2023two}. For all heuristic algorithms, the termination criterion is established based on maximum CPU runtime, defined as $Ft_{max} = n$ seconds (s). The output population size is uniformly maintained at $\textcolor{blue}{\mathcal{P}} = 10$ across all experiments. To ensure statistical significance, $10$ independent runs are performed for these algorithms on each test instance. For the MILP approach~\cite{gurobi}, computational experiments are conducted under two different temporal bounds: $Ft_{max} = n \,$s and $Ft_{max} = 7200$s. All algorithmic implementations are executed on a computing platform equipped with an Intel Core i7-12700 CPU (2.1GHz) and 32GB of RAM.

\subsection{Comparative algorithms}

To evaluate the performance of AEDGA, comprehensive comparative experiments are conducted against \textcolor{blue}{eight} advanced algorithms, comprising \textcolor{blue}{seven} state-of-the-art heuristic methods and one MILP model proposed by this research. Specifically, GA-NN~\cite{de2018hybrid} integrates nearest neighbor heuristic within the GA framework, facilitating efficient local search through greedy mechanisms. ETSA~\cite{rabbouch2020empirical} extends the traditional SA paradigm by incorporating dynamic acceptance criteria, addressing the challenges of parameter tuning and convergence efficiency. GSACO~\cite{li2022ant} features adaptive greedy strategies and dynamic parameter adjustment mechanisms, demonstrating notable advantages in maintaining population diversity and accelerating convergence. HGSA~\cite{sajid2022routing} synthesizes GA's random crossover exploration with SA's local search exploitation, striving to achieve an optimal balance between global and local optimization. \textcolor{blue}{HABC~\cite{djebbar2025hybrid} integrates the GA exploratory operators within the artificial bee colony (ABC) framework, strategically leveraging the complementary strengths of GA's population diversification and ABC's rapid convergence properties to enhance overall search performance. mGWOA~\cite{pham2025innovative} employs a novel partitioned population strategy that synergistically leverages the strong exploitation of the grey wolf optimizer (GWO) and the exploratory strengths of the whale optimization algorithm (WOA), further enhancing global search capabilities and solution diversity through a terminal opposition-based learning (OBL) phase.} MA~\cite{he2023memetic} implements a sophisticated framework combining generalized edge assembly crossover with multi-level optimization, achieving enhanced solution precision through systematic local search strategies.

\subsection{Sensitivity analysis}
AEDGA dynamically selects local search targets based on historical experience. It generates a uniform sequence within [$0.1$,$p$] and calculates the selection probability for each individual in the current population based on their success frequency. When $p$ is small, the algorithm prioritizes higher-ranked individuals, favoring convergence but risking local optima entrapment in complex problems. Conversely, larger $p$ values promote diversity by considering more individuals for local search, though this may compromise performance in large search spaces under limited computational resources.

The iteration count ($I$) serves as an adaptive parameter for both the local search iterations per solution and the ACO-based path optimization iterations after generating new routes. This parameter directly influences both the number of optimizable paths per solution and the optimization degree of each path. To control $I$, we introduce the iteration parameter $\sigma$. For local search operations, $I$ is calculated as $I = \lceil$ total paths in a solution $\times$ $\sigma \rceil$, while for single path optimization, $I = \lceil$ total tasks in a path $\times \sigma \rceil$.

Parameter sensitivity analysis for $p$ and $\sigma$ is essential. Initial uniform sampling analysis~\cite{chen2024archive} set both parameters' ranges to $\{0.2,0.4,0.6,0.8,1\}$. We extend $\sigma$ to $\{0.05, 0.1, 0.2, 0.4, 0.6, 0.8, 1\}$ based on the preliminary study. To optimize experimental efficiency, we adopt a streamlined approach~\cite{qiao2024constraints,qiao2024benchmark}: first analyzing $p$ with a fixed $\sigma$ value, then using the best-performing $p$ value for $\sigma$ analysis, followed by $p$ parameter validation again using the optimal $\sigma$ value.

\begin{figure*}[!hbpt]
    \centering
    \includegraphics[width=16cm]{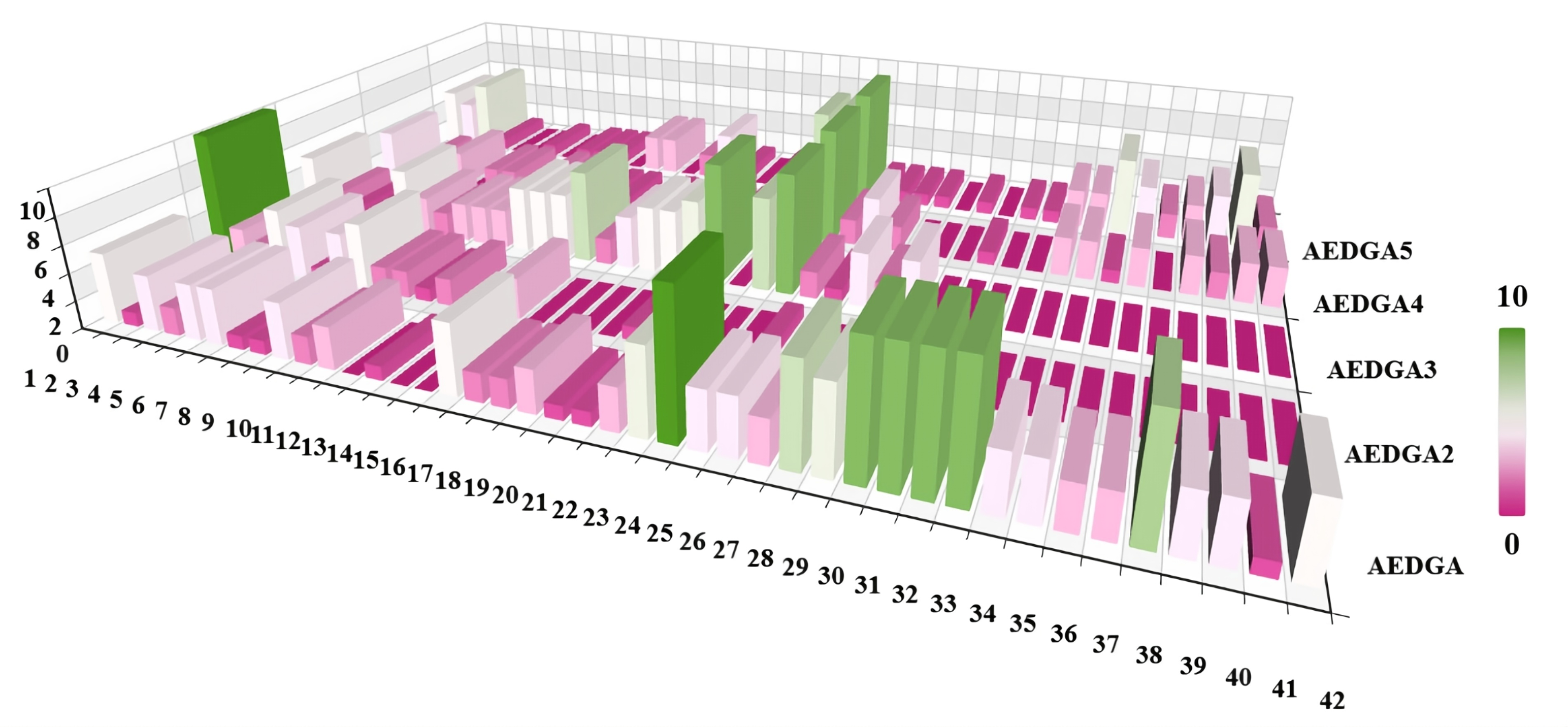}
    \caption{Test results of ablation experiment}
    \label{Test results of ablation experiment}
\end{figure*}

Fig.~\ref{Test results of parameters} illustrates the performance of these variants across $42$ test instances. Variants p$1$-p$5$ in~Fig.~\ref{Test results of parameters}(a) correspond to $p$ sampling points, while variants $\sigma1$-$\sigma7$ in~Fig.~\ref{Test results of parameters}(b) correspond to $\sigma$ sampling points. The 'Optimal count per problem' represents the total number of best performances across all algorithms in $10$ experiments per problem, typically equaling the experiment count. However, when parameter variations minimally impact certain test problems, multiple variants may achieve identical optimal results, causing the shared optimal count to exceed the experiment count. This phenomenon indirectly indicates the lower complexity of these problem instances. The 'Optimal count per variant' tracks each parameter variant's optimal performance frequency across $420$ experiments, with higher counts indicating better parameter performance. Furthermore, the instances are arranged vertically from problem $1$ to problem $42$. Color intensity represents variant performance across test problems, with darker shades indicating higher frequency of achieving optimal values in $10$ experiments.

Comprehensive analysis reveals variant p$3$ demonstrates superior overall performance in~Fig.~\ref{Test results of parameters}(a). In~Fig.~\ref{Test results of parameters}(b), $\sigma4$ performs relatively well in small and medium-scale problems, while $\sigma3$ shows excellent performance across different problem scales with notably better performance in large-scale problems. Consequently, parameter values are set to $p=0.6$ and $\sigma=0.2$ for subsequent experiments.

\subsection{Ablation experiment}

Three critical components significantly influence AEDGA's performance: ILBIM, CLSM, and EASS. To provide comprehensive validation of our research methodology and strengthen its credibility, these components are verified and analyzed collectively in this subsection.

To validate ILBIM's effectiveness, we develop AEDGA2, which replaces the initialization method with random population initialization. AEDGA3 omits CLSM to verify its contribution. Since EASS dynamically adjusts individual selection probabilities based on historical experience, for comparison, AEDGA4 maintains uniform selection probabilities for all individuals within range $p$. Additionally, AEDGA5 is designed to evaluate the effectiveness of performing local search exclusively on the population's best individual.

Fig.~\ref{Test results of ablation experiment} illustrates the performance of these $5$ variants across $42$ test instances, with $10$ runs per instance. The bar height and color intensity (trending towards green) correspond to the frequency of achieving superior results for each problem instance. AEDGA demonstrates outstanding performance across most instances, with particularly significant advantages in large-scale problems. The results direct validate the effectiveness of each algorithmic component. Notably, AEDGA3 shows competitive performance in small-scale instances, suggesting that excessive local search operations may impede algorithm progression under limited computational time.

\subsection{Comparative experiments and analysis of route generation phase}
This subsection focuses on validating the performance of the RG phase by comprehensively comparing AEDGA with other algorithms across the test instances. The detailed experimental results for each algorithm are presented in subsequent content through tabulated formats.

In the results tables~\ref{Comparative results on traditional CVRP} and~\ref{Comparative results on route generation}, the performance of heuristic algorithms is presented as mean-(standard deviation) for each instance. For MILP, results are shown as optimal value-(gap(\%)). MILP\_N represents the MILP approach with a maximum runtime of $n \,$s. To explore the potential optimal solutions, we additionally implement MILP\_{7200} with an extended maximum runtime of $7200$s, whose detailed results are displayed in the supplementary materials due to page limitation. Given the deterministic nature of MILP approaches, each test instance is executed once, and the final results are directly compared with the mean values obtained by AEDGA. The symbols '$+$', '$-$', and '$=$' indicate that the comparative algorithm performs better than, worse than, or approximately equal to AEDGA, respectively. The last row summarizes the total count of '$+$', '$-$', and '$=$' outcomes across all $42$ test instances for each algorithm.

\subsubsection{Comparative experiments on traditional CVRP}
To comprehensively evaluate the versatility and robustness of AEDGA, we first conducted comparative experiments on the traditional CVRP model, despite our algorithm being specifically designed for MTPRTS. Compared to our proposed model, the traditional CVRP eliminates the constraints corresponding to Eqs.~(\ref{eq5}) to (\ref{eq7}),~(\ref{eq9}),~(\ref{eq12}),~(\ref{eq13}), and~(\ref{eq15}) to (\ref{eq17}), while modifying the objective function in Eq.~(\ref{eq1}) to: 
\begin{equation}
 \min Z = \sum_{i=0}^n \sum_{j \neq i}^n d_{ij}x_{ij} + \sum_{i=1}^n d_{i0}b_i 
\end{equation}

The results are presented in Table~\ref{Comparative results on traditional CVRP}. For systematic analysis, we categorize the test instances into three scales: small-scale ($n \leq 50$), medium-scale ($50 < n \leq 100$), and large-scale ($n > 100$). The results reveal distinct performance characteristics of algorithms across different problem scales.

\begin{table*}[!hbpt]
\centering
\caption{\textcolor{blue}{Comparative results on traditional CVRP}}
\label{Comparative results on traditional CVRP}
\setlength{\tabcolsep}{6.5pt}
\centering\small
\scalebox{0.51}{
\begin{tabular}{clcccccccccc}
\hline
$Pro$ & \multicolumn{1}{c}{$n$} & GA-NN                   & ETSA                    & GSACO                   & HGSA                    & \textcolor{blue}{HABC}                    & \textcolor{blue}{mGWOA}                   & MA                               & MILP\_N                 & MILP\_7200                       & AEDGA                          \\
\hline
1   & 16                 & 4.66e+02 (4.87e+00) = & 5.05e+02 (1.07e+01) - & 4.63e+02 (4.33e-01) + & 4.62e+02 (8.55e-01) + & 4.63e+02 (1.92e+00) + & 4.62e+02 (5.09e+00) + & 4.65e+02 (1.39e+00) =         & 4.51e+02 (0.00e+00) + & \textbf{4.51e+02 (0.00e+00) +} & 4.70e+02 (5.29e+00)         \\
2   & 19                 & 2.66e+02 (1.22e+01) - & 4.72e+02 (1.64e+01) - & 2.28e+02 (4.61e+00) + & 2.25e+02 (7.85e+00) + & 2.34e+02 (9.90e+00) + & 2.24e+02 (1.39e+01) + & 2.51e+02 (2.15e+00) -         & 2.20e+02 (2.31e+01) + & \textbf{2.13e+02 (0.00e+00) +} & 2.44e+02 (1.21e+01)         \\
3   & 20                 & 2.74e+02 (1.51e+01) - & 4.80e+02 (2.36e+01) - & 2.23e+02 (3.74e+00) + & 2.25e+02 (7.32e+00) + & 2.30e+02 (1.39e+01) + & 2.59e+02 (1.88e+01) = & 2.54e+02 (1.15e+00) -         & 2.19e+02 (1.83e+01) + & \textbf{2.17e+02 (0.00e+00) +} & 2.39e+02 (1.77e+01)         \\
4   & 21                 & 2.72e+02 (1.86e+01) - & 4.91e+02 (2.08e+01) - & 2.19e+02 (3.52e+00) - & 2.25e+02 (1.19e+01) = & 2.41e+02 (1.39e+01) - & 2.22e+02 (9.26e+00) = & 2.57e+02 (1.44e+00) -         & 2.13e+02 (1.51e+01) + & \textbf{2.13e+02 (0.00e+00) +} & 2.15e+02 (2.52e+00)         \\
5   & 22                 & 2.84e+02 (1.95e+01) - & 5.11e+02 (3.52e+01) - & 2.26e+02 (2.10e+00) + & 2.25e+02 (6.39e+00) + & 2.20e+02 (1.45e+01) + & 2.56e+02 (2.03e+01) = & 2.64e+02 (1.17e+00) -         & 2.18e+02 (1.51e+01) + & \textbf{2.18e+02 (4.60e+00) +} & 2.44e+02 (2.11e+01)         \\
6   & 22                 & 6.25e+02 (1.83e+01) - & 7.97e+02 (1.75e+01) - & 6.22e+02 (9.60e+00) - & 5.98e+02 (1.52e+01) = & 5.94e+02 (1.31e+01) = & 6.15e+02 (2.46e+01) - & 5.94e+02 (1.36e+00) -         & 5.89e+02 (1.02e+01) + & \textbf{5.89e+02 (0.00e+00) +} & 5.90e+02 (5.18e-01)         \\
7   & 23                 & 5.56e+02 (6.94e+00) = & 7.09e+02 (9.13e+00) - & 5.73e+02 (7.76e+00) - & 5.53e+02 (7.45e+00) = & 5.61e+02 (7.35e+00) = & 5.63e+02 (1.35e+01) = & 5.62e+02 (5.89e+00) -         & 5.31e+02 (1.40e+01) + & \textbf{5.31e+02 (0.00e+00) +} & 5.54e+02 (4.97e+00)         \\
8   & 40                 & 7.18e+02 (3.48e+01) - & 1.25e+03 (2.90e+01) - & 5.08e+02 (7.83e+00) = & 5.39e+02 (3.20e+01) - & 5.84e+02 (4.32e+01) - & 5.87e+02 (4.32e+01) - & 5.09e+02 (1.17e+00) -         & 4.75e+02 (1.49e+01) + & \textbf{4.62e+02 (3.16e+00) +} & 4.97e+02 (1.38e+01)         \\
9   & 45                 & 8.40e+02 (4.14e+01) - & 1.50e+03 (4.40e+01) - & 5.79e+02 (7.73e+00) - & 6.10e+02 (2.30e+01) - & 6.69e+02 (3.17e+01) - & 6.79e+02 (5.23e+01) - & 5.75e+02 (1.11e+00) -         & 5.16e+02 (1.36e+01) + & \textbf{5.13e+02 (4.24e+00) +} & 5.49e+02 (2.36e+01)         \\
10  & 50                 & 1.03e+03 (3.13e+01) - & 1.61e+03 (3.65e+01) - & 7.72e+02 (4.63e+00) - & 8.08e+02 (3.56e+01) - & 8.17e+02 (3.41e+01) - & 8.42e+02 (3.43e+01) - & 7.72e+02 (7.30e+00) -         & 7.40e+02 (1.84e+01) + & \textbf{7.14e+02 (8.04e+00) +} & 7.46e+02 (1.27e+01)         \\
11  & 50                 & 9.69e+02 (4.25e+01) - & 1.59e+03 (5.23e+01) - & 6.20e+02 (5.56e+00) - & 6.76e+02 (3.84e+01) - & 7.08e+02 (4.18e+01) - & 7.11e+02 (3.53e+01) - & 6.06e+02 (1.82e+00) -         & 5.96e+02 (1.97e+01) + & \textbf{5.60e+02 (5.61e+00) +} & 5.99e+02 (6.90e+00)         \\
12  & 50                 & 9.87e+02 (4.23e+01) - & 1.59e+03 (2.50e+01) - & 6.96e+02 (4.09e+00) - & 7.26e+02 (1.66e+01) - & 7.68e+02 (4.43e+01) - & 7.73e+02 (3.62e+01) - & 6.79e+02 (1.63e+00) -         & 6.52e+02 (1.76e+01) + & \textbf{6.39e+02 (7.92e+00) +} & 6.70e+02 (5.24e+00)         \\
13  & 51                 & 1.11e+03 (2.08e+01) - & 1.71e+03 (4.06e+01) - & 8.53e+02 (1.10e+01) - & 8.20e+02 (2.10e+01) - & 9.14e+02 (2.84e+01) - & 9.10e+02 (2.53e+01) - & 7.78e+02 (1.60e+00) -         & 7.78e+02 (1.68e+01) - & \textbf{7.50e+02 (6.77e+00) +} & 7.76e+02 (1.07e-13)         \\
14  & 55                 & 9.92e+02 (3.05e+01) - & 1.76e+03 (3.15e+01) - & 7.55e+02 (4.67e+00) - & 8.11e+02 (3.84e+01) - & 8.32e+02 (4.73e+01) - & 8.46e+02 (3.09e+01) - & 7.88e+02 (1.01e+01) -         & 7.33e+02 (1.92e+01) - & \textbf{7.03e+02 (7.42e+00) +} & 7.30e+02 (7.36e+00)         \\
15  & 55                 & 1.22e+03 (2.34e+01) - & 1.82e+03 (3.72e+01) - & 1.02e+03 (1.02e+01) - & 1.04e+03 (1.92e+01) - & 1.08e+03 (2.84e+01) - & 1.07e+03 (2.52e+01) - & 9.87e+02 (1.49e+00) -         & 9.61e+02 (1.26e+01) + & \textbf{9.52e+02 (5.55e+00) +} & 9.84e+02 (8.47e-14)         \\
16  & 55                 & 1.01e+03 (5.60e+01) - & 1.71e+03 (7.06e+01) - & 6.40e+02 (9.44e+00) - & 7.04e+02 (3.23e+01) - & 7.61e+02 (3.45e+01) - & 7.31e+02 (2.16e+01) - & 6.31e+02 (1.30e+00) -         & 6.01e+02 (2.36e+01) + & \textbf{5.82e+02 (1.03e+01) +} & 6.23e+02 (4.80e+00)         \\
17  & 55                 & 1.08e+03 (3.86e+01) - & 1.75e+03 (4.07e+01) - & 6.42e+02 (4.64e+00) - & 7.13e+02 (2.51e+01) - & 7.42e+02 (3.44e+01) - & 7.61e+02 (4.49e+01) - & 6.33e+02 (1.48e+00) -         & 6.24e+02 (2.46e+01) + & \textbf{5.90e+02 (9.37e+00) +} & 6.27e+02 (3.03e+00)         \\
18  & 60                 & 1.26e+03 (4.65e+01) - & 1.99e+03 (4.35e+01) - & 8.42e+02 (9.64e+00) - & 9.07e+02 (4.95e+01) - & 9.08e+02 (3.95e+01) - & 9.38e+02 (3.12e+01) - & 8.09e+02 (1.63e+00) -         & 7.83e+02 (2.08e+01) + & \textbf{7.53e+02 (9.81e+00) +} & 7.87e+02 (4.58e+00)         \\
19  & 60                 & 1.37e+03 (4.38e+01) - & 2.08e+03 (2.36e+01) - & 1.07e+03 (8.57e+00) - & 1.08e+03 (3.05e+01) - & 1.12e+03 (5.29e+01) - & 1.12e+03 (2.27e+01) - & 1.00e+03 (4.92e+00) =         & 1.03e+03 (1.75e+01) - & \textbf{9.81e+02 (8.21e+00) +} & 1.01e+03 (0.00e+00)         \\
20  & 65                 & 1.32e+03 (4.82e+01) - & 2.25e+03 (3.98e+01) - & 9.00e+02 (1.02e+01) - & 9.46e+02 (1.58e+01) - & 9.71e+02 (3.23e+01) - & 1.02e+03 (5.53e+01) - & 8.42e+02 (1.59e+00) -         & 8.56e+02 (2.27e+01) - & \textbf{8.03e+02 (7.94e+00) +} & 8.39e+02 (2.40e-13)         \\
21  & 70                 & 1.40e+03 (3.69e+01) - & 2.44e+03 (5.59e+01) - & 9.59e+02 (9.73e+00) - & 9.99e+02 (4.64e+01) - & 1.05e+03 (4.36e+01) - & 1.10e+03 (6.30e+01) - & 9.30e+02 (7.23e+00) =         & 9.09e+02 (2.39e+01) - & \textbf{8.48e+02 (1.06e+01) +} & 9.05e+02 (3.84e+01)         \\
22  & 76                 & 1.27e+03 (7.67e+01) - & 2.51e+03 (1.19e+02) - & 6.83e+02 (7.36e+00) - & 7.94e+02 (2.48e+01) - & 9.12e+02 (6.35e+01) - & 9.48e+02 (6.20e+01) - & 7.94e+02 (1.37e+00) -         & 7.21e+02 (2.43e+01) - & \textbf{6.08e+02 (4.85e+00) +} & 6.70e+02 (2.03e+01)         \\
23  & 76                 & 1.30e+03 (3.65e+01) - & 2.63e+03 (1.04e+02) - & 7.20e+02 (1.13e+01) = & 8.46e+02 (3.49e+01) - & 9.54e+02 (5.61e+01) - & 9.76e+02 (4.28e+01) - & 7.95e+02 (1.76e+00) -         & 7.17e+02 (2.34e+01) - & \textbf{6.59e+02 (1.25e+01) +} & 7.03e+02 (1.86e+01)         \\
24  & 101                & 1.78e+03 (1.95e+02) - & 1.16e+03 (4.19e+01) - & 8.06e+02 (1.13e+01) - & 1.05e+03 (4.54e+01) - & 1.12e+03 (1.00e+02) - & 1.28e+03 (7.95e+01) - & 9.30e+02 (1.22e+00) -         & 7.79e+02 (1.34e+01) - & \textbf{6.91e+02 (7.80e-01) +} & 7.73e+02 (1.90e+01)         \\
25  & 41                 & 2.65e+02 (1.36e+01) - & 5.02e+02 (1.10e+01) - & 2.52e+02 (1.02e+00) - & 2.55e+02 (4.55e+00) - & 2.71e+02 (1.33e+01) - & 2.64e+02 (7.28e+00) - & 2.55e+02 (1.90e+00) -         & 2.50e+02 (2.43e+01) - & \textbf{2.44e+02 (6.81e+00) +} & 2.47e+02 (1.21e+00)         \\
26  & 61                 & 4.11e+02 (3.70e+00) - & 8.19e+02 (5.25e+01) - & 4.02e+02 (1.52e+00) - & 4.03e+02 (5.53e+00) - & 3.69e+03 (4.12e+01) - & 4.49e+03 (9.14e+01) - & 3.94e+02 (1.88e+00) -         & 4.19e+02 (2.20e+01) - & \textbf{3.81e+02 (9.55e+00) +} & 3.89e+02 (3.69e-01)         \\
27  & 81                 & 6.37e+02 (2.88e+00) - & 1.21e+03 (4.25e+01) - & 6.20e+02 (2.82e+00) - & 6.17e+02 (5.88e+00) - & 6.26e+03 (5.93e+01) - & 7.79e+03 (7.63e+01) - & 6.08e+02 (1.49e+00) -         & 6.35e+02 (2.73e+01) - & \textbf{5.91e+02 (1.34e+01) +} & 6.06e+02 (1.20e-13)         \\
28  & 91                 & 8.72e+02 (7.32e+00) - & 1.77e+03 (8.62e+01) - & 8.24e+02 (1.57e+00) - & 8.34e+02 (2.48e+01) - & 7.57e+03 (7.80e+01) - & 9.72e+03 (9.62e+01) - & 8.03e+02 (6.36e+00) =         & 8.66e+02 (3.87e+01) - & \textbf{7.70e+02 (1.97e+01) +} & 7.97e+02 (4.89e+00)         \\
29  & 136                & 1.40e+03 (1.56e+01) - & 2.84e+03 (1.01e+02) - & 1.28e+03 (1.43e+01) - & 1.32e+03 (1.58e+01) - & 7.90e+03 (6.04e+01) - & 9.47e+03 (1.20e+02) - & 1.26e+03 (6.91e+00) -         & 1.39e+03 (7.73e+01) - & 1.26e+03 (2.08e+01) -         & \textbf{1.24e+03 (3.38e+00)} \\
30  & 181                & 1.82e+03 (2.75e+01) - & 3.59e+03 (7.91e+01) - & 1.59e+03 (1.04e+01) - & 1.60e+03 (3.18e+01) - & 1.21e+04 (1.25e+02) - & 1.49e+04 (1.10e+02) - & 1.53e+03 (5.63e+00) -         & 1.72e+03 (7.85e+01) - & 1.57e+03 (2.37e+01) -         & \textbf{1.51e+03 (7.57e+00)} \\
31  & 161                & 2.52e+03 (1.72e+01) - & 4.81e+03 (1.48e+02) - & 2.34e+03 (9.23e+00) - & 2.36e+03 (1.93e+01) - & 1.36e+04 (1.05e+02) - & 1.78e+04 (2.00e+02) - & 2.25e+03 (3.87e+00) -         & 2.50e+03 (8.15e+01) - & 2.38e+03 (4.71e+01) -         & \textbf{2.23e+03 (8.75e+00)} \\
32  & 251                & 3.91e+03 (6.91e+01) - & 7.11e+03 (1.86e+02) - & 3.41e+03 (1.54e+01) - & 3.46e+03 (3.25e+01) - & 1.11e+04 (6.42e+01) - & 1.39e+04 (1.70e+02) - & \textbf{3.27e+03 (9.59e+00) +} & 3.51e+03 (8.49e+01) - & 3.36e+03 (3.67e+01) -         & 3.30e+03 (1.22e+01)         \\
33  & 321                & 4.55e+03 (2.91e+01) - & 8.55e+03 (1.91e+02) - & 3.76e+03 (1.17e+01) - & 3.82e+03 (4.56e+01) - & 1.87e+04 (1.46e+02) - & 2.37e+04 (1.27e+02) - & \textbf{3.67e+03 (2.78e+01) +} & 4.42e+03 (8.52e+01) - & 3.89e+03 (3.78e+01) -         & 3.72e+03 (2.91e+01)         \\
34  & 251                & 4.25e+03 (5.99e+01) - & 8.15e+03 (1.23e+02) - & 3.53e+03 (2.06e+01) - & 3.62e+03 (7.25e+01) - & 2.24e+04 (1.61e+02) - & 2.98e+04 (3.15e+02) - & \textbf{3.36e+03 (1.25e+01) +} & 3.83e+03 (8.33e+01) - & 3.53e+03 (2.71e+01) -         & 3.40e+03 (9.59e-13)         \\
35  & 376                & 7.42e+03 (4.57e+01) - & 1.30e+04 (2.15e+02) - & 5.91e+03 (2.07e+01) - & 6.03e+03 (4.11e+01) - & 4.19e+02 (1.03e+01) - & 4.26e+02 (1.12e+01) - & \textbf{5.60e+03 (3.16e+01) +} & 7.44e+03 (8.90e+01) - & 6.57e+03 (8.50e+01) -         & 5.73e+03 (0.00e+00)         \\
36  & 501                & 9.29e+03 (6.14e+01) - & 1.64e+04 (2.83e+02) - & 6.91e+03 (2.12e+01) - & 7.19e+03 (1.23e+02) - & 6.28e+02 (1.36e+01) - & 6.53e+02 (1.04e+01) - & \textbf{6.64e+03 (4.85e+01) +} & 1.24e+04 (9.18e+01) - & 8.21e+03 (8.76e+01) -         & 6.72e+03 (3.34e+01)         \\
37  & 361                & 9.06e+03 (1.28e+02) - & 1.56e+04 (2.52e+02) - & 7.39e+03 (3.16e+01) - & 7.53e+03 (1.47e+02) - & 8.50e+02 (2.06e+01) - & 9.05e+02 (1.80e+01) - & 7.15e+03 (4.01e+01) =         & 8.15e+03 (8.90e+01) - & 8.09e+03 (8.16e+01) -         & \textbf{7.13e+03 (2.76e+01)} \\
38  & 541                & 1.43e+04 (1.26e+02) - & 2.35e+04 (3.67e+02) - & 1.10e+04 (2.59e+01) - & 1.15e+04 (8.69e+01) - & 1.33e+03 (1.81e+01) - & 1.50e+03 (3.20e+01) - & \textbf{1.05e+04 (1.75e+01) +} & 1.35e+04 (9.12e+01) - & 1.15e+04 (8.94e+01) -         & 1.06e+04 (4.81e+01)         \\
39  & 721                & 1.71e+04 (1.40e+02) - & 2.86e+04 (4.84e+02) - & 1.18e+04 (3.24e+01) - & 1.26e+04 (1.14e+02) - & 1.62e+03 (2.23e+01) - & 1.89e+03 (3.41e+01) - & \textbf{1.12e+04 (5.54e+01) +} & 2.15e+04 (9.32e+01) - & 1.56e+04 (9.06e+01) -         & 1.13e+04 (4.83e+01)         \\
40  & 491                & 1.33e+04 (1.27e+02) - & 2.30e+04 (2.73e+02) - & 1.00e+04 (3.22e+01) - & 1.04e+04 (1.38e+02) - & 2.41e+03 (3.27e+01) - & 2.69e+03 (3.75e+01) - & \textbf{9.54e+03 (4.39e+01) =} & 1.11e+04 (8.91e+01) - & 1.10e+04 (8.89e+01) -         & 9.57e+03 (4.20e+01)         \\
41  & 736                & 2.25e+04 (9.82e+01) - & 3.66e+04 (5.48e+02) - & 1.65e+04 (2.83e+01) - & 1.74e+04 (1.81e+02) - & 3.54e+03 (4.21e+01) - & 4.12e+03 (5.19e+01) - & \textbf{1.61e+04 (1.12e+02) +} & 2.88e+04 (9.45e+01) - & 2.20e+04 (9.27e+01) -         & 1.62e+04 (6.46e+01)         \\
42  & 981                & 2.83e+04 (8.91e+01) - & 4.55e+04 (4.44e+02) - & 1.85e+04 (3.67e+01) - & 2.02e+04 (1.05e+02) - & 3.96e+03 (3.80e+01) - & 4.87e+03 (6.14e+01) - & \textbf{1.76e+04 (8.89e+01) +} & 3.10e+04 (9.36e+01) - & 3.10e+04 (9.36e+01) -         & 1.83e+04 (0.00e+00)              \\
\hline
\multicolumn{2}{c}{+/-/=}   & \textbf{0/40/2}         & \textbf{0/42/0}         & \textbf{4/36/2}         & \textbf{4/35/3}         & \textbf{4/36/2}         & \textbf{2/37/4}         & \textbf{9/27/6}                  & \textbf{16/26/0}        & \textbf{28/14/0}                 &                               
\\ \hline
\end{tabular}
}
\end{table*}

\begin{table}[]
\centering 
  \caption{\textcolor{blue}{Multi-problem Wilcoxon's test results on traditional CVRP}}
  \setlength{\tabcolsep}{2.0pt}
\begin{tabular}{ccccc}
\hline
AEDGA VS        & $R^{+}$ & $R^{-}$ & Asymptotic P-value & P-value $\leq 0.05$   \\ \hline
GA-NN     & 901.0   & 2.0     & 0    & Yes                \\ 
ETSA      & 903.0   & 0.0     & 0     & Yes                   \\ 
GSACO     & 868.0   & 35.0    & 0       & Yes                   \\ 
HGSA      & 872.5   & 30.5    & 0     & Yes                   \\ 
\textcolor{blue}{HABC}      & 616.0 & 287.0    & 0.038189     & Yes                   \\ 
\textcolor{blue}{mGWOA}      & 635.5 & 267.5    & 0.020659     & Yes                   \\ 
MA        & 534.5   & 368.5   & 0.269387   &      No      \\ 
MILP\_N    & 675.0   & 228.0   & 0.004011      &  Yes           \\ 
MILP\_7200 & 471.0   & 432.0   & 0.800701    &      No      \\ \hline
\end{tabular}
\label{Wil_cvrp}
\end{table}

For small-scale instances, MILP demonstrates superior performance through its exact solving capability. GSACO and HGSA also achieve notable performance through their unique path-independent optimization strategies. \textcolor{blue}{Similarly, the proposed HABC and mGWOA exhibit competitive performance, effectively leveraging their hybrid structures to navigate the search space and often outperforming baseline heuristics such as GA-NN and ETSA, thereby validating their foundational design.} Notably, the MA exhibits excessive local search capabilities, with its small standard deviation indicating high stability, while simultaneously suggesting the strong tendency for premature convergence to local optima. Although the proposed AEDGA does not achieve optimal solutions across all instances, it demonstrates competitive performance with minimal gaps from the best-known solutions and maintains consistent stability.

As the problem scale extends to medium-size instances, the performance differentiation among algorithms becomes pronounced. Particularly noteworthy is that under the same time constraint ($n \,$s), MILP\_N struggles to maintain its advantage observed in small-scale problems. Even with extended runtime to $7200$s (MILP\_{7200}), it cannot guarantee global optimal solutions, significantly limiting its practical applicability. \textcolor{blue}{In this range, the performance of HGSA, HABC and mGWOA begins to lag behind more refined heuristics like MA and AEDGA. Their gradually increasing standard deviations suggest a challenge in consistently converging to high-quality solutions as the search space complexity grows.} In contrast, AEDGA obtains high-quality solutions within time proportional to problem size, offering a more practical balance between efficiency and solution quality. Moreover, AEDGA maintains excellent stability despite increased problem complexity, progressively outperforming other heuristic methods in both solution quality and convergence stability, demonstrating superior scalability.

In large-scale problem domains, the performance divergence becomes more evident. Particularly, even MILP\_$7200$ shows limited improvement in solution quality, further confirming the limitations of exact methods in handling large-scale problems. By comparison, the MA algorithm, leveraging its strong local search capability, rapidly identifies high-quality local optimal solutions. Although AEDGA shows slightly inferior performance on traditional CVRPs due to its comprehensive consideration of load and distance relationships during initialization, it still demonstrates good adaptability to large-scale problems.

Through in-depth analysis of experimental data, AEDGA demonstrates significant advantages in balancing solution quality and algorithmic stability. This characteristic contrasts sharply with other algorithms: especially, ETSA shows significant stability deterioration as problem size increases (standard deviation increasing from $10$ to $500$), while GA-NN also exhibits moderate stability decay (standard deviation rising from $5$ to $100$). Considering the diversity and complexity of problem characteristics in practical applications, AEDGA's low sensitivity to problem features makes it particularly valuable, positioning it as a promising universal solution for practical CVRPs.

The Wilcoxon test results presented in Table~\ref{Wil_cvrp} further substantiate AEDGA's significant advantages over comparative algorithms. In comparisons with GA-NN, ETSA, GSACO, HGSA, \textcolor{blue}{HABC, mGWOA} and MILP\_N, AEDGA achieves notably high $R^{+}$ values and low $R^{-}$ values, with corresponding $P$-values substantially below the $0.05$ significance level. These statistics strongly indicate AEDGA's superior performance on problem-solving capabilities. In comparisons to more challenging competitors, namely MA and MILP\_$7200$, while the $P$-values suggest smaller statistical significance differences for this model, AEDGA maintains strong competitiveness with higher $R^{+}$ values compared to $R^{-}$ values. These findings comprehensively validate not only AEDGA's effectiveness in problem-solving but also its significant superiority in solution quality.

\subsubsection{Comparative experiments on route geneartion}
The RG model excludes constraints~\ref{eq15} to~\ref{eq17}. Compared to the traditional CVRP model, it introduces additional complexity through the consideration of real-time load impacts. The experimental results are presented in Table~\ref{Comparative results on route generation}.

\begin{table*}[!hbpt]
\centering
\caption{\textcolor{blue}{Comparative results on route generation}}
\label{Comparative results on route generation}
\setlength{\tabcolsep}{6.5pt}
\centering\small
\scalebox{0.51}{
\begin{tabular}{clcccccccccc}
\hline
$Pro$ & \multicolumn{1}{c}{$n$} & GA-NN                   & ETSA                    & GSACO                   & HGSA                    & \textcolor{blue}{HABC}                    & \textcolor{blue}{mGWOA}                   & MA                               & MILP\_N                 & MILP\_7200                       & AEDGA                          \\
 \hline
1   & 16                    & 1.15e+04 (1.49e+02) = & 1.24e+04 (1.51e+02) - & 1.20e+04 (6.76e+01) - & 1.14e+04 (7.81e+01) + & 1.14e+04 (1.32e+02) + & 1.14e+04 (1.86e+02) + & 1.14e+04 (1.21e-12) =          & 1.14e+04 (0.00e+00) + & \textbf{1.14e+04 (0.00e+00) +} & 1.15e+04 (1.40e+02)          \\
2   & 19                    & 2.60e+04 (8.50e+02) - & 3.75e+04 (9.28e+02) - & 2.81e+04 (9.23e+02) - & 2.52e+04 (4.20e+02) = & 2.74e+04 (1.99e+03) - & 2.84e+04 (1.37e+03) - & 2.53e+04 (3.76e+02) -          & 2.42e+04 (3.52e+01) + & \textbf{2.38e+04 (1.03e+01) +} & 2.47e+04 (1.97e+02)          \\
3   & 20                    & 2.71e+04 (1.12e+03) - & 3.89e+04 (6.27e+02) - & 2.91e+04 (6.58e+02) - & 2.65e+04 (5.73e+02) - & 2.98e+04 (2.06e+03) - & 3.08e+04 (1.23e+03) - & 2.60e+04 (6.71e+02) -          & 2.46e+04 (3.95e+01) + & \textbf{2.42e+04 (1.68e+01) +} & 2.51e+04 (2.87e+02)          \\
4   & 21                    & 2.66e+04 (1.09e+03) - & 4.01e+04 (1.12e+03) - & 2.87e+04 (8.32e+02) - & 2.61e+04 (6.59e+02) - & 2.92e+04 (1.51e+03) - & 3.11e+04 (2.68e+03) - & 2.60e+04 (6.57e+02) -          & 2.43e+04 (4.00e+01) + & \textbf{2.42e+04 (2.24e+01) +} & 2.50e+04 (1.32e+02)          \\
5   & 22                    & 2.82e+04 (1.06e+03) - & 4.17e+04 (1.05e+03) - & 2.99e+04 (4.96e+02) - & 2.71e+04 (1.13e+03) - & 2.95e+04 (1.77e+03) - & 3.32e+04 (2.16e+03) - & 2.70e+04 (6.06e+02) -          & 2.54e+04 (4.06e+01) + & \textbf{2.51e+04 (2.35e+01) +} & 2.62e+04 (1.34e+02)          \\
6   & 22                    & 1.34e+06 (1.45e+04) - & 1.34e+06 (1.31e+04) - & 1.41e+06 (8.09e+03) - & 1.31e+06 (6.58e+03) - & 1.36e+06 (3.19e+04) - & 1.35e+06 (1.68e+04) - & 1.30e+06 (3.41e+03) =          & 1.31e+06 (2.70e+01) - & \textbf{1.30e+06 (0.00e+00) +} & 1.30e+06 (5.56e+02)          \\
7   & 23                    & 1.59e+04 (9.51e+01) - & 1.87e+04 (2.73e+02) - & 1.72e+04 (1.90e+02) - & 1.56e+04 (6.11e+01) + & 1.63e+04 (2.75e+02) - & 1.64e+04 (3.53e+02) - & 1.60e+04 (1.56e+02) -          & 1.56e+04 (3.05e+00) + & \textbf{1.56e+04 (0.00e+00) +} & 1.58e+04 (7.14e+01)          \\
8   & 40                    & 5.17e+04 (1.08e+03) - & 8.42e+04 (1.27e+03) - & 5.89e+04 (8.88e+02) - & 5.27e+04 (2.11e+03) - & 6.16e+04 (2.52e+03) - & 6.30e+04 (4.91e+03) - & 5.17e+04 (5.79e+02) -          & 4.78e+04 (3.20e+01) + & \textbf{4.48e+04 (1.29e+01) +} & 4.91e+04 (1.43e+03)          \\
9   & 45                    & 6.27e+04 (1.19e+03) - & 1.07e+05 (1.13e+03) - & 7.41e+04 (1.08e+03) - & 6.43e+04 (3.38e+03) - & 7.67e+04 (3.51e+03) - & 7.73e+04 (1.85e+03) - & 5.98e+04 (1.31e+03) -          & 5.49e+04 (2.99e+01) + & \textbf{5.28e+04 (1.73e+01) +} & 5.76e+04 (8.73e+02)          \\
10  & 50                    & 5.55e+04 (5.99e+02) - & 8.85e+04 (7.39e+02) - & 6.28e+04 (6.60e+02) - & 5.62e+04 (1.29e+03) - & 5.95e+04 (2.26e+03) - & 6.00e+04 (2.17e+03) - & 5.63e+04 (5.55e+02) -          & 5.20e+04 (1.68e+01) + & \textbf{4.96e+04 (1.60e+00) +} & 5.33e+04 (7.10e+02)          \\
11  & 50                    & 6.79e+04 (6.93e+02) - & 1.17e+05 (1.81e+03) - & 7.81e+04 (8.23e+02) - & 7.21e+04 (3.11e+03) - & 7.99e+04 (4.77e+03) - & 8.05e+04 (5.91e+03) - & 6.86e+04 (5.24e+02) -          & 6.26e+04 (2.47e+01) + & \textbf{5.86e+04 (6.92e+00) +} & 6.41e+04 (9.89e+02)          \\
12  & 50                    & 6.08e+04 (9.10e+02) - & 1.00e+05 (7.00e+02) - & 6.77e+04 (1.36e+03) - & 6.29e+04 (3.23e+03) - & 6.79e+04 (3.56e+03) - & 6.99e+04 (3.14e+03) - & 5.86e+04 (3.28e+02) =          & 5.42e+04 (1.69e+01) + & \textbf{5.31e+04 (2.79e+00) +} & 5.83e+04 (7.82e+02)          \\
13  & 51                    & 4.78e+04 (7.06e+02) - & 7.55e+04 (9.26e+02) - & 5.55e+04 (9.78e+02) - & 4.79e+04 (1.66e+03) - & 5.35e+04 (2.68e+03) - & 5.52e+04 (1.74e+03) - & 4.69e+04 (1.22e+02) -          & 4.29e+04 (1.54e+01) + & \textbf{4.27e+04 (4.98e+00) +} & 4.49e+04 (3.79e+02)          \\
14  & 55                    & 6.56e+04 (1.10e+03) - & 1.07e+05 (2.66e+03) - & 7.33e+04 (6.16e+02) - & 6.49e+04 (3.30e+03) - & 7.11e+04 (2.44e+03) - & 7.42e+04 (3.73e+03) - & 6.38e+04 (8.59e+02) -          & 6.13e+04 (2.63e+01) + & \textbf{5.61e+04 (4.16e+00) +} & 6.14e+04 (6.60e+02)          \\
15  & 55                    & 5.18e+04 (4.98e+02) - & 7.70e+04 (9.14e+02) - & 5.78e+04 (5.50e+02) - & 5.14e+04 (9.90e+02) = & 5.41e+04 (1.03e+03) - & 5.53e+04 (1.15e+03) - & 5.10e+04 (3.61e+02) =          & 4.88e+04 (1.35e+01) + & \textbf{4.83e+04 (9.50e-01) +} & 5.09e+04 (6.31e+02)          \\
16  & 55                    & 8.24e+04 (1.76e+03) - & 1.41e+05 (1.35e+03) - & 8.93e+04 (1.31e+03) - & 8.14e+04 (3.39e+03) - & 9.44e+04 (5.52e+03) - & 1.01e+05 (4.04e+03) - & 7.48e+04 (5.33e+02) -          & 7.40e+04 (3.25e+01) - & \textbf{6.80e+04 (1.71e+01) +} & 7.26e+04 (1.42e+03)          \\
17  & 55                    & 7.74e+04 (9.22e+02) - & 1.35e+05 (2.02e+03) - & 8.66e+04 (1.22e+03) - & 7.92e+04 (3.50e+03) - & 8.98e+04 (6.18e+03) - & 9.21e+04 (5.38e+03) - & 7.23e+04 (9.02e+02) -          & 6.93e+04 (2.96e+01) + & \textbf{6.54e+04 (1.30e+01) +} & 7.02e+04 (1.51e+03)          \\
18  & 60                    & 7.49e+04 (1.15e+03) - & 1.26e+05 (1.99e+03) - & 8.26e+04 (9.63e+02) - & 7.65e+04 (3.65e+03) - & 8.27e+04 (4.61e+03) - & 8.80e+04 (4.99e+03) - & 7.43e+04 (6.72e+02) -          & 6.99e+04 (2.58e+01) - & \textbf{6.31e+04 (5.83e+00) +} & 6.86e+04 (5.32e+02)          \\
19  & 60                    & 6.23e+04 (8.84e+02) - & 9.61e+04 (1.38e+03) - & 6.98e+04 (8.77e+02) - & 6.22e+04 (2.21e+03) - & 6.50e+04 (2.27e+03) - & 6.69e+04 (2.47e+03) - & 6.13e+04 (3.70e+02) -          & 5.76e+04 (1.27e+01) + & \textbf{5.64e+04 (2.71e+00) +} & 5.85e+04 (3.97e+02)          \\
20  & 65                    & 8.86e+04 (1.66e+03) - & 1.50e+05 (1.13e+03) - & 9.79e+04 (6.85e+02) - & 8.86e+04 (3.09e+03) - & 9.71e+04 (3.85e+03) - & 1.02e+05 (5.22e+03) - & 8.42e+04 (6.36e+02) -          & 8.22e+04 (2.81e+01) - & \textbf{7.38e+04 (8.44e+00) +} & 7.97e+04 (5.01e+02)          \\
21  & 70                    & 9.86e+04 (9.94e+02) - & 1.67e+05 (2.02e+03) - & 1.06e+05 (1.55e+03) - & 9.81e+04 (3.72e+03) - & 1.05e+05 (6.09e+03) - & 1.12e+05 (6.16e+03) - & 9.12e+04 (8.15e+02) -          & 9.05e+04 (2.91e+01) - & \textbf{7.99e+04 (1.04e+01) +} & 8.65e+04 (5.07e+02)          \\
22  & 76                    & 2.02e+05 (5.00e+03) - & 3.71e+05 (8.15e+03) - & 2.05e+05 (2.61e+03) - & 2.14e+05 (9.26e+03) - & 2.47e+05 (1.50e+04) - & 2.72e+05 (1.80e+04) - & 1.51e+05 (9.73e+02) +          & 1.61e+05 (3.96e+01) - & \textbf{1.41e+05 (2.73e+01) +} & 1.54e+05 (2.57e+03)          \\
23  & 76                    & 1.72e+05 (1.76e+03) - & 3.15e+05 (3.23e+03) - & 1.68e+05 (1.18e+03) - & 1.80e+05 (1.02e+04) - & 2.06e+05 (2.01e+04) - & 2.24e+05 (9.55e+03) - & 1.32e+05 (1.08e+03) +          & 1.42e+05 (4.01e+01) - & \textbf{1.21e+05 (2.53e+01) +} & 1.36e+05 (2.58e+03)          \\
24  & 101                   & 2.81e+05 (1.08e+04) - & 5.71e+05 (4.97e+03) - & 2.60e+05 (5.26e+03) - & 2.99e+05 (1.54e+04) - & 3.44e+05 (2.52e+04) - & 4.10e+05 (2.11e+04) - & 1.98e+05 (2.58e+02) +          & 2.10e+05 (3.36e+01) - & \textbf{1.81e+05 (2.11e+01) +} & 2.00e+05 (3.05e+02)          \\
25  & 41                    & 5.99e+04 (7.83e+02) - & 9.08e+04 (9.31e+02) - & 6.32e+04 (8.06e+02) - & 5.68e+04 (1.66e+03) - & 5.92e+04 (2.39e+03) - & 5.98e+04 (1.55e+03) - & 5.85e+04 (1.02e+03) -          & 5.74e+04 (3.31e+01) - & \textbf{5.26e+04 (2.59e+00) +} & 5.45e+04 (5.30e+02)          \\
26  & 61                    & 9.58e+04 (9.91e+02) - & 1.52e+05 (2.00e+03) - & 1.01e+05 (7.53e+02) - & 9.36e+04 (2.29e+03) - & 8.48e+05 (7.59e+03) - & 1.06e+06 (1.74e+04) - & 9.11e+04 (8.35e+02) -          & 9.68e+04 (2.80e+01) - & \textbf{8.56e+04 (3.60e+00) +} & 8.89e+04 (1.02e+03)          \\
27  & 81                    & 1.51e+05 (7.50e+02) - & 2.42e+05 (2.48e+03) - & 1.58e+05 (1.25e+03) - & 1.44e+05 (2.57e+03) - & 1.45e+06 (1.28e+04) - & 1.80e+06 (2.46e+04) - & 1.46e+05 (4.17e+02) -          & 1.52e+05 (2.52e+01) - & \textbf{1.33e+05 (2.98e+00) +} & 1.40e+05 (1.26e+03)          \\
28  & 91                    & 2.10e+05 (4.09e+03) - & 3.39e+05 (3.37e+03) - & 2.06e+05 (1.18e+03) - & 1.97e+05 (3.83e+03) - & 1.75e+06 (1.20e+04) - & 2.31e+06 (2.26e+04) - & 1.86e+05 (8.32e+02) -          & 2.00e+05 (2.32e+01) - & \textbf{1.73e+05 (2.88e+00) +} & 1.82e+05 (1.33e+03)          \\
29  & 136                   & 3.44e+05 (3.55e+03) - & 5.63e+05 (5.04e+03) - & 3.26e+05 (1.08e+03) - & 3.10e+05 (5.80e+03) - & 1.82e+06 (1.32e+04) - & 2.23e+06 (1.78e+04) - & 2.97e+05 (5.61e+02) -          & 3.15e+05 (6.59e+01) - & \textbf{2.86e+05 (5.97e+00) +} & 2.93e+05 (7.42e+02)          \\
30  & 181                   & 4.38e+05 (4.85e+03) - & 7.12e+05 (5.48e+03) - & 4.00e+05 (1.36e+03) - & 3.85e+05 (5.12e+03) - & 2.79e+06 (1.21e+04) - & 3.52e+06 (3.45e+04) - & 3.59e+05 (9.74e+02) -          & 4.15e+05 (8.71e+01) - & \textbf{3.50e+05 (9.19e+00) +} & 3.54e+05 (1.13e+03)          \\
31  & 161                   & 6.33e+05 (2.49e+03) - & 1.01e+06 (8.25e+03) - & 5.73e+05 (1.76e+03) - & 5.60e+05 (1.08e+04) - & 3.15e+06 (1.22e+04) - & 4.26e+06 (4.59e+04) - & 5.35e+05 (9.92e+02) -          & 5.93e+05 (8.34e+01) - & \textbf{5.24e+05 (6.00e+00) +} & 5.31e+05 (1.56e+03)          \\
32  & 251                   & 9.74e+05 (3.65e+03) - & 1.54e+06 (1.18e+04) - & 8.38e+05 (2.32e+03) - & 8.27e+05 (1.30e+04) - & 2.56e+06 (2.13e+04) - & 3.31e+06 (3.79e+04) - & 7.80e+05 (4.87e+02) =          & 8.67e+05 (9.15e+01) - & \textbf{7.78e+05 (9.37e+00) +} & 7.80e+05 (1.06e+03)          \\
33  & 321                   & 1.13e+06 (9.43e+03) - & 1.77e+06 (1.59e+04) - & 9.30e+05 (2.11e+03) - & 9.38e+05 (1.59e+04) - & 4.32e+06 (2.93e+04) - & 5.60e+06 (5.96e+04) - & \textbf{8.49e+05 (1.33e+03) =} & 9.34e+05 (8.53e+01) - & 8.81e+05 (1.36e+01) -          & 8.49e+05 (1.01e+03)          \\
34  & 251                   & 1.07e+06 (9.04e+03) - & 1.65e+06 (1.21e+04) - & 8.67e+05 (2.75e+03) - & 8.79e+05 (1.05e+04) - & 5.25e+06 (3.18e+04) - & 7.16e+06 (6.59e+04) - & 8.00e+05 (2.07e+02) -          & 8.87e+05 (8.75e+01) - & \textbf{7.93e+05 (9.66e+00) +} & 7.97e+05 (1.92e+03)          \\
35  & 376                   & 1.87e+06 (6.95e+03) - & 2.81e+06 (2.03e+04) - & 1.44e+06 (4.10e+03) - & 1.49e+06 (1.80e+04) - & 9.35e+04 (2.42e+03) - & 9.61e+04 (2.14e+03) - & 1.34e+06 (1.28e+03) =          & 1.48e+06 (8.91e+01) - & 1.40e+06 (3.36e+01) -          & \textbf{1.34e+06 (9.72e+02)} \\
36  & 501                   & 2.33e+06 (1.05e+04) - & 3.41e+06 (1.29e+04) - & 1.69e+06 (1.93e+03) - & 1.81e+06 (1.39e+04) - & 1.42e+05 (2.81e+03) = & 1.49e+05 (2.46e+03) - & \textbf{1.56e+06 (8.86e+02) =} & 1.71e+06 (9.12e+01) - & 1.70e+06 (4.30e+01) -          & 1.56e+06 (5.35e+02)          \\
37  & 361                   & 2.30e+06 (2.16e+04) - & 3.43e+06 (1.66e+04) - & 1.80e+06 (2.79e+03) - & 1.87e+06 (2.43e+04) - & 1.92e+05 (3.82e+03) - & 2.07e+05 (3.01e+03) - & 1.70e+06 (8.95e+02) -          & 3.16e+06 (9.51e+01) - & 1.90e+06 (1.92e+01) -          & \textbf{1.70e+06 (6.54e+02)} \\
38  & 541                   & 3.60e+06 (5.46e+04) - & 5.25e+06 (2.94e+04) - & 2.67e+06 (2.99e+03) - & 2.84e+06 (1.90e+04) - & 3.00e+05 (4.64e+03) - & 3.41e+05 (4.86e+03) - & 2.52e+06 (7.95e+02) -          & 2.80e+06 (9.38e+01) - & 2.76e+06 (9.32e+01) -          & \textbf{2.52e+06 (6.96e+02)} \\
39  & 721                   & 4.17e+06 (2.57e+04) - & 5.94e+06 (3.48e+04) - & 2.86e+06 (4.76e+03) - & 3.19e+06 (2.04e+04) - & 3.70e+05 (5.01e+03) - & 4.45e+05 (1.03e+04) - & 2.64e+06 (7.89e+02) =          & 5.90e+06 (9.70e+01) - & 3.65e+06 (9.37e+01) -          & \textbf{2.64e+06 (6.68e+02)} \\
40  & 491                   & 3.31e+06 (2.68e+04) - & 4.88e+06 (1.60e+04) - & 2.43e+06 (4.20e+03) - & 2.63e+06 (2.78e+04) - & 5.47e+05 (7.49e+03) - & 6.20e+05 (1.01e+04) - & \textbf{2.27e+06 (1.60e+03) =} & 5.13e+06 (9.67e+01) - & 2.94e+06 (7.94e+01) -          & 2.27e+06 (5.38e+02)          \\
41  & 736                   & 5.67e+06 (6.39e+04) - & 8.09e+06 (1.88e+04) - & 3.97e+06 (4.84e+03) - & 4.41e+06 (3.70e+04) - & 8.11e+05 (5.00e+03) - & 9.56e+05 (7.37e+03) - & \textbf{3.74e+06 (5.80e+02) =} & 8.90e+06 (9.77e+01) - & 4.64e+06 (9.46e+01) -          & 3.74e+06 (4.48e+02)          \\
42  & 981                   & 6.96e+06 (7.02e+04) - & 9.59e+06 (3.67e+04) - & 4.47e+06 (5.00e+03) - & 5.27e+06 (1.98e+04) - & 9.14e+05 (9.08e+03) - & 1.13e+06 (1.15e+04) - & 4.18e+06 (1.48e+03) -          & 9.72e+06 (9.75e+01) - & 6.68e+06 (9.55e+01) -          & \textbf{4.18e+06 (1.06e+03)} \\
 \hline
\multicolumn{2}{c}{+/-/=}   & \textbf{0/41/1}         & \textbf{0/42/0}         & \textbf{0/42/0}         & \textbf{2/38/2}         & \textbf{1/40/1}         & \textbf{1/41/0}         & \textbf{3/28/11}                 & \textbf{16/26/0}        & \textbf{33/9/0}                  &                               
\\ \hline
\end{tabular}
}
\end{table*}

\begin{table}[]
\centering 
  \caption{\textcolor{blue}{Multi-problem Wilcoxon's test results on RG}}
  \setlength{\tabcolsep}{2.0pt}
\begin{tabular}{ccccc}
\hline
AEDGA VS        & $R^{+}$ & $R^{-}$ & Asymptotic P-value & P-value $\leq 0.05$   \\ \hline
GA-NN     & 861.0 & 0.0      & 0    & Yes                \\ 
ETSA      & 903.0 & 0.0     & 0     & Yes                   \\ 
GSACO     & 903.0 & 0.0    & 0       & Yes                   \\ 
HGSA      & 900.0 & 3.0    & 0     & Yes                   \\ 
\textcolor{blue}{HABC}      & 633.0 & 270.0    & 0.022653     & Yes                   \\ 
\textcolor{blue}{mGWOA}    & 644.0 & 259.0    & 0.015813     & Yes                   \\ 
MA        & 779.5 & 123.5  & 0.000016   &      Yes      \\ 
MILP\_N    & 726.0 & 135.0   & 0.000105      &  Yes           \\ 
MILP\_7200 & 343.5 & 559.5   & 1    &      No      \\ \hline
\end{tabular}
\label{Wil_RG}
\end{table}

For small-scale instances, AEDGA demonstrates superior performance, surpassing MILP\_N in solution quality across several instances while maintaining enhanced stability. In contrast to the performance on CVRP, GSACO and HGSA show diminished performance advantages, indicating limitations in their distance-based local search strategies when handling complex objective functions. \textcolor{blue}{The other hybrid algorithms, HABC and mGWOA, also struggle to adapt to the RG model's increased complexity; while competitive on the simplest instances, their solution quality and stability quickly deteriorate as problem size grows, revealing challenges in managing the intricate load-dependent objective.}

The experimental results for medium-scale problems further highlight AEDGA's strengths. Although MA maintains good stability, it exhibits significant gaps in solution quality compared to AEDGA. Notably, MILP's performance deteriorates dramatically, reflecting the substantial impact of increased problem complexity on exact solution methods.

In large-scale problems, AEDGA exhibits exceptional performance advantages. Other methods exhibit significant performance deterioration, particularly MILP. MA also fails to maintain its previously demonstrated strong search capabilities. These results validate the significance of AEDGA's balanced consideration of load and distance factors in its design architecture, while highlighting its distinctive advantages in handling complex problems that closely approximate real-world scenarios.

Table~\ref{Wil_RG} presents the Wilcoxon test results comparing AEDGA with comparative algorithms on the RG model. In comparison to GA-NN, ETSA, GSACO, and HGSA, AEDGA achieves substantially high $R^{+}$ values and near-zero $R^{-}$ values, with extremely small $P$-values, strongly indicating its statistically significant superiority over these algorithms. AEDGA also demonstrates distinct advantages over \textcolor{blue}{HABC, mGWOA, }MA and MILP\_$N$. Notably, while the $R$ values suggest marginally inferior performance compared to MILP\_$7200$, the $P$-value approaching $1$ indicates negligible statistical significance in their performance difference. When considering the substantial disparity in computational time, these results further emphasize AEDGA's significant practical value for real-world applications.

\subsubsection{Case study}

\begin{figure}[htbp]
    \centering
     
    \subfigure[ETSA]{
        \includegraphics[width=0.206\textwidth]{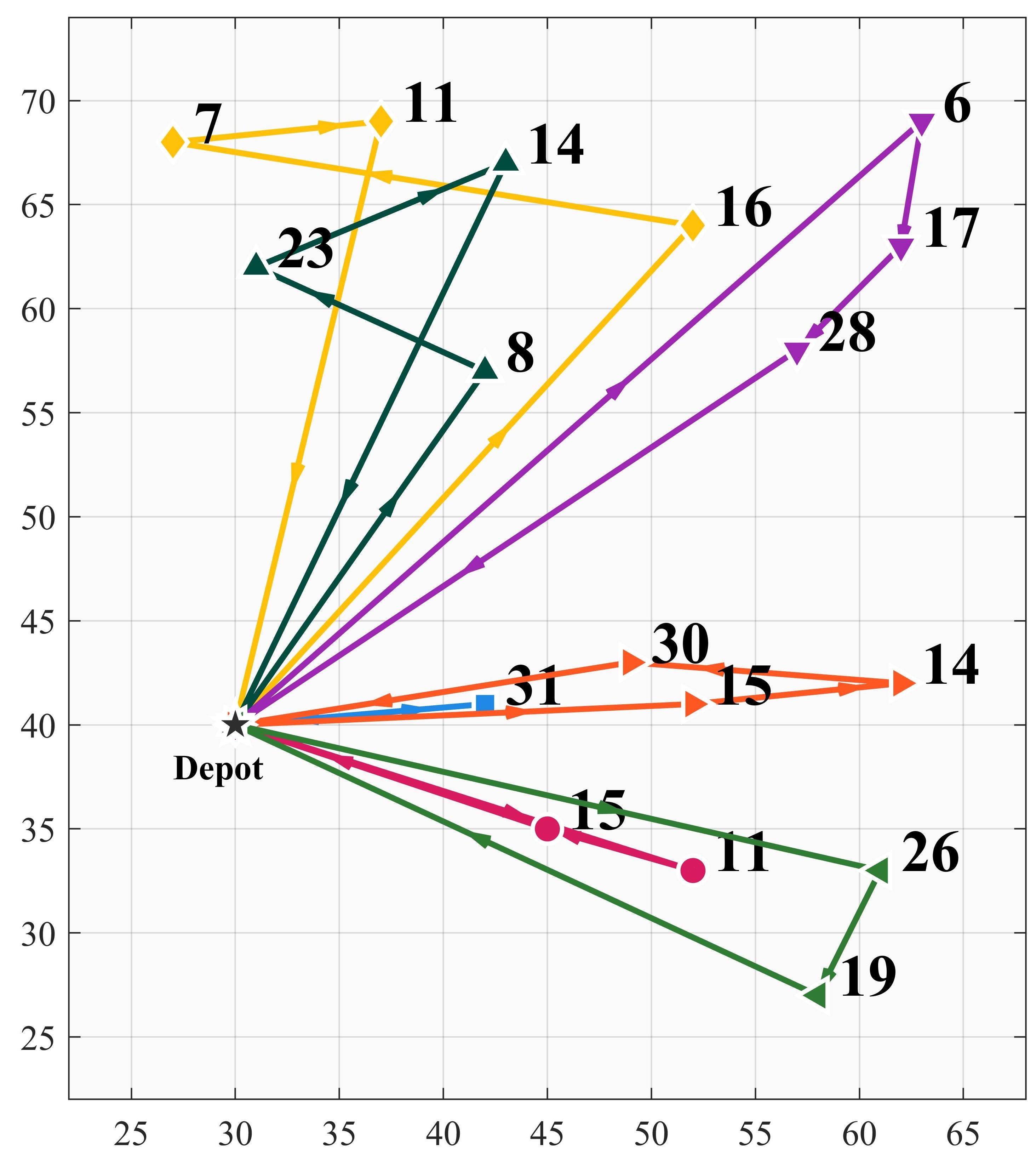}
    }
    \subfigure[GA-NN]{
        \includegraphics[width=0.206\textwidth]{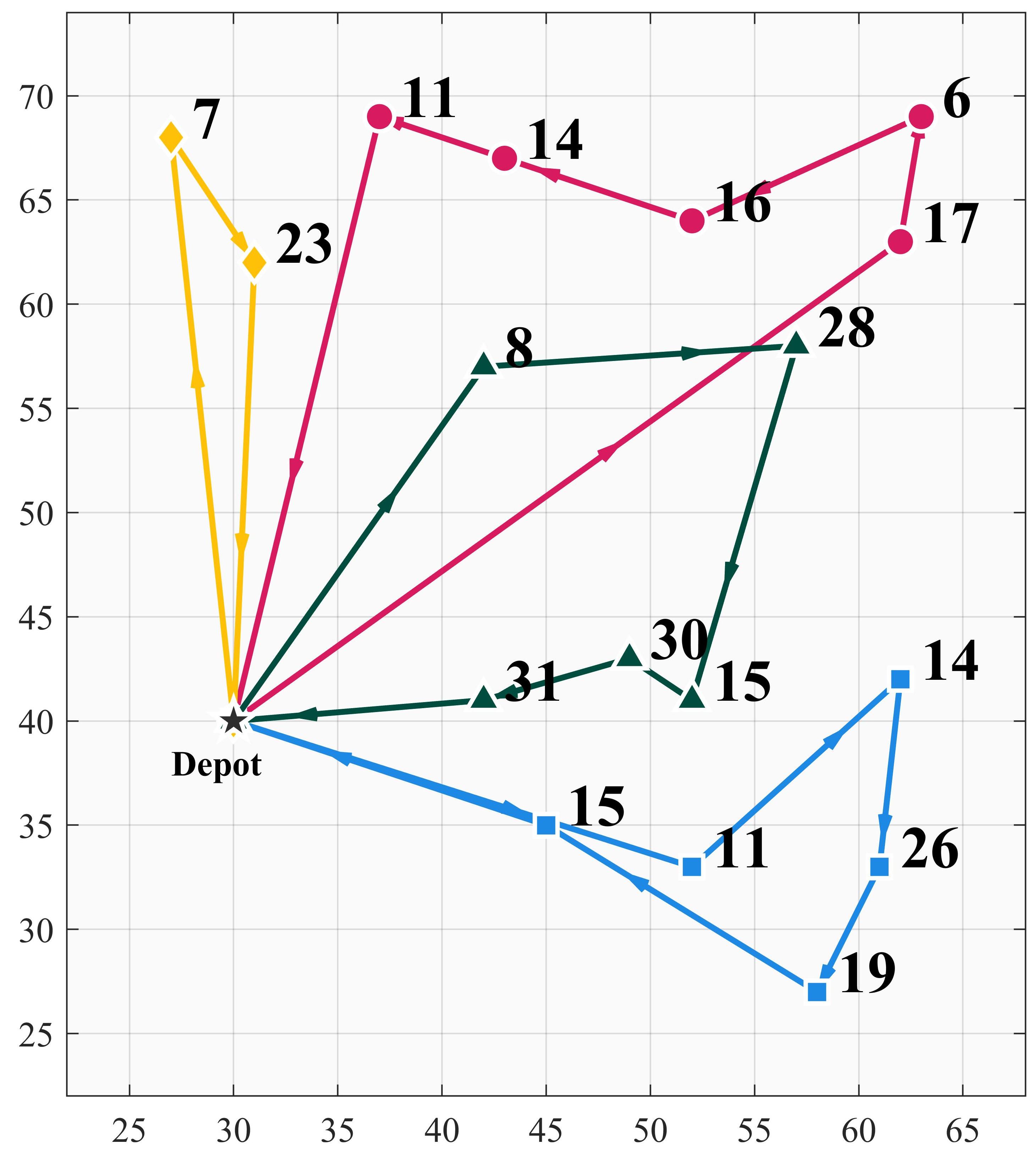}
    }
    
    \subfigure[GSACO]{
        \includegraphics[width=0.206\textwidth]{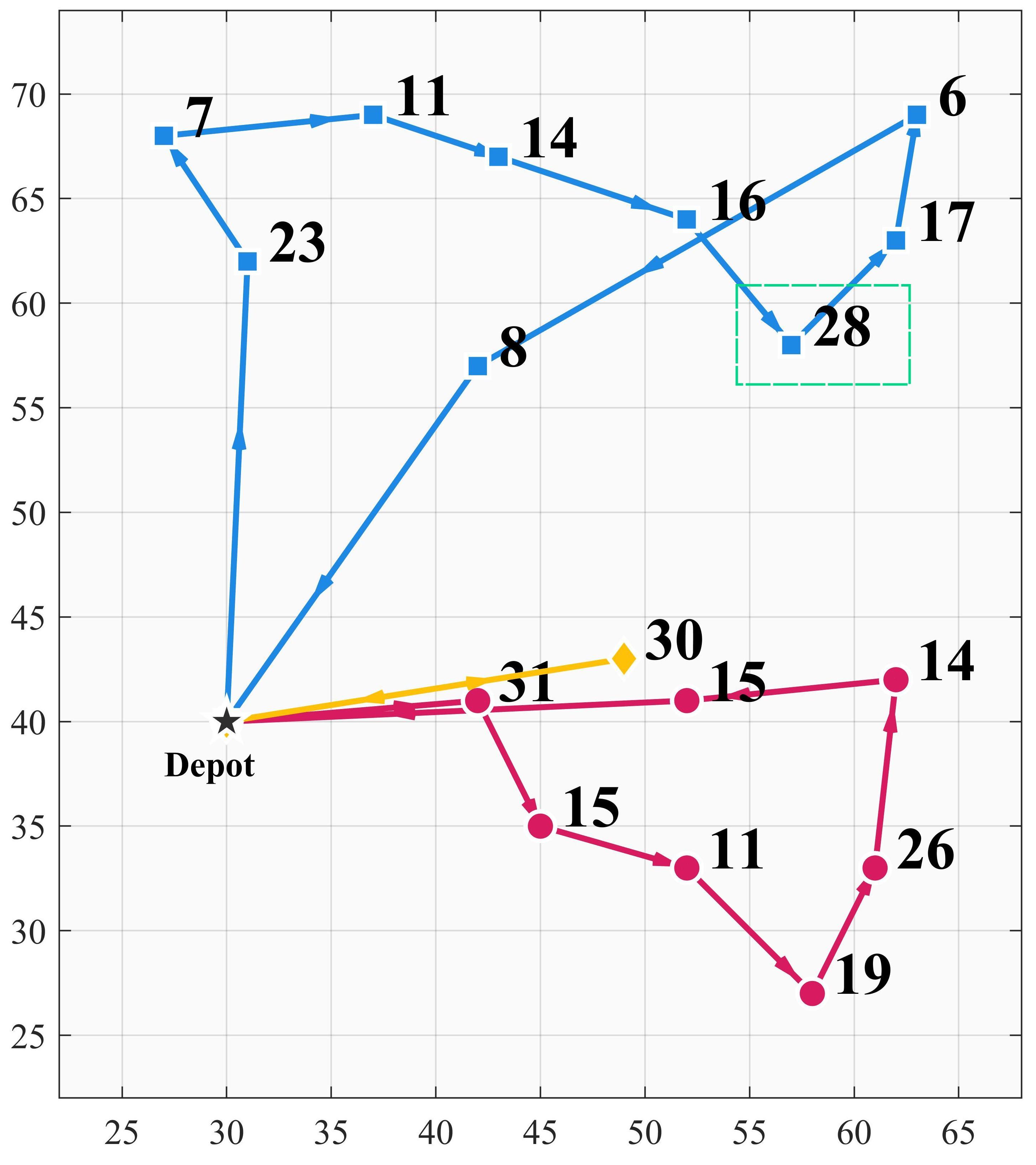}
    }
    \subfigure[HGSA]{
        \includegraphics[width=0.206\textwidth]{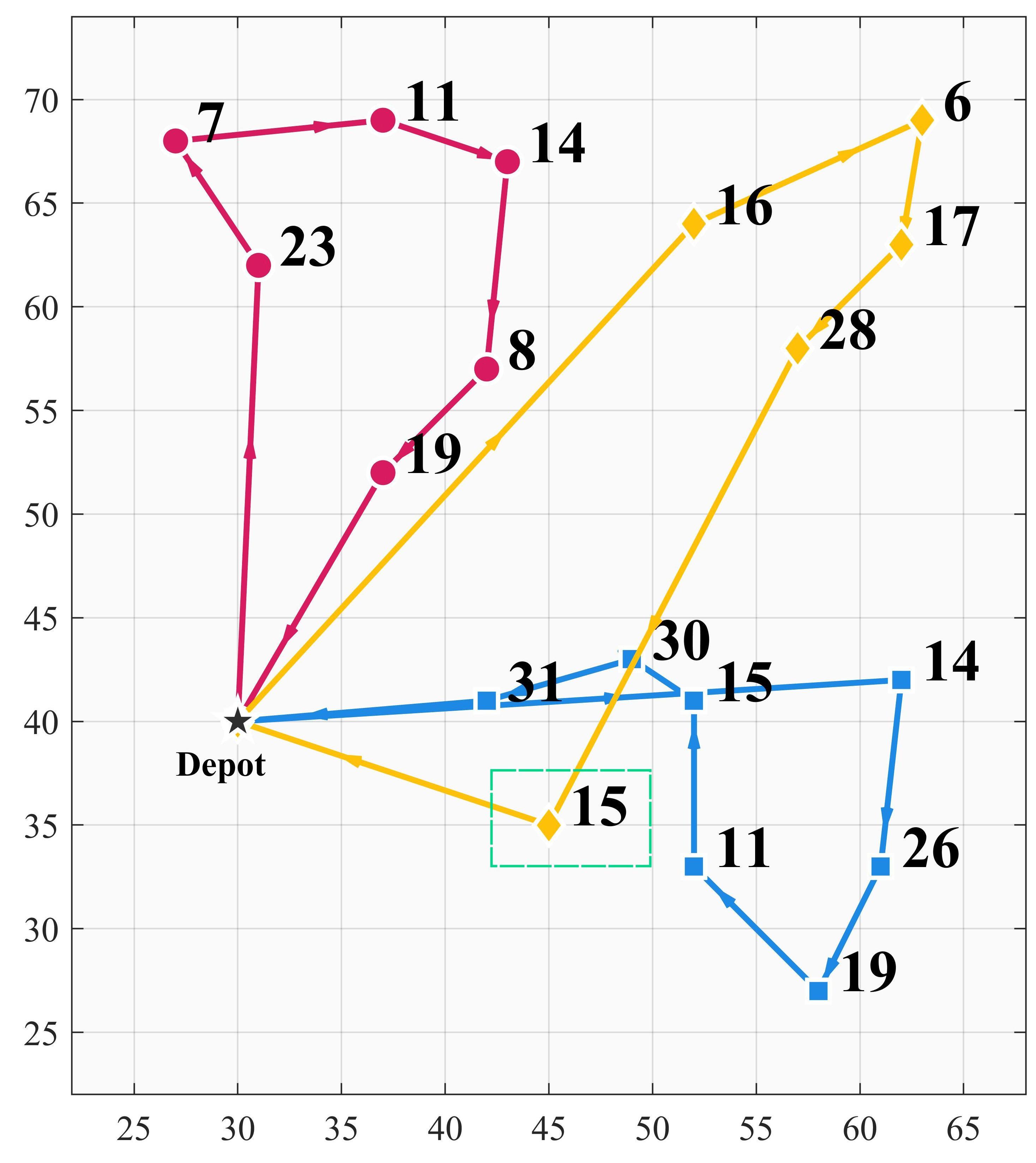}
        }
   \subfigure[\textcolor{blue}{HABC}]{
        \includegraphics[width=0.206\textwidth]{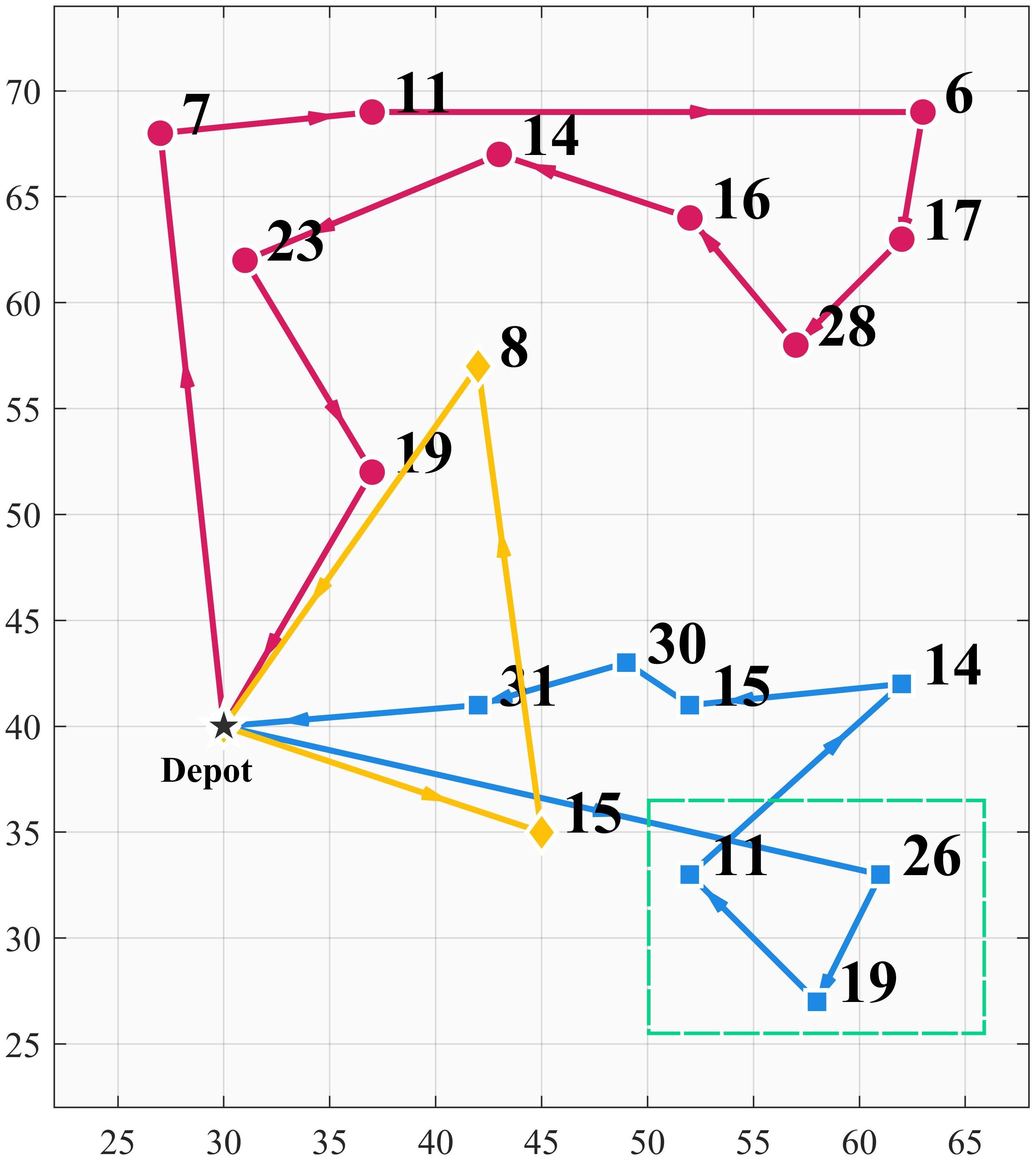}
    }     
    \subfigure[\textcolor{blue}{mGWOA}]{
        \includegraphics[width=0.206\textwidth]{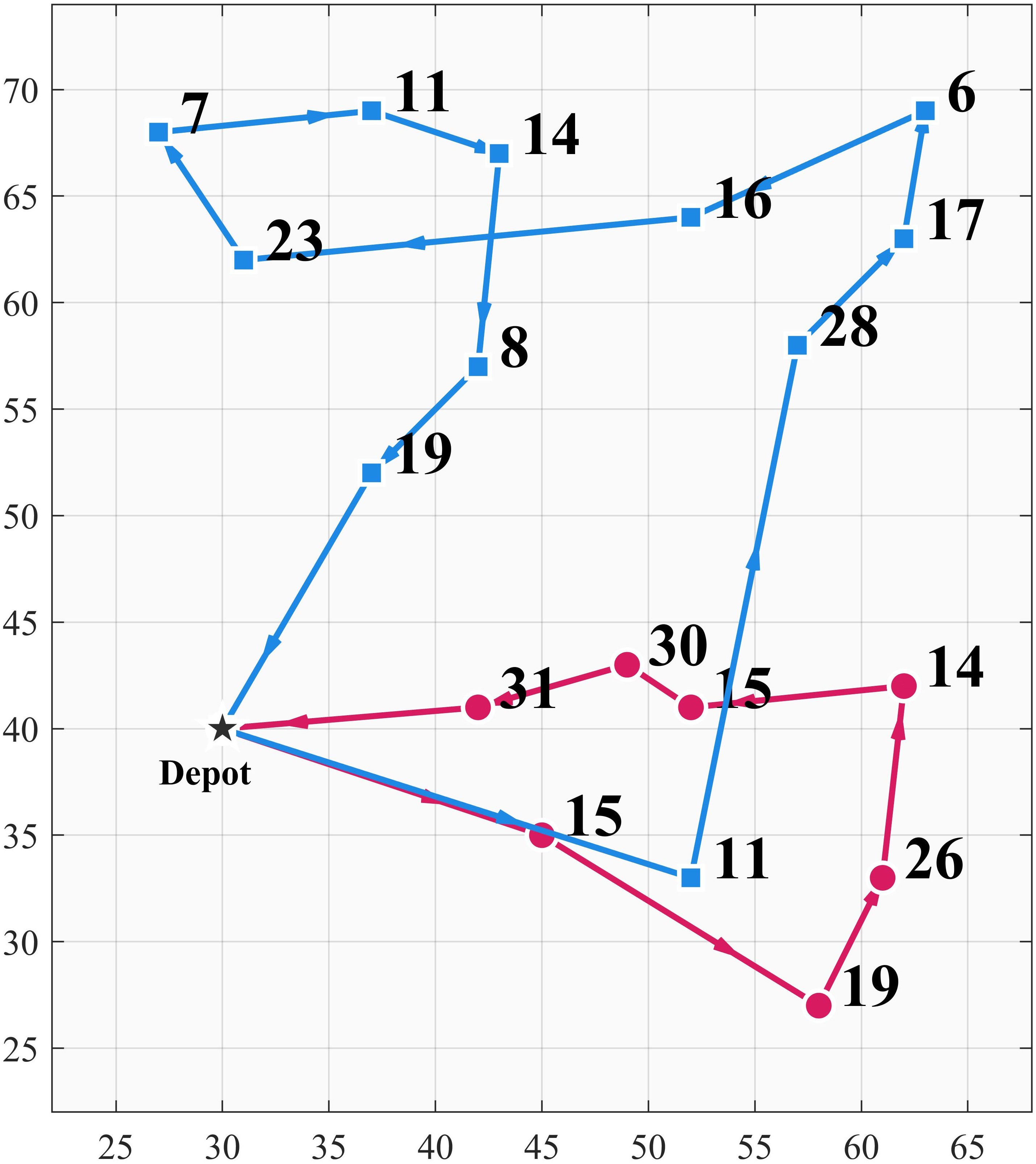}
    }
    \subfigure[MA]{
        \includegraphics[width=0.206\textwidth]{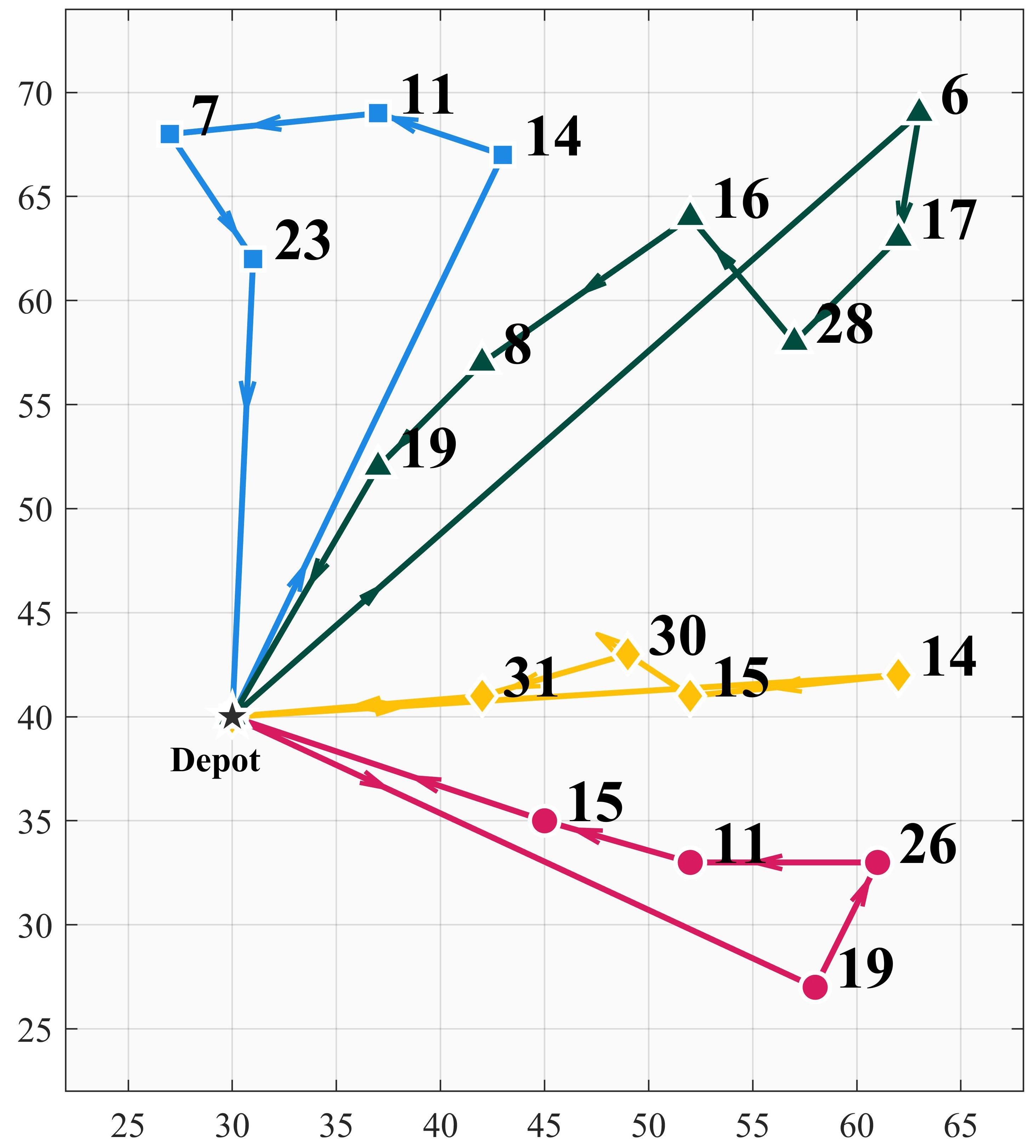}
    }
    \subfigure[MILP\_N]{
        \includegraphics[width=0.206\textwidth]{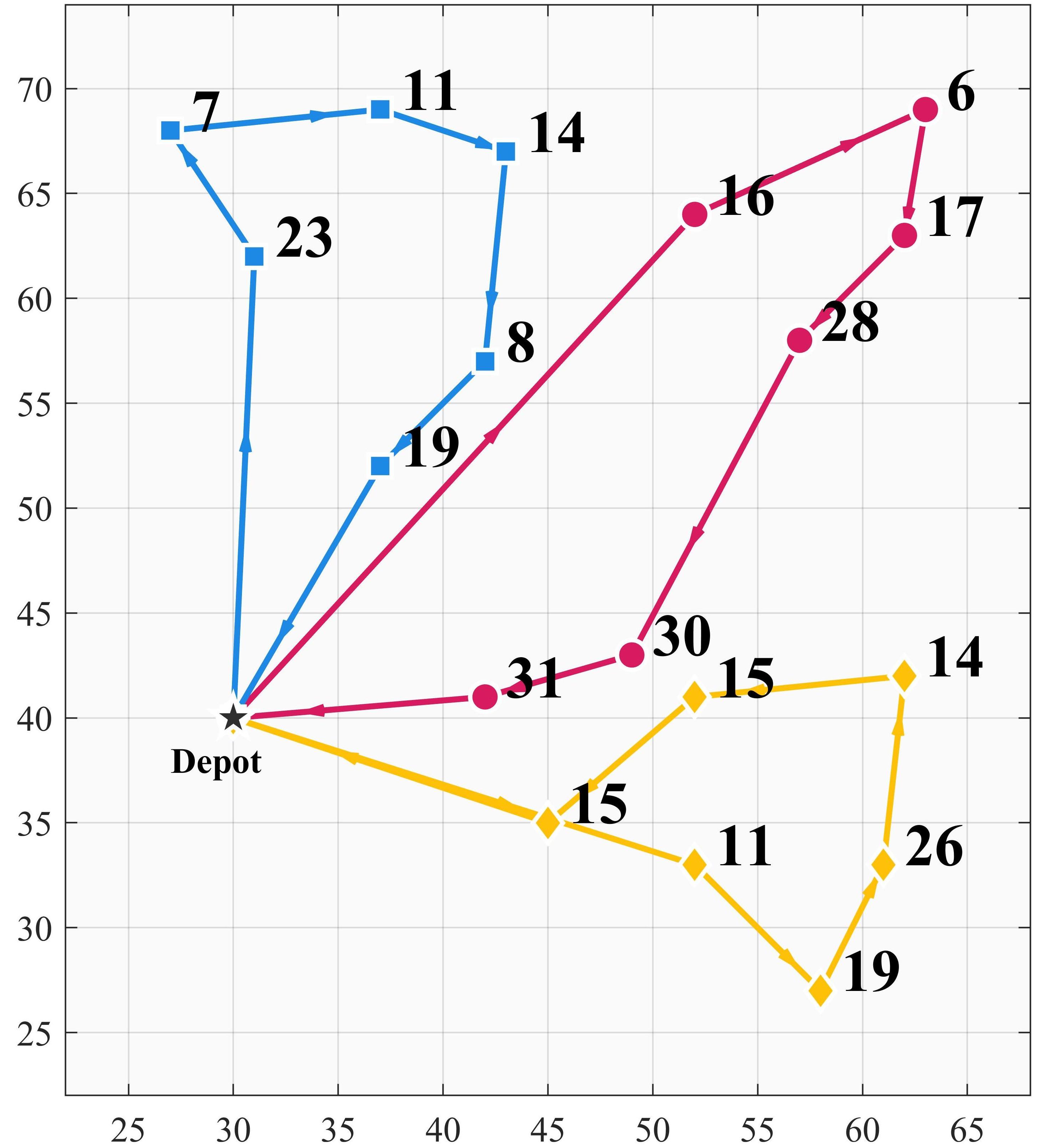}
    }
    
    \subfigure[MILP\_7200]{
        \includegraphics[width=0.206\textwidth]{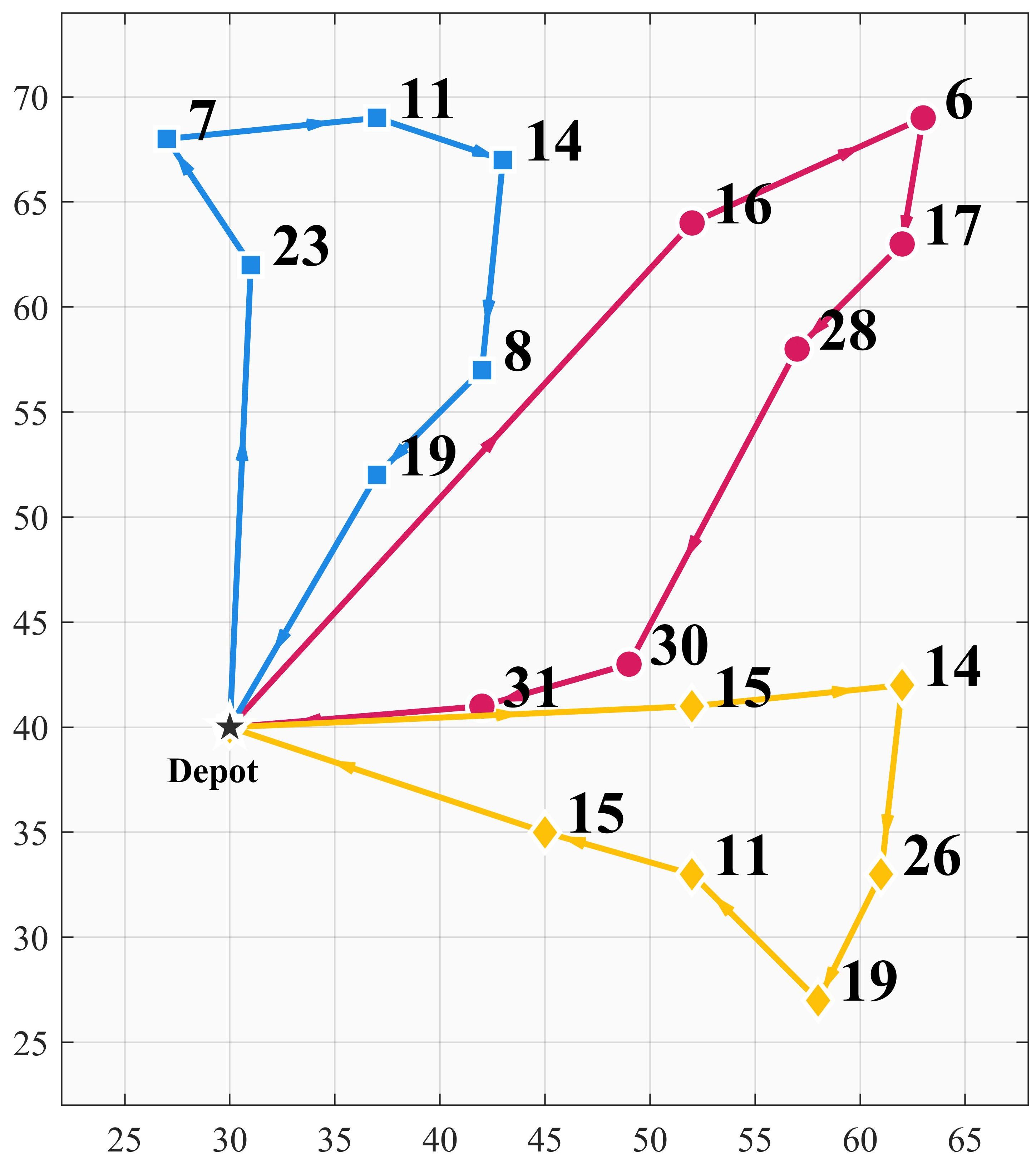}
    }
    \subfigure[AEDGA]{
        \includegraphics[width=0.206\textwidth]{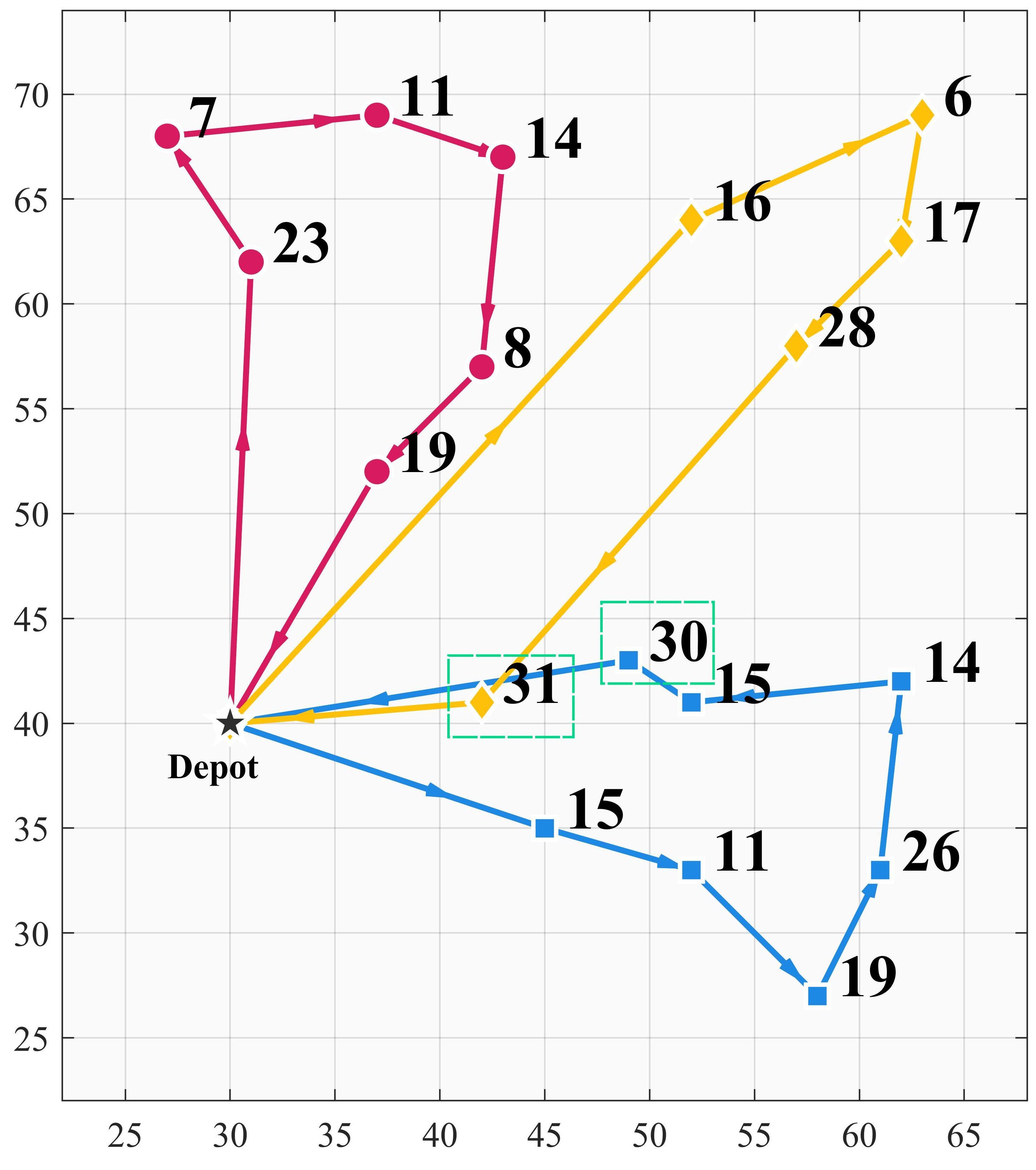}
    }
    
    \caption{\textcolor{blue}{Results of case study}}
    \label{fig:case study}
\end{figure}

In this subsection, we select test problem $2$ on route construction model as a representative case study. Fig.~\ref{fig:case study} illustrates the task allocation patterns of optimal solutions obtained by different algorithms. Each route is represented by a distinct color. Different symbols denote tasks on different routes, and the production volume of each task point is explicitly labeled.

The results reveal significant differences among algorithms. ETSA generates notably chaotic routes, demonstrating its limitations in handling complex problems. While GA-NN, GSACO, HGSA, \textcolor{blue}{HABC, mGWOA, }and MA produce more structured routes, they exhibit certain inefficiencies: \textcolor{blue}{GSACO, HGSA, and HABC show evident detours (highlighted in green boxes)}, with GSACO's priority on high-yield tasks leading to unnecessary energy consumption. GA-NN and MA underutilize robot load capacity, resulting in excessive route generation. \textcolor{blue}{In contrast, mGWOA overly prioritizes maximizing robot loads, which compromises the efficiency of individual trips.} MILP\_N and MILP\_7200 yield identical task allocations, differing only in task execution sequences, demonstrating MILP's advantage in small-scale problem solving. After 14,266 seconds of computation, MILP confirms that MILP\_7200 indeed achieves the optimal solution.

AEDGA demonstrates clear, logically structured route planning. Its unique strategy prioritizes the optimization of tasks distant from the depot while balancing high-yield tasks near the depot across multiple routes (highlighted in green boxes), positioning them at the end of execution sequences. This approach effectively minimizes additional energy consumption from excessive loads, maintaining errors within acceptable ranges. Although this strategy leads to local optima in this instance, it reflects the algorithm's comprehensive consideration of spatial distribution and load balancing.

\subsubsection{Results analysis}
Based on the algorithmic performance in the above experiments, we provide the following comprehensive analysis:

GA-NN employs greedy path adjustments, leveraging its computational efficiency and implementation simplicity to quickly identify local optima under limited computational resources. However, its \textcolor{blue}{local search architecture is fundamentally flawed due to its rigid, decoupled two-stage process. The nearest neighbor heuristic greedily constructs routes by always selecting the closest unvisited task point, a method that completely ignores the global route structure and often leads to inefficient, tangled paths. Consequently, the GA's strategic partitioning is guided by highly unreliable and misleading cost feedback.}

ETSA maintains convergence and enhances search efficiency through parameter adjustment and local search. \textcolor{blue}{However, its innovation focuses on the acceptance strategy rather than the local search operator itself, and the efficiency of this new strategy heavily depends on density function fitting accuracy and parameter optimization.} Moreover, solution quality improvements through parameter adjustments alone cannot overcome SA's inherent limitations of local optima entrapment and initial solution sensitivity, potentially facing dual challenges of computational efficiency and fitting accuracy in large-scale instances.

\textcolor{blue}{GSACO's local search is implicitly guided by the pheromone matrix, which is updated based on an adaptive greedy strategy. Its reliance on greedy strategies may suppress exploration through over-exploitation, while its adaptive adjustment mechanism increases computational complexity.} Its performance largely depends on balancing greedy strategies with pheromone updates. Additionally, ACO's inherent computational complexity in calculating task correlations for each solution update significantly constrains its efficiency in large-scale problem solving.

\textcolor{blue}{HGSA implements a superficial hybridization, where GA's crossover and SA's local search operate largely in sequence rather than in synergy. The local search (SA) is applied as a refinement operator to solutions generated by the GA, meaning the global exploration phase (GA) is not immediately informed by the fine-grained landscape information discovered by the local search. This one-way flow of information is inefficient and fails to create a truly co-evolutionary process, limiting the algorithm's ability to effectively balance its explorative and exploitative tendencies.}

\textcolor{blue}{HABC's local search framework is undermined by its flawed mechanism for escaping local optima. It integrates GA's evolutionary operators into the ABC framework, but its mechanism for escaping local optima is a significant weakness. The scout bee phase, triggered when a solution fails to improve, may completely discards the stagnated solution and replaces it with a new random one. This acts as a disruptive reset that throws away potentially valuable structural information learned during the local search. This lack of a more intelligent diversification strategy limits its ability to perform nuanced exploration in complex problem landscapes.}

\textcolor{blue}{mGWOA's partitioned population strategy creates a structural imbalance in its local search. One population segment, inspired by GWO, exerts strong and potentially premature convergence pressure by aggressively targeting the elite solutions, which can rapidly diminish diversity. The other segment, guided by WOA, alternates between exploitation and exploration but lacks full synergy with the first. This dominant exploitative tendency is often not sufficiently counteracted by the terminal OBL phase, resulting in an architectural imbalance that hinders its ability to consistently find high-quality solutions, particularly as problem complexity increases.}

MA, despite employing a multi-level optimization framework, incurs substantial computational overhead by performing local search on all offspring solutions, easily converging to local optima under limited computational resources. Its excessive reliance on local search leads to premature convergence and insufficient adaptability to complex objective functions, limiting its practical effectiveness. These issues highlight the need for better balance among global search, local optimization, solution quality, diversity maintenance, and algorithmic efficiency in hybrid optimization algorithm design.

In summary, while existing heuristic methods incorporate local search through algorithm hybridization or local operator addition, attempting to maintain exploration-exploitation balance through mere parameter adjustment, inappropriate local search target selection or applying greedy search to all paths is prone to result in redundant and time-consuming operations. This significantly reduces search efficiency and may guide the population toward local optima. \textcolor{blue}{In contrast, AEDGA systematically addresses these deficiencies through its key components, as validated by our ablation study and comparative experiments: the EASS dynamically allocates search resources to high-potential individuals, while the CLSM further intensifies this search by optimizing the routes within these individuals that offer the greatest potential for improvement. This design philosophy of targeted and synergistic optimization allows AEDGA to achieve a more robust balance between exploration and exploitation, making it a more effective and scalable solution for complex routing problems.}

MILP requires substantial computational resources, imposing high demands on computing environments. MILP\_N shows limited scalability under equal runtime constraints. Although MILP\_7200 achieves superior solutions for small and medium-scale problems, such extended computation times may cause scheduling delays in practical applications. MILP's poor performance on large-scale problems further diminishes its practicality. In comparison, AEDGA demonstrates unique advantages in computational efficiency, stability, scalability, and practicality. Particularly in practical applications, AEDGA provides high-quality solutions within reasonable timeframes while maintaining excellent algorithmic stability and problem adaptability, making it an ideal choice for solving real-world VRP instances.

\subsection{Comparative experiments and analysis of route scheduling phase}
\label{results of RS}

The RS phase primarily focuses on allocating task paths, generated during the RG phase, to a limited number of robots within a constrained makespan. To address infeasible solutions, this study introduces three remediation frameworks, detailed in subsection \ref{rs framework}. These frameworks correspond to variants Fr$_1$, Fr$_2$, and Fr$_3$, respectively. As referenced in~\cite{cattaruzza2014memetic}, two time thresholds for task completion are established: $TH_1$ = $\frac{1.5 \times Z}{m}$ and $TH_2$ = $\frac{1.7 \times Z}{m}$, where $Z$ represents the mean performance achieved by AEDGA during the RG phase, and $m$ denotes the number of robots. With $m$ configured at $2, 5,$ and $8$, six distinct scenarios are constructed. It is worth noting that these threshold values should be adjusted according to specific real-world requirements; the current experimental settings serve solely to validate the framework's effectiveness rather than carrying any particular practical significance. For clarity, detailed comparative results are presented in the supplementary materials. The experimental results reveal an interesting pattern: as scenario complexity increases, the probability of finding feasible solutions decreases significantly for Fr$_2$ and Fr$_3$, while Fr$_1$ consistently maintains its ability to identify viable solutions across all scenarios. 

\begin{figure}[!hbpt]
    \centering
    \includegraphics[width=9cm]{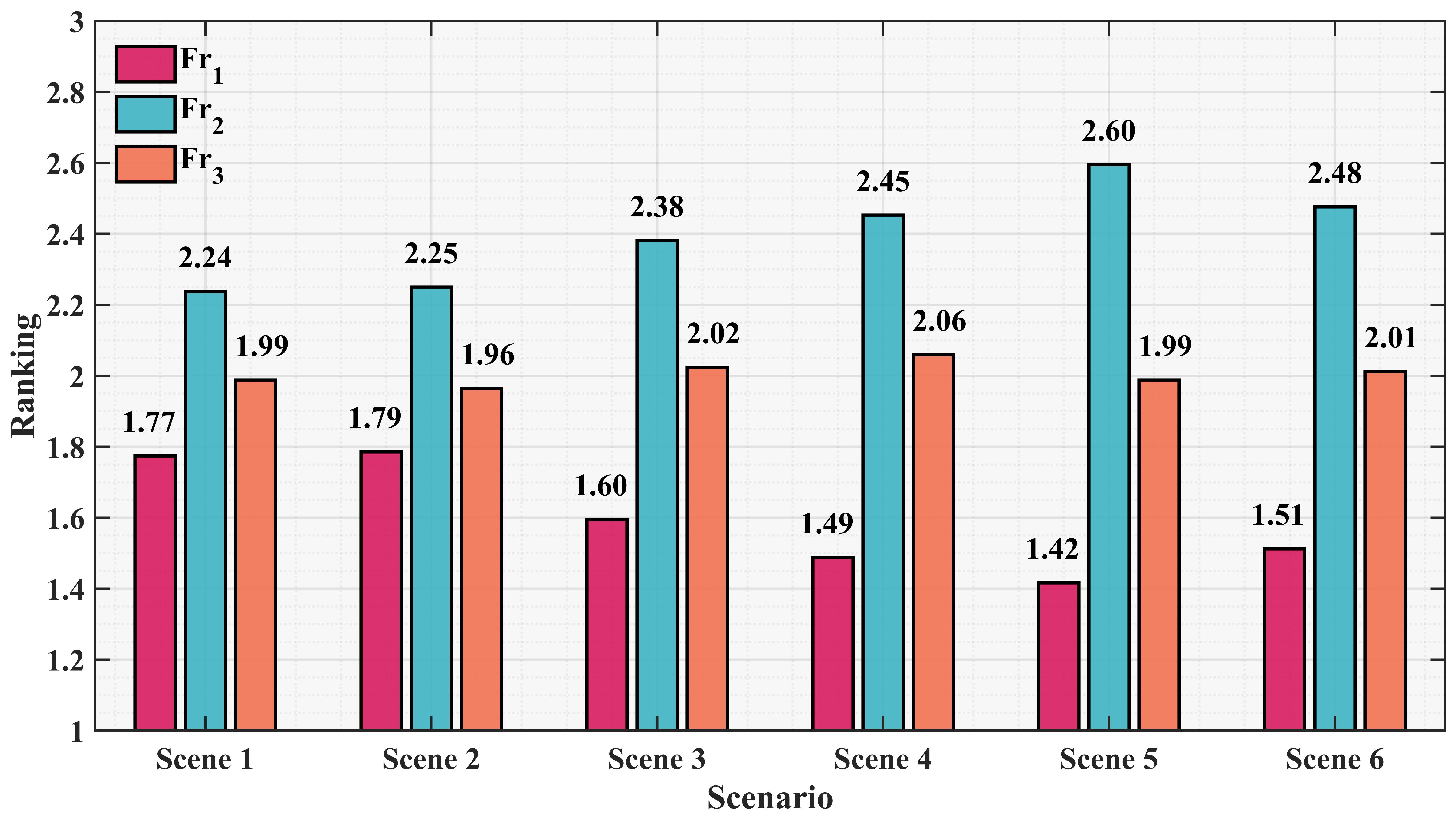}
    \caption{Friedman test results in $6$ scenarios}
    \label{Fried}
\end{figure}

The Friedman test~\cite{liang2021clustering} results across all scenarios, illustrated in Fig.~\ref{Fried}, clearly demonstrate Fr$_1$'s superior performance, consistently achieving the highest rankings across all test scenarios. This empirical evidence strongly suggests that the generational solution processing and infeasibility remediation approach is more compatible with AEDGA. Furthermore, the consistent lowest ranking of variant Fr$_2$ provides direct validation of the remediation procedure's effectiveness.

\section{Conclusion and future research}
\label{section 5}
This paper presents a comprehensive investigation into the MTPRTS problem. We first establish a mathematical model that incorporates the relationship between robot energy consumption, self-weight, and real-time load, while allowing robots to perform multiple trips within a specified makespan constraint. Based on this framework, the AEDGA is proposed as a solution approach.

AEDGA enhances solution quality through three key components: the ILBIM for generating high-quality initial solutions, the CLSM for reducing redundant local search operations, and the EASS for dynamically adjusting the probability of local search targets based on accumulated successful experiences. Additionally, a solution repair procedure is developed to ensure feasibility under makespan constraints, which is proved most effective when applied after each generation's population update. Comprehensive experiments conducted on $24$ existing test problems and $18$ proposed test instances demonstrate that AEDGA significantly outperforms existing heuristic algorithms in both solution quality and computational efficiency, particularly exhibiting superior scalability in medium to large-scale scenarios. Furthermore, compared to the proposed MILP approach, AEDGA demonstrates superior practicality in addressing real-world applications. While the problem-specific operators may limit direct application to other task allocation problems, the clustering-based local search mechanism and experience-based dynamic selection strategy provide valuable insights for solving related optimization challenges.

The innovative solutions presented in this research offer practical technical support for multi-robot coordination in smart orchards, contributing to improved harvesting efficiency and reduced operational costs. However, several limitations warrant consideration. First, the current model assumes static task conditions, whereas real-world orchard environments often involve dynamic factors such as weather conditions and merchants' demand variations. Second, the current model addresses the scheduling problem for a single type of robot. However, real-world smart farming environments may involve collaborative operations among heterogeneous robotic platforms to enhance harvesting efficiency. Third, practical applications often involve balancing multiple conflicting objectives, necessitating the development of trade-off solutions. \textcolor{blue}{Furthermore, our current model enforces the constraint that each task must be completed by a single robot. Relaxing this constraint to allow for task partitioning (i.e., a large task being divided and serviced by multiple robots) represents a vital and challenging avenue for future research to further enhance operational flexibility.} In the future, these advances will significantly contribute to bridging the gap between theoretical models and practical implementation in intelligent agricultural systems. This would ultimately facilitate the widespread adoption of autonomous agricultural systems, leading to improved operational efficiency and sustainable farming practices in modern agriculture.

\bibliographystyle{unsrt}
\bibliography{ref}

\end{document}


\begin{table*}[!hbpt]
\centering
\caption{Comparative results of AEDGA and MILP\_7200 on traditional CVRP}
\begin{tabular}{llllll}
\hline
\multicolumn{1}{c}{\multirow{2}{*}{$Pro$}} & \multicolumn{1}{c}{\multirow{2}{*}{$n$}} & \multicolumn{2}{c}{Traditional CVRP}                                  & \multicolumn{2}{c}{RG}                                               \\ \cline{3-6} 
\multicolumn{1}{c}{}                       & \multicolumn{1}{c}{}                     & \multicolumn{1}{c}{MILP\_7200}       & \multicolumn{1}{c}{AEDGA}      & \multicolumn{1}{c}{MILP\_7200}      & \multicolumn{1}{c}{AEDGA}      \\ \hline
1                                          & 16                                       & \textbf{4.5134e+02 (0.00e+00) +}     & 4.6976e+02 (5.29e+00)          & \textbf{1.1374e+04 (0.00e+00) +}    & 1.1468e+04 (1.40e+02)          \\
2                                          & 19                                       & \textbf{2.1266e+02 (0.00e+00) +}     & 2.4354e+02 (1.21e+01)          & \textbf{2.3805e+04 (1.03e+01) +}    & 2.4742e+04 (1.97e+02)          \\
3                                          & 20                                       & \textbf{2.1742e+02 (0.00e+00) +}     & 2.3910e+02 (1.77e+01)          & \textbf{2.4191e+04 (1.68e+01) +}    & 2.5103e+04 (2.87e+02)          \\
4                                          & 21                                       & \textbf{2.1271e+02 (0.00e+00) +}     & 2.1510e+02 (2.52e+00)          & \textbf{2.4229e+04 (2.24e+01) +}    & 2.4983e+04 (1.32e+02)          \\
5                                          & 22                                       & \textbf{2.1785e+02 (4.60e+00) +}     & 2.4354e+02 (2.11e+01)          & \textbf{2.5129e+04 (2.35e+01) +}    & 2.6165e+04 (1.34e+02)          \\
6                                          & 22                                       & \textbf{5.8879e+02 (0.00e+00) +}     & 5.9038e+02 (5.18e-01)          & \textbf{1.3024e+06 (0.00e+00) +}    & 1.3032e+06 (5.56e+02)          \\
7                                          & 23                                       & \textbf{5.3117e+02 (0.00e+00) +}     & 5.5407e+02 (4.97e+00)          & \textbf{1.5601e+04 (0.00e+00) +}    & 1.5776e+04 (7.14e+01)          \\
8                                          & 40                                       & \textbf{4.6173e+02 (3.16e+00) +}     & 4.9743e+02 (1.38e+01)          & \textbf{4.4778e+04 (1.29e+01) +}    & 4.9092e+04 (1.43e+03)          \\
9                                          & 45                                       & \textbf{5.1279e+02 (4.24e+00) +}     & 5.4869e+02 (2.36e+01)          & \textbf{5.2797e+04 (1.73e+01) +}    & 5.7617e+04 (8.73e+02)          \\
10                                         & 50                                       & \textbf{7.1358e+02 (8.04e+00) +}     & 7.4561e+02 (1.27e+01)          & \textbf{4.9575e+04 (1.60e+00) +}    & 5.3261e+04 (7.10e+02)          \\
11                                         & 50                                       & \textbf{5.6015e+02 (5.61e+00) +}     & 5.9921e+02 (6.90e+00)          & \textbf{5.8587e+04 (6.92e+00) +}    & 6.4077e+04 (9.89e+02)          \\
12                                         & 50                                       & \textbf{6.3905e+02 (7.92e+00) +}     & 6.7048e+02 (5.24e+00)          & \textbf{5.3092e+04 (2.79e+00) +}    & 5.8313e+04 (7.82e+02)          \\
13                                         & 51                                       & \textbf{7.5001e+02 (6.77e+00) +}     & 7.7610e+02 (1.07e-13)          & \textbf{4.2739e+04 (4.98e+00) +}    & 4.4856e+04 (3.79e+02)          \\
14                                         & 55                                       & \textbf{7.0252e+02 (7.42e+00) +}     & 7.2990e+02 (7.36e+00)          & \textbf{5.6065e+04 (4.16e+00) +}    & 6.1412e+04 (6.60e+02)          \\
15                                         & 55                                       & \textbf{9.5184e+02 (5.55e+00) +}     & 9.8408e+02 (8.47e-14)          & \textbf{4.8251e+04 (9.50e-01) +}    & 5.0897e+04 (6.31e+02)          \\
16                                         & 55                                       & \textbf{5.8164e+02 (1.03e+01) +}     & 6.2343e+02 (4.80e+00)          & \textbf{6.8016e+04 (1.71e+01) +}    & 7.2594e+04 (1.42e+03)          \\
17                                         & 55                                       & \textbf{5.8954e+02 (9.37e+00) +}     & 6.2666e+02 (3.03e+00)          & \textbf{6.5444e+04 (1.30e+01) +}    & 7.0243e+04 (1.51e+03)          \\
18                                         & 60                                       & \textbf{7.5305e+02 (9.81e+00) +}     & 7.8696e+02 (4.58e+00)          & \textbf{6.3129e+04 (5.83e+00) +}    & 6.8584e+04 (5.32e+02)          \\
19                                         & 60                                       & \textbf{9.8100e+02 (8.21e+00) +}     & 1.0055e+03 (0.00e+00)          & \textbf{5.6404e+04 (2.71e+00) +}    & 5.8525e+04 (3.97e+02)          \\
20                                         & 65                                       & \textbf{8.0329e+02 (7.94e+00) +}     & 8.3948e+02 (2.40e-13)          & \textbf{7.3838e+04 (8.44e+00) +}    & 7.9711e+04 (5.01e+02)          \\
21                                         & 70                                       & \textbf{8.4820e+02 (1.06e+01) +}     & 9.0515e+02 (3.84e+01)          & \textbf{7.9861e+04 (1.04e+01) +}    & 8.6541e+04 (5.07e+02)          \\
22                                         & 76                                       & \textbf{6.0762e+02 (4.85e+00) +}     & 6.6965e+02 (2.03e+01)          & \textbf{1.4060e+05 (2.73e+01) +}    & 1.5413e+05 (2.57e+03)          \\
23                                         & 76                                       & \textbf{6.5884e+02 (1.25e+01) +}     & 7.0261e+02 (1.86e+01)          & \textbf{1.2147e+05 (2.53e+01) +}    & 1.3616e+05 (2.58e+03)          \\
24                                         & 101                                      & \textbf{6.9129e+02 (7.80e-01) +}     & 7.7281e+02 (1.90e+01)          & \textbf{1.8119e+05 (2.11e+01) +}    & 1.9984e+05 (3.05e+02)          \\
25                                         & 41                                       & \textbf{2.4449e+02 (6.81e+00) +}     & 2.4659e+02 (1.21e+00)          & \textbf{5.2634e+04 (2.59e+00) +}    & 5.4489e+04 (5.30e+02)          \\
26                                         & 61                                       & \textbf{3.8096e+02 (9.55e+00) +}     & 3.8948e+02 (3.69e-01)          & \textbf{8.5554e+04 (3.60e+00) +}    & 8.8930e+04 (1.02e+03)          \\
27                                         & 81                                       & \textbf{5.9100e+02 (1.34e+01) +}     & 6.0564e+02 (1.20e-13)          & \textbf{1.3297e+05 (2.98e+00) +}    & 1.3989e+05 (1.26e+03)          \\
28                                         & 91                                       & \textbf{7.6969e+02 (1.97e+01) +}     & 7.9731e+02 (4.89e+00)          & \textbf{1.7343e+05 (2.88e+00) +}    & 1.8248e+05 (1.33e+03)          \\
29                                         & 131                                      & 1.2551e+03 (2.08e+01) -              & \textbf{1.2424e+03 (3.38e+00)} & \textbf{2.8624e+05 (5.97e+00) +}    & 2.9288e+05 (7.42e+02)          \\
30                                         & 181                                      & 1.5688e+03 (2.37e+01) -              & \textbf{1.5062e+03 (7.57e+00)} & \textbf{3.4989e+05 (9.19e+00) +}    & 3.5413e+05 (1.13e+03)          \\
31                                         & 161                                      & 2.3804e+03 (4.71e+01) -              & \textbf{2.2346e+03 (8.75e+00)} & \textbf{5.2445e+05 (6.00e+00) +}    & 5.3105e+05 (1.56e+03)          \\
32                                         & 251                                      & 3.3627e+03 (3.67e+01) -              & \textbf{3.2988e+03 (1.22e+01)} & \textbf{7.7779e+05 (9.37e+00) +}    & 7.7968e+05 (1.06e+03)          \\
33                                         & 321                                      & 3.8895e+03 (3.78e+01) -              & \textbf{3.7189e+03 (2.91e+01)} & 8.8054e+05 (1.36e+01) -             & \textbf{8.4926e+05 (1.01e+03)} \\
34                                         & 251                                      & 3.5274e+03 (2.71e+01) -              & \textbf{3.3992e+03 (9.59e-13)} & \textbf{7.9291e+05 (9.66e+00) +}    & 7.9701e+05 (1.92e+03)          \\
35                                         & 376                                      & 6.5743e+03 (8.50e+01) -              & \textbf{5.7280e+03 (0.00e+00)} & 1.4026e+06 (3.36e+01) -             & \textbf{1.3380e+06 (9.72e+02)} \\
36                                         & 501                                      & 8.2082e+03 (8.76e+01) -              & \textbf{6.7168e+03 (3.34e+01)} & 1.6988e+06 (4.30e+01) -             & \textbf{1.5570e+06 (5.35e+02)} \\
37                                         & 361                                      & 8.0877e+03 (8.16e+01) -              & \textbf{7.1348e+03 (2.76e+01)} & 1.8990e+06 (1.92e+01) -             & \textbf{1.7007e+06 (6.54e+02)} \\
38                                         & 541                                      & 1.1532e+04 (8.94e+01) -              & \textbf{1.0586e+04 (4.81e+01)} & 2.7571e+06 (9.32e+01) -             & \textbf{2.5193e+06 (6.96e+02)} \\
39                                         & 721                                      & 1.5571e+04 (9.06e+01) -              & \textbf{1.1313e+04 (4.83e+01)} & 3.6494e+06 (9.37e+01) -             & \textbf{2.6437e+06 (6.68e+02)} \\
40                                         & 491                                      & 1.1033e+04 (8.89e+01) -              & \textbf{9.5738e+03 (4.20e+01)} & 2.9421e+06 (7.94e+01) -             & \textbf{2.2735e+06 (5.38e+02)} \\
41                                         & 736                                      & 2.2046e+04 (9.27e+01) -              & \textbf{1.6167e+04 (6.46e+01)} & 4.6392e+06 (9.46e+01) -             & \textbf{3.7416e+06 (4.48e+02)} \\
42                                         & 981                                      & 3.1040e+04 (9.36e+01) -              & \textbf{1.8259e+04 (0.00e+00)} & 6.6844e+06 (9.55e+01) -             & \textbf{4.1775e+06 (1.06e+03)} \\ \hline
\multicolumn{2}{c}{+/-/=}                                                             & \multicolumn{1}{c}{\textbf{28/14/0}} &                                & \multicolumn{1}{c}{\textbf{33/9/0}} &                                \\ \hline
\end{tabular}
\end{table*}

\begin{table*}[!hbpt]
\centering
\caption{Results of $TH_1$ with $r = 2$}
\label{Comparative results on traditional CVRP}
\setlength{\tabcolsep}{6.5pt}
\centering\small
\begin{tabular}{lllllllll}
\hline
Pro   & AEDGA1                  & AEDGA2                  & AEDGA3                \\
\hline
1     & 4.8824e+04 (2.05e+04) + & 1.9973e+05 (4.73e+02) + & 1.9994e+05 (0.00e+00) \\
2     & 1.1771e+04 (1.67e+02) - & 1.1954e+04 (1.55e+02) - & 1.1374e+04 (0.00e+00) \\
3     & 1.2899e+04 (1.48e+03) + & 2.5193e+04 (1.45e+02) - & 2.1819e+04 (4.08e+03) \\
4     & 1.2145e+04 (1.01e+03) + & 2.5543e+04 (3.58e+02) - & 2.1658e+04 (3.72e+03) \\
5     & 1.2718e+04 (1.68e+03) + & 2.5816e+04 (5.81e+02) = & 2.2842e+04 (4.12e+03) \\
6     & 1.0026e+04 (2.43e+03) + & 2.6892e+04 (1.99e+02) - & 2.5728e+04 (4.65e+02) \\
7     & 1.3311e+06 (9.22e+03) - & 1.3373e+06 (7.77e+03) - & 1.3058e+06 (5.89e+03) \\
8     & 1.6415e+04 (2.86e+02) - & 1.6426e+04 (3.96e+02) - & 1.5753e+04 (1.18e+02) \\
9     & 1.8670e+04 (3.93e+03) + & 5.1895e+04 (1.09e+03) - & 4.7352e+04 (1.06e+03) \\
10    & 1.6967e+04 (3.35e+03) + & 6.0809e+04 (4.82e+02) - & 5.7412e+04 (9.78e+02) \\
11    & 5.7743e+04 (4.45e+02) - & 5.7680e+04 (4.74e+02) - & 5.3215e+04 (6.54e+02) \\
12    & 6.7158e+04 (2.80e+03) = & 6.8324e+04 (9.43e+02) - & 6.3840e+04 (3.67e+02) \\
13    & 5.9389e+04 (6.30e+02) = & 6.0775e+04 (5.76e+02) - & 5.8360e+04 (1.11e+03) \\
14    & 4.6621e+04 (4.14e+02) - & 4.6483e+04 (3.69e+02) - & 4.4944e+04 (5.02e+02) \\
15    & 6.3244e+04 (3.68e+02) - & 6.3530e+04 (6.67e+02) - & 6.0542e+04 (7.58e+02) \\
16    & 5.1863e+04 (3.03e+02) - & 5.2537e+04 (0.00e+00) - & 5.1021e+04 (6.34e+02) \\
17    & 7.5252e+04 (3.14e+02) = & 7.5139e+04 (4.59e+02) = & 7.3715e+04 (2.12e+03) \\
18    & 7.3295e+04 (6.17e+02) - & 7.2951e+04 (8.59e+02) - & 7.1007e+04 (1.46e+03) \\
19    & 7.1791e+04 (9.26e+02) - & 7.2539e+04 (6.42e+02) - & 6.9391e+04 (3.90e+02) \\
20    & 6.2095e+04 (0.00e+00) - & 6.1803e+04 (5.17e+02) - & 5.8649e+04 (4.39e+02) \\
21    & 8.2356e+04 (1.02e+03) - & 8.2614e+04 (5.59e+02) - & 8.0723e+04 (4.05e+02) \\
22    & 9.2385e+04 (1.98e+03) - & 9.2085e+04 (1.30e+03) - & 8.7297e+04 (9.05e+02) \\
23    & 3.4484e+04 (1.03e+04) + & 1.5496e+05 (1.23e+03) + & 1.5535e+05 (3.66e+02) \\
24    & 2.4325e+04 (1.23e+04) + & 1.3659e+05 (9.76e+02) + & 1.3702e+05 (1.54e+01) \\
25    & 5.8340e+04 (6.31e+02) - & 5.9018e+04 (7.24e+02) - & 5.4731e+04 (6.81e+02) \\
26    & 1.4472e+06 (7.88e+05) + & Inf (     NaN) =        & Inf (     NaN)        \\
27    & 6.3001e+04 (1.29e+04) + & Inf (     NaN) -        & 7.7524e+04 (7.79e+02) \\
28    & 5.6390e+04 (2.47e+03) = & Inf (     NaN) =        & Inf (     NaN)        \\
29    & 6.2514e+04 (8.89e+03) + & Inf (     NaN) =        & Inf (     NaN)        \\
30    & 4.5936e+06 (2.53e+06) + & Inf (     NaN) =        & Inf (     NaN)        \\
31    & 1.5250e+05 (5.76e+04) + & 2.6457e+06 (5.87e+02) = & 2.6446e+06 (9.98e+02) \\
32    & 5.2100e+06 (7.86e+04) + & Inf (     NaN) =        & Inf (     NaN)        \\
33    & 1.0041e+05 (3.26e+04) = & Inf (     NaN) -        & 1.3536e+05 (4.06e+02) \\
34    & 8.9483e+04 (4.26e+03) + & Inf (     NaN) =        & Inf (     NaN)        \\
35    & 1.7479e+04 (4.61e+03) + & 9.1701e+04 (4.88e+02) - & 8.9053e+04 (3.79e+02) \\
36    & 2.4721e+04 (8.37e+03) + & 1.4163e+05 (1.76e+02) - & 1.4002e+05 (1.09e+03) \\
37    & 1.9638e+04 (6.09e+03) + & 1.8915e+05 (2.12e+02) - & 1.2233e+05 (6.24e+03) \\
38    & 2.1515e+04 (4.02e+03) + & Inf (     NaN) -        & 5.2084e+04 (1.49e+04) \\
39    & 3.6243e+04 (2.15e+04) = & Inf (     NaN) -        & 6.3574e+04 (2.86e+03) \\
40    & 2.5227e+04 (0.00e+00) = & Inf (     NaN) -        & 2.5058e+04 (4.72e+02) \\
41    & 7.9710e+04 (1.50e+04) = & Inf (     NaN) -        & 9.4994e+04 (2.37e+03) \\
42    & 6.3604e+04 (1.83e+04) = & Inf (     NaN) -        & 5.5086e+04 (4.71e+02) \\
\hline
+/-/= & 20/13/9                 & 4/30/18                 & \multicolumn{1}{l}{} \\
\hline
\end{tabular}

\end{table*}

\begin{table*}[!hbpt]
\centering
\caption{Results of $TH_2$ with $r = 2$}
\setlength{\tabcolsep}{6.5pt}
\centering\small
\begin{tabular}{lllllllll}
\hline
Pro   & AEDGA1                  & AEDGA2                  & AEDGA3                \\
\hline
1     & 4.4418e+04 (1.75e+04) + & 1.9990e+05 (9.25e+01) - & 1.9935e+05 (1.31e+03) \\
2     & 1.1929e+04 (1.28e+02) - & 1.1994e+04 (2.52e+02) - & 1.1432e+04 (1.29e+02) \\
3     & 1.4536e+04 (8.02e+02) + & 2.6315e+04 (1.07e+03) = & 2.4866e+04 (1.76e+02) \\
4     & 1.2952e+04 (3.05e+03) + & 2.5910e+04 (3.76e+02) - & 2.4232e+04 (1.95e+03) \\
5     & 1.1596e+04 (3.22e+03) + & 2.6552e+04 (7.51e+02) - & 2.5140e+04 (2.91e+02) \\
6     & 1.3271e+04 (1.37e+03) + & 2.8324e+04 (4.70e+02) - & 2.6520e+04 (5.56e+02) \\
7     & 1.3878e+06 (1.81e+04) - & 1.3992e+06 (4.09e+03) - & 1.3089e+06 (6.22e+03) \\
8     & 1.6885e+04 (0.00e+00) - & 1.6885e+04 (0.00e+00) - & 1.5799e+04 (8.05e+01) \\
9     & 4.2102e+04 (7.57e+03) = & 5.3020e+04 (0.00e+00) - & 4.8491e+04 (1.93e+03) \\
10    & 3.2119e+04 (1.24e+04) + & 6.1202e+04 (2.06e+02) - & 5.8473e+04 (7.61e+02) \\
11    & 5.8003e+04 (8.13e-12) - & 5.7919e+04 (1.17e+02) - & 5.4143e+04 (2.61e+02) \\
12    & 6.9006e+04 (0.00e+00) - & 6.9006e+04 (0.00e+00) - & 6.4475e+04 (9.89e+02) \\
13    & 6.0028e+04 (5.83e+02) = & 6.0947e+04 (1.90e+02) = & 5.9356e+04 (1.03e+03) \\
14    & 4.7047e+04 (2.75e+02) - & 4.7133e+04 (1.79e+02) - & 4.5131e+04 (2.60e+02) \\
15    & 6.4092e+04 (6.38e+02) - & 6.4562e+04 (3.54e+02) - & 6.1331e+04 (6.25e+02) \\
16    & 5.2174e+04 (1.33e+02) - & 5.2537e+04 (0.00e+00) - & 5.1206e+04 (4.54e+02) \\
17    & 7.4443e+04 (2.12e+03) = & 7.5222e+04 (2.99e+02) = & 7.4521e+04 (1.18e+03) \\
18    & 7.3571e+04 (0.00e+00) - & 7.3321e+04 (3.97e+02) - & 7.1841e+04 (1.15e+03) \\
19    & 7.3832e+04 (5.37e+02) - & 7.5329e+04 (3.78e+02) - & 6.9987e+04 (6.28e+02) \\
20    & 6.1889e+04 (2.85e+02) - & 6.2065e+04 (6.69e+01) - & 5.9920e+04 (4.08e+02) \\
21    & 8.3808e+04 (5.35e+02) - & 8.4996e+04 (1.02e+03) - & 8.1080e+04 (4.91e+02) \\
22    & 9.3039e+04 (2.95e+02) - & 9.3313e+04 (0.00e+00) - & 8.9020e+04 (2.34e+03) \\
23    & 4.5249e+04 (6.83e+03) + & 1.5542e+05 (1.97e+02) + & 1.5548e+05 (6.23e+01) \\
24    & 3.7741e+04 (2.22e+04) + & 1.3703e+05 (0.00e+00) + & 1.3703e+05 (0.00e+00) \\
25    & 6.0190e+04 (5.23e+02) - & 6.1068e+04 (6.52e+02) - & 5.5008e+04 (6.12e+02) \\
26    & 7.5768e+05 (9.82e+05) + & Inf (     NaN) =        & Inf (     NaN)        \\
27    & 5.8950e+04 (1.21e+04) + & Inf (     NaN) -        & 7.7368e+04 (9.38e+02) \\
28    & 5.6594e+04 (2.01e+03) = & Inf (     NaN) =        & Inf (     NaN)        \\
29    & 6.6585e+04 (1.03e+04) + & Inf (     NaN) =        & Inf (     NaN)        \\
30    & 8.5413e+04 (2.52e+04) = & Inf (     NaN) -        & 1.0317e+05 (7.68e+02) \\
31    & 2.1279e+05 (8.97e+04) + & 2.6464e+06 (4.00e+02) - & 2.6442e+06 (3.08e+02) \\
32    & 5.2032e+06 (7.72e+04) + & Inf (     NaN) =        & Inf (     NaN)        \\
33    & 1.0943e+05 (2.64e+04) + & Inf (     NaN) -        & 1.3579e+05 (3.90e+02) \\
34    & 9.0379e+04 (1.73e+03) + & Inf (     NaN) =        & Inf (     NaN)        \\
35    & 4.8146e+04 (1.03e+04) + & 9.2642e+04 (6.60e+02) - & 8.9550e+04 (2.74e+02) \\
36    & 2.8569e+04 (1.33e+04) + & 1.4592e+05 (1.03e+03) - & 1.4088e+05 (1.18e+03) \\
37    & 2.2731e+04 (8.51e+03) + & 1.9064e+05 (1.05e+03) - & 1.8622e+05 (1.25e+03) \\
38    & 2.0611e+04 (5.87e+03) + & Inf (     NaN) -        & 7.3264e+04 (3.11e+04) \\
39    & 4.6659e+04 (1.74e+04) = & Inf (     NaN) -        & 1.4352e+05 (7.84e+04) \\
40    & 2.4802e+04 (6.65e+02) = & Inf (     NaN) -        & 1.1376e+05 (6.11e+04) \\
41    & 6.6064e+04 (1.72e+04) + & Inf (     NaN) -        & 2.4472e+05 (1.25e+03) \\
42    & 4.9675e+04 (1.28e+04) + & Inf (     NaN) -        & 2.1552e+05 (1.74e+04) \\
\hline
+/-/= & 21/14/7                 & 2/32/8                  & \multicolumn{1}{l}{} 
\\
\hline
\end{tabular}
\end{table*}

\begin{table*}[!hbpt]
\centering
\caption{Results of $TH_1$ with $r = 5$}
\setlength{\tabcolsep}{6.5pt}
\centering\small
\begin{tabular}{lllllllll}
\hline
Pro   & AEDGA1                  & AEDGA2                  & AEDGA3                \\
\hline
1     & 2.9503e+04 (1.27e+04) + & 1.9994e+05 (0.00e+00) - & 7.8048e+04 (3.14e+04) \\
2     & 9.5719e+03 (6.32e+02) + & 1.1860e+04 (1.46e+02) - & 1.1374e+04 (1.58e-12) \\
3     & 1.1382e+04 (5.45e+02) + & Inf (     NaN) =        & Inf (     NaN)        \\
4     & 2.1275e+04 (1.99e+04) + & Inf (     NaN) =        & Inf (     NaN)        \\
5     & 1.0534e+04 (1.40e+03) = & Inf (     NaN) -        & 1.0820e+04 (5.62e+02) \\
6     & 1.0977e+04 (9.80e+02) = & Inf (     NaN) -        & 1.0408e+04 (1.14e+03) \\
7     & 6.6074e+05 (1.52e+05) + & 1.3749e+06 (1.96e+04) - & 1.3096e+06 (6.50e+03) \\
8     & 1.2416e+04 (2.44e+03) = & 1.6731e+04 (2.63e+02) - & 1.5891e+04 (2.25e+02) \\
9     & 1.2346e+04 (2.46e+03) + & Inf (     NaN) -        & 3.9572e+04 (7.90e+03) \\
10    & 1.6659e+04 (1.62e+03) = & 6.0593e+04 (4.38e+02) - & 4.0548e+04 (1.76e+04) \\
11    & 2.0402e+04 (4.68e+03) + & 5.7893e+04 (2.45e+02) - & 5.3331e+04 (4.65e+02) \\
12    & 1.5163e+04 (3.42e+03) + & 6.8854e+04 (2.95e+02) - & 6.1799e+04 (4.99e+03) \\
13    & 1.7813e+04 (4.03e+03) + & 6.0383e+04 (6.20e+02) - & 5.8169e+04 (4.17e+02) \\
14    & 1.5154e+04 (3.13e+03) + & 4.6767e+04 (4.24e+02) - & 4.4702e+04 (4.99e+02) \\
15    & 2.1862e+04 (5.18e+03) + & 6.4222e+04 (1.75e+02) - & 6.1073e+04 (7.11e+02) \\
16    & 5.2281e+04 (1.10e+02) - & 5.2537e+04 (0.00e+00) - & 5.1128e+04 (3.46e+02) \\
17    & 2.0658e+04 (5.01e+03) + & 7.4839e+04 (7.64e+02) = & 7.4788e+04 (1.30e+03) \\
18    & 1.6736e+04 (5.15e+03) + & 7.3366e+04 (4.59e+02) - & 6.7822e+04 (8.78e+03) \\
19    & 2.2485e+04 (1.63e+03) + & 7.2833e+04 (1.41e+03) - & 6.9278e+04 (6.87e+02) \\
20    & 4.3496e+04 (1.24e+04) = & 6.2042e+04 (1.18e+02) - & 5.9282e+04 (6.50e+02) \\
21    & 2.2985e+04 (5.46e+03) + & 8.3062e+04 (1.44e+03) - & 8.0820e+04 (6.48e+02) \\
22    & 2.3256e+04 (6.28e+03) + & 9.2627e+04 (1.07e+03) - & 8.7262e+04 (8.50e+02) \\
23    & 2.4024e+04 (2.59e+03) + & 1.5467e+05 (9.93e+02) - & 7.7899e+04 (2.04e+04) \\
24    & 2.3261e+04 (6.89e+03) + & 1.3611e+05 (1.11e+03) + & 1.3703e+05 (0.00e+00) \\
25    & 6.0368e+04 (5.15e+02) - & 6.0872e+04 (6.68e+02) - & 5.4920e+04 (7.35e+02) \\
26    & 1.7728e+06 (2.04e+04) + & Inf (     NaN) =        & Inf (     NaN)        \\
27    & 3.0411e+06 (4.86e+04) + & Inf (     NaN) =        & Inf (     NaN)        \\
28    & 3.6217e+06 (5.77e+04) + & Inf (     NaN) =        & Inf (     NaN)        \\
29    & 3.7709e+06 (6.56e+04) + & Inf (     NaN) =        & Inf (     NaN)        \\
30    & 5.6814e+06 (6.95e+04) + & Inf (     NaN) =        & Inf (     NaN)        \\
31    & 1.8272e+05 (9.56e+04) + & Inf (     NaN) -        & 1.4961e+06 (5.14e+05) \\
32    & 5.2796e+06 (6.48e+04) + & Inf (     NaN) =        & Inf (     NaN)        \\
33    & 8.6598e+06 (1.13e+05) + & Inf (     NaN) =        & Inf (     NaN)        \\
34    & 1.0115e+07 (1.09e+05) + & Inf (     NaN) =        & Inf (     NaN)        \\
35    & 1.7313e+04 (4.19e+03) + & Inf (     NaN) -        & 4.7797e+04 (6.84e+03) \\
36    & 1.7680e+04 (4.78e+03) + & Inf (     NaN) -        & 4.1897e+04 (1.06e+04) \\
37    & 1.6610e+04 (1.03e+03) + & Inf (     NaN) -        & 4.6603e+04 (2.08e+04) \\
38    & 1.5303e+04 (1.15e+03) + & Inf (     NaN) -        & 3.2119e+04 (1.29e+04) \\
39    & 2.4515e+04 (4.94e+03) + & Inf (     NaN) -        & 5.0223e+04 (1.90e+04) \\
40    & 2.4176e+04 (2.35e+03) = & Inf (     NaN) =        & Inf (     NaN)        \\
41    & 8.1397e+04 (8.11e+03) + & Inf (     NaN) =        & Inf (     NaN)        \\
42    & 4.8164e+04 (1.04e+04) = & Inf (     NaN) -        & 5.4705e+04 (3.89e+02) \\
\hline
+/-/= & 33/2/7                  & 1/28/13                 & \multicolumn{1}{l}{} 
\\
\hline
\end{tabular}
\end{table*}

\begin{table*}[!hbpt]
\centering
\caption{Results of $TH_2$ with $r = 5$}
\setlength{\tabcolsep}{6.5pt}
\centering\small
\begin{tabular}{lllllllll}
\hline
Pro   & AEDGA1                  & AEDGA2                  & AEDGA3                \\
\hline
1     & 3.2817e+04 (0.00e+00) = & Inf (     NaN) -        & 3.2817e+04 (0.00e+00) \\
2     & 7.9159e+03 (8.83e+02) + & 1.2031e+04 (1.03e+02) - & 1.0863e+04 (1.55e+03) \\
3     & 4.5297e+04 (4.47e+03) + & Inf (     NaN) =        & Inf (     NaN)        \\
4     & 4.9104e+04 (4.02e+03) + & Inf (     NaN) =        & Inf (     NaN)        \\
5     & 3.9325e+04 (1.69e+04) = & Inf (     NaN) =        & Inf (     NaN)        \\
6     & 5.0572e+04 (3.69e+03) + & Inf (     NaN) =        & Inf (     NaN)        \\
7     & 4.6645e+05 (1.61e+05) + & 1.3892e+06 (2.36e+04) - & 1.2739e+06 (7.32e+04) \\
8     & 5.5702e+03 (2.53e+03) + & 1.6885e+04 (0.00e+00) - & 1.4946e+04 (1.92e+03) \\
9     & 1.2258e+04 (1.26e+03) + & Inf (     NaN) -        & 1.9114e+04 (7.24e+03) \\
10    & 1.4346e+04 (1.62e+03) + & Inf (     NaN) -        & 2.1034e+04 (4.64e+03) \\
11    & 1.3967e+04 (2.83e+03) + & 5.7944e+04 (1.30e+02) - & 5.1462e+04 (7.15e+03) \\
12    & 1.4211e+04 (3.18e+03) + & Inf (     NaN) -        & 2.9329e+04 (6.57e+03) \\
13    & 1.3446e+04 (1.95e+03) + & 6.0929e+04 (2.31e+02) - & 4.8235e+04 (1.40e+04) \\
14    & 1.2250e+04 (2.89e+03) + & 4.6974e+04 (1.64e+02) - & 3.9953e+04 (7.52e+03) \\
15    & 1.5860e+04 (2.69e+03) + & 6.4267e+04 (2.07e+02) - & 5.7955e+04 (8.46e+03) \\
16    & 2.1205e+04 (4.42e+03) + & 5.2489e+04 (1.07e+02) = & 5.2267e+04 (3.11e+02) \\
17    & 1.7880e+04 (3.15e+03) + & 7.5393e+04 (0.00e+00) - & 5.9054e+04 (1.44e+04) \\
18    & 1.4790e+04 (4.47e+03) + & 7.2987e+04 (5.42e+02) - & 4.7855e+04 (7.02e+03) \\
19    & 1.4404e+04 (3.59e+03) + & 7.4713e+04 (6.82e+02) - & 6.8155e+04 (3.89e+03) \\
20    & 2.0135e+04 (3.95e+03) + & 6.2095e+04 (0.00e+00) - & 5.9801e+04 (1.41e+03) \\
21    & 1.9583e+04 (3.88e+03) + & 8.3873e+04 (4.30e+02) - & 7.7140e+04 (7.70e+03) \\
22    & 1.9112e+04 (5.45e+03) + & 9.2749e+04 (1.09e+03) - & 8.3583e+04 (1.13e+04) \\
23    & 3.2453e+04 (1.02e+04) = & Inf (     NaN) -        & 3.7883e+04 (2.31e+03) \\
24    & 2.0669e+04 (5.68e+03) + & Inf (     NaN) -        & 4.8360e+04 (1.92e+04) \\
25    & 6.0742e+04 (6.08e+02) - & 6.1270e+04 (2.01e+02) - & 5.5543e+04 (6.62e+02) \\
26    & 1.7932e+06 (4.73e+04) + & Inf (     NaN) =        & Inf (     NaN)        \\
27    & 3.0463e+06 (5.27e+04) + & Inf (     NaN) =        & Inf (     NaN)        \\
28    & 3.6618e+06 (5.58e+04) + & Inf (     NaN) =        & Inf (     NaN)        \\
29    & 3.7468e+06 (2.29e+04) + & Inf (     NaN) =        & Inf (     NaN)        \\
30    & 5.6537e+06 (9.66e+04) + & Inf (     NaN) =        & Inf (     NaN)        \\
31    & 2.6145e+05 (8.81e+04) + & Inf (     NaN) -        & 1.3496e+06 (4.91e+03) \\
32    & 5.2891e+06 (9.14e+04) + & Inf (     NaN) =        & Inf (     NaN)        \\
33    & 8.6084e+06 (5.26e+04) + & Inf (     NaN) =        & Inf (     NaN)        \\
34    & 1.0046e+07 (1.88e+05) + & Inf (     NaN) =        & Inf (     NaN)        \\
35    & 1.0774e+04 (2.57e+03) = & Inf (     NaN) -        & 1.5628e+04 (9.11e+03) \\
36    & 1.5298e+04 (3.14e+03) + & Inf (     NaN) -        & 2.6989e+04 (9.01e+03) \\
37    & 1.9901e+04 (1.86e+03) = & Inf (     NaN) -        & 2.5404e+04 (1.18e+04) \\
38    & 1.5795e+04 (5.22e+01) + & Inf (     NaN) =        & Inf (     NaN)        \\
39    & 2.6742e+04 (1.18e+03) + & Inf (     NaN) =        & Inf (     NaN)        \\
40    & 2.4508e+04 (1.61e+03) + & Inf (     NaN) =        & Inf (     NaN)        \\
41    & 6.8476e+04 (1.96e+04) + & Inf (     NaN) =        & Inf (     NaN)        \\
42    & 5.2573e+04 (9.01e+03) + & Inf (     NaN) -        & 9.7312e+04 (7.06e+02) \\
\hline
+/-/= & 36/1/5                  & 0/25/17                 & \multicolumn{1}{l}{} 
\\
\hline
\end{tabular}
\end{table*}

\begin{table*}[!hbpt]
\centering
\caption{Results of $TH_1$ with $r = 8$}
\setlength{\tabcolsep}{6.5pt}
\centering\small
\begin{tabular}{lllllllll}
\hline
Pro   & AEDGA1                  & AEDGA2                  & AEDGA3                \\
\hline
1     & 3.2025e+04 (1.13e+03) = & Inf (     NaN) -        & 3.2817e+04 (0.00e+00) \\
2     & 6.7293e+03 (3.21e+02) = & 1.2167e+04 (3.12e+02) - & 1.0129e+04 (2.12e+03) \\
3     & 4.7800e+04 (3.43e+03) + & Inf (     NaN) =        & Inf (     NaN)        \\
4     & 4.7483e+04 (7.28e+03) + & Inf (     NaN) =        & Inf (     NaN)        \\
5     & 4.3084e+04 (2.08e+04) + & Inf (     NaN) =        & Inf (     NaN)        \\
6     & 3.5270e+04 (2.58e+04) + & Inf (     NaN) =        & Inf (     NaN)        \\
7     & 3.9445e+05 (1.02e+05) + & Inf (     NaN) -        & 1.0351e+06 (2.67e+05) \\
8     & 4.1959e+03 (1.91e+03) + & 1.6816e+04 (1.54e+02) - & 1.5883e+04 (1.81e+02) \\
9     & 1.2314e+04 (1.59e+03) + & Inf (     NaN) -        & 2.2366e+04 (5.32e+03) \\
10    & 1.2908e+04 (3.27e+03) + & Inf (     NaN) -        & 2.4376e+04 (6.49e+03) \\
11    & 1.4262e+04 (3.06e+03) + & Inf (     NaN) -        & 4.8318e+04 (7.26e+03) \\
12    & 1.3694e+04 (9.61e+02) + & Inf (     NaN) -        & 3.2068e+04 (9.43e+03) \\
13    & 1.3664e+04 (2.74e+03) + & Inf (     NaN) -        & 5.4451e+04 (7.91e+03) \\
14    & 1.2335e+04 (1.51e+03) + & 4.6770e+04 (1.78e+02) - & 4.2378e+04 (5.24e+03) \\
15    & 1.6389e+04 (3.89e+03) + & Inf (     NaN) -        & 5.6243e+04 (7.52e+03) \\
16    & 1.5877e+04 (2.70e+03) + & 5.2537e+04 (0.00e+00) - & 5.0896e+04 (5.78e+02) \\
17    & 1.7752e+04 (4.68e+03) + & Inf (     NaN) -        & 3.5399e+04 (9.44e+03) \\
18    & 1.5066e+04 (3.22e+03) + & Inf (     NaN) -        & 5.3874e+04 (9.93e+03) \\
19    & 1.9880e+04 (1.57e+03) + & 7.3663e+04 (6.61e+02) - & 6.8934e+04 (5.98e+02) \\
20    & 1.6597e+04 (3.31e+03) + & 6.1911e+04 (2.66e+02) - & 5.8867e+04 (1.01e+03) \\
21    & 2.0674e+04 (4.69e+03) + & 8.3403e+04 (1.16e+03) - & 6.9174e+04 (1.61e+04) \\
22    & 1.4823e+04 (3.87e+03) + & 9.2124e+04 (1.10e+03) - & 7.6139e+04 (1.87e+04) \\
23    & 2.5632e+04 (6.17e+03) = & Inf (     NaN) -        & 3.0953e+04 (2.26e+03) \\
24    & 2.1204e+04 (6.43e+03) = & Inf (     NaN) -        & 2.8059e+04 (1.81e+03) \\
25    & 5.3940e+04 (1.62e+04) = & 6.0871e+04 (1.09e+03) - & 5.4953e+04 (1.04e+03) \\
26    & 1.8000e+06 (4.76e+04) + & Inf (     NaN) =        & Inf (     NaN)        \\
27    & 3.0607e+06 (3.80e+04) + & Inf (     NaN) =        & Inf (     NaN)        \\
28    & 3.6386e+06 (2.13e+04) + & Inf (     NaN) =        & Inf (     NaN)        \\
29    & 3.7330e+06 (5.62e+04) + & Inf (     NaN) =        & Inf (     NaN)        \\
30    & 5.6329e+06 (8.98e+04) + & Inf (     NaN) =        & Inf (     NaN)        \\
31    & 1.1992e+05 (3.23e+04) + & Inf (     NaN) -        & 1.1369e+06 (2.85e+04) \\
32    & 5.3078e+06 (5.02e+04) + & Inf (     NaN) =        & Inf (     NaN)        \\
33    & 8.6177e+06 (9.44e+04) + & Inf (     NaN) =        & Inf (     NaN)        \\
34    & 1.0065e+07 (6.23e+04) + & Inf (     NaN) =        & Inf (     NaN)        \\
35    & 9.8556e+03 (1.01e+03) = & Inf (     NaN) -        & 9.0114e+03 (8.66e+02) \\
36    & 1.4647e+04 (1.70e+03) + & Inf (     NaN) -        & 2.9337e+04 (4.17e+03) \\
37    & 1.6235e+04 (1.36e+03) + & Inf (     NaN) -        & 2.3494e+04 (9.04e+03) \\
38    & 1.5385e+04 (9.69e+02) + & Inf (     NaN) =        & Inf (     NaN)        \\
39    & 2.5877e+04 (6.24e+03) = & Inf (     NaN) =        & Inf (     NaN)        \\
40    & 2.3250e+04 (2.76e+03) = & Inf (     NaN) =        & Inf (     NaN)        \\
41    & 1.3760e+06 (7.42e+05) + & Inf (     NaN) =        & Inf (     NaN)        \\
42    & 4.9340e+04 (9.56e+03) = & Inf (     NaN) -        & 5.4642e+04 (4.14e+02) \\
\hline
+/-/= & 33/0/9                  & 0/26/16                 & \multicolumn{1}{l}{} 
\\
\hline
\end{tabular}
\end{table*}

\begin{table*}[!hbpt]
\centering
\caption{Results of $TH_2$ with $r = 8$}
\setlength{\tabcolsep}{6.5pt}
\centering\small
\begin{tabular}{lllllllll}
\hline
Pro   & AEDGA1                  & AEDGA2                  & AEDGA3                \\
\hline
1     & 3.1467e+04 (3.02e+03) = & Inf (     NaN) -        & 3.2817e+04 (0.00e+00) \\
2     & 6.9790e+03 (4.60e+02) + & 1.2152e+04 (8.24e+01) - & 1.1353e+04 (3.87e+02) \\
3     & 4.4183e+04 (4.56e+03) + & Inf (     NaN) =        & Inf (     NaN)        \\
4     & 4.9546e+04 (4.13e+03) + & Inf (     NaN) =        & Inf (     NaN)        \\
5     & 3.0763e+04 (2.01e+04) + & Inf (     NaN) =        & Inf (     NaN)        \\
6     & 3.4253e+04 (2.37e+04) + & Inf (     NaN) =        & Inf (     NaN)        \\
7     & 3.2729e+05 (4.93e+04) + & 1.3915e+06 (2.12e+04) - & 1.2383e+06 (1.54e+05) \\
8     & 4.6973e+03 (1.94e+03) + & 1.6885e+04 (0.00e+00) - & 1.4125e+04 (2.26e+03) \\
9     & 1.2067e+04 (2.48e+03) + & Inf (     NaN) -        & 2.1002e+04 (4.35e+03) \\
10    & 1.2845e+04 (2.69e+03) + & Inf (     NaN) -        & 2.2687e+04 (4.84e+03) \\
11    & 1.3195e+04 (4.82e+03) + & 5.8003e+04 (8.13e-12) - & 4.9860e+04 (8.50e+03) \\
12    & 1.3175e+04 (2.20e+03) + & Inf (     NaN) -        & 3.4008e+04 (8.67e+03) \\
13    & 1.3606e+04 (3.11e+03) + & 6.1014e+04 (4.04e+01) = & 5.5242e+04 (9.27e+03) \\
14    & 1.3740e+04 (3.74e+03) + & 4.7242e+04 (1.00e+02) - & 4.4783e+04 (3.77e+02) \\
15    & 1.5843e+04 (5.74e+03) + & 6.4708e+04 (5.48e+02) - & 6.1791e+04 (1.61e+02) \\
16    & 2.4541e+04 (9.16e+03) + & 5.2537e+04 (0.00e+00) - & 5.1603e+04 (3.63e+02) \\
17    & 2.0503e+04 (2.40e+03) + & 7.5382e+04 (2.44e+01) - & 5.6356e+04 (1.84e+04) \\
18    & 1.8214e+04 (1.50e+03) + & 7.3490e+04 (1.81e+02) - & 4.8832e+04 (1.05e+04) \\
19    & 1.6600e+04 (2.91e+03) + & 7.5315e+04 (3.76e+02) - & 6.9459e+04 (6.19e+02) \\
20    & 1.8827e+04 (4.03e+03) + & 6.2095e+04 (0.00e+00) - & 5.9269e+04 (9.00e+02) \\
21    & 1.9040e+04 (5.57e+03) + & 8.4604e+04 (8.95e+02) - & 8.0185e+04 (1.71e+03) \\
22    & 1.9218e+04 (6.27e+03) + & 9.3136e+04 (3.95e+02) - & 7.6974e+04 (2.47e+04) \\
23    & 2.8475e+04 (5.26e+03) = & Inf (     NaN) -        & 3.2425e+04 (9.54e+03) \\
24    & 1.5645e+04 (5.24e+03) + & Inf (     NaN) -        & 7.8936e+04 (1.88e+04) \\
25    & 6.0587e+04 (6.11e+02) - & 6.1360e+04 (0.00e+00) - & 5.4574e+04 (5.44e+02) \\
26    & 1.7787e+06 (4.37e+04) + & Inf (     NaN) =        & Inf (     NaN)        \\
27    & 3.0377e+06 (3.99e+04) + & Inf (     NaN) =        & Inf (     NaN)        \\
28    & 3.6637e+06 (2.43e+04) + & Inf (     NaN) =        & Inf (     NaN)        \\
29    & 3.7666e+06 (2.47e+04) + & Inf (     NaN) =        & Inf (     NaN)        \\
30    & 5.6541e+06 (6.59e+04) + & Inf (     NaN) =        & Inf (     NaN)        \\
31    & 1.9519e+05 (4.18e+04) = & Inf (     NaN) -        & 9.1055e+05 (5.90e+05) \\
32    & 5.1689e+06 (8.22e+04) + & Inf (     NaN) =        & Inf (     NaN)        \\
33    & 8.6491e+06 (8.48e+04) + & Inf (     NaN) =        & Inf (     NaN)        \\
34    & 8.0506e+06 (4.47e+06) + & Inf (     NaN) =        & Inf (     NaN)        \\
35    & 1.0685e+04 (2.35e+03) = & Inf (     NaN) -        & 1.7496e+04 (1.69e+04) \\
36    & 1.4176e+04 (2.41e+03) + & Inf (     NaN) -        & 3.2953e+04 (5.99e+03) \\
37    & 1.7279e+04 (1.22e+03) = & Inf (     NaN) -        & 2.1035e+04 (6.12e+03) \\
38    & 1.5819e+04 (0.00e+00) + & Inf (     NaN) -        & 2.3527e+04 (5.75e+03) \\
39    & 2.5400e+04 (2.43e+03) + & Inf (     NaN) -        & 2.9441e+04 (3.31e+02) \\
40    & 2.5227e+04 (0.00e+00) = & Inf (     NaN) =        & Inf (     NaN)        \\
41    & 7.0752e+04 (1.65e+04) + & Inf (     NaN) =        & Inf (     NaN)        \\
42    & 5.9086e+04 (2.49e+04) = & Inf (     NaN) -        & 5.5403e+04 (3.48e+02) \\
\hline
+/-/= & 34/1/7                  & 0/27/15                 & \multicolumn{1}{l}{} 
\\
\hline
\end{tabular}
\end{table*}